\documentclass{article}

 \usepackage[main, final]{neurips_2026}

\usepackage[utf8]{inputenc} 
\usepackage[T1]{fontenc}    
\usepackage{hyperref}       
\usepackage{url}            
\usepackage{booktabs}       
\usepackage{amsfonts}       
\usepackage{nicefrac}       
\usepackage{microtype}      
\usepackage{xcolor}         

\usepackage{graphicx}
\usepackage{subcaption}

\usepackage{amsmath,amsfonts,bm}

\def\eqref#1{equation~\ref{#1}}

\def\1{\bm{1}}

\def\rva{{\mathbf{a}}}

\def\rvv{{\mathbf{v}}}

\def\rvx{{\mathbf{x}}}

\def\vzero{{\bm{0}}}
\def\vone{{\bm{1}}}

\def\va{{\bm{a}}}

\def\ve{{\bm{e}}}

\def\vg{{\bm{g}}}
\def\vh{{\bm{h}}}

\def\vm{{\bm{m}}}

\def\vx{{\bm{x}}}

\def\evx{{x}}

\def\mE{{\bm{E}}}

\def\mI{{\bm{I}}}

\def\mW{{\bm{W}}}
\def\mX{{\bm{X}}}

\DeclareMathAlphabet{\mathsfit}{\encodingdefault}{\sfdefault}{m}{sl}
\SetMathAlphabet{\mathsfit}{bold}{\encodingdefault}{\sfdefault}{bx}{n}

\def\gG{{\mathcal{G}}}

\def\gL{{\mathcal{L}}}

\def\gS{{\mathcal{S}}}

\def\gU{{\mathcal{U}}}

\def\gX{{\mathcal{X}}}

\def\sR{{\mathbb{R}}}
\def\sS{{\mathbb{S}}}

\def\sV{{\mathbb{V}}}

\usepackage{booktabs}
\usepackage{float}
\usepackage{cancel}
\usepackage{multirow}
\usepackage{wrapfig}
\newtheorem{theorem}{Theorem}[section]
\newtheorem{proposition}[theorem]{Proposition}
\newtheorem{assumption}[theorem]{Assumption}
\title{Unrestrained Simplex Denoising for Discrete Data \\ A Generative Method Applied to Graphs}

\author{%
  Yoann Boget \\
  University of Applied Sciences \\
  Western Switzerland (HES-SO)\\
  Department of Computer Science\\
  University of Geneva \\
  \texttt{yoann.boget@hes-so.ch} \\
  \And
  Alexandros Kalousis \\
    University of Applied Sciences \\
    Western Switzerland (HES-SO)\\
}

\begin{document}

\maketitle

\begin{abstract}
    Denoising models such as Diffusion or Flow Matching have recently advanced generative modeling for discrete structures, yet most approaches operate directly in the discrete state space, leading to abrupt state changes. We introduce simplex denoising, a simple yet effective generative framework that operates on the probability simplex. The key idea is a non-Markovian noising scheme in which, for a given clean data point, noisy representations at different times are conditionally independent. While preserving the theoretical guarantees of denoising-based generative models, our method removes unnecessary constraints, thereby improving performance and simplifying the formulation. Empirically, \emph{unrestrained simplex denoising} surpasses strong discrete diffusion and flow-matching baselines across synthetic and real-world graph benchmarks. These results highlight the probability simplex as an effective framework for discrete generative modeling.
\end{abstract}

\section{Introduction}

Denoising models, such as discrete diffusion and discrete flow matching, have substantially advanced generative modeling for discrete structures \citep{discdiff_austin, discdiff_campbell, Discrete_FlowMatching_campbell, discrete_flow_matching_gat}, including graphs \citep{discdiff_haefeli, digress, DeFog, SID}. Yet, surprisingly few works study denoising on simplices for modeling discrete data \citep{richemond2022, avdeyev23a, DirichletFlowMating, BetaGraph}. 
This may result from the pathological behavior of interpolant approaches on the simplex \citep{DirichletFlowMating}, which creates abrupt discontinuities in the support of the noisy distributions (Section~\ref{sec:noising}). 

Moreover, standard approaches such as diffusion or flow matching strongly constrain the reverse dynamics by enforcing a continuous denoising trajectory in the continuous-time limit, i.e., $\lim_{h \to 0} p_{t+h \mid t}(\rvx_{t+h} \mid \rvx_{t}) = \delta(\rvx_{t+h} - \rvx_{t})$. 
In discrete diffusion, this constraint has been linked to compounding denoising errors \citep{compounding_denoising_error, SID, Non_Markov_discrete}. We argue that a similar issue arises in the continuous-state setting. 

To overcome these issues, we introduce \emph{Unrestrained Simplex Denoising (UNSID)}, a simple yet effective generative model for discrete data that operates directly on the probability simplex. Our model leverages a non-Markovian noising process in which noisy states are conditionally independent given the clean input. By removing dependence on the previous noisy state, our formulation eliminates unnecessary constraints and simplifies the denoising process.

Our key contributions are as follows:

\begin{itemize}
    \item We introduce a new non-Markovian noising approach for categorical data that operates on the probability simplex (Section~\ref{sec:method}) and apply it to graph generation. The formulation substantially simplifies standard denoising paradigms such as diffusion or flow matching and addresses the issue of compounding denoising errors. 
    
    \item We provide theoretical guarantees, proving convergence of the proposed procedure assuming access to the true denoising distribution $p_{1 \mid t}(\rvx_1 \mid \rvx_t)$ (Section~\ref{sec:method}).
    
    \item Leveraging explicit probability paths, we propose new theoretical tools for the analysis of the noising dynamics on the probability simplex. Specifically, we introduce the Voronoi probability and derive a closed-form expression for it under the Dirichlet probability path (Section~\ref{sec:noising}). 

    \item Empirically, our unrestrained simplex denoising method surpasses strong discrete diffusion and flow-matching baselines on synthetic and real-world benchmarks for graph generation, advancing the state of the art on three of the five datasets considered while remaining competitive with the strongest baselines on the other two (Section~\ref{sec:eval}).
\end{itemize}
\section{Background}\label{sec:background}

In this paper, we focus on modeling multivariate categorical data, with applications to graphs that may have discrete node and edge attributes. In this section, we first review the related literature on denoising models for discrete data, focusing on graph generative models, and then present the non-Markovian noising processes relevant to our method. 

\subsection{Related Work}

Recently, \emph{denoising models} for discrete data have attracted considerable attention. Coupled with equivariant Graph Neural Networks (GNNs), they have become the dominant approach to graph generation, as they have proven highly effective (see Appendix~\ref{ap:related_works} for a comprehensive review of generative methods for graphs). 
We group existing denoising approaches for discrete data in general, and for graphs in particular, into three broad families: (1) continuous relaxation, (2) discrete diffusion, and (3) simplex-based denoising.

Approaches based on \emph{continuous relaxation} embed discrete variables into the unconstrained Euclidean space $\sR^K$ and train conventional continuous diffusion models with a standard Gaussian prior \citep{gdss, drum, Niu2020}. A related class of methods similarly injects continuous noise, but trains the denoiser to predict the categorical distribution directly \citep{CatFlow, graphBFN}. These approaches have two important shortcomings. First, points on the simplex have $K-1$ degrees of freedom; embedding them in $\sR^K$ adds a redundant dimension that carries no information and introduces unnecessary noise. Second, simplex-based noising naturally defines explicit and interpretable probability paths over categorical distributions (Section~\ref{sec:noising}), whereas Euclidean noising requires nonlinear projection or renormalization steps whose behavior is harder to characterize.

Initiated by D3PM \citep{discdiff_austin}, \emph{discrete diffusion} corrupts data by injecting discrete noise and models transitions directly in the discrete domain. Subsequent work extends this idea to continuous time \citep{discdiff_campbell} and to discrete flow matching \citep{Discrete_FlowMatching_campbell, discrete_flow_matching_gat}. 
In graph generation, this approach has been widely adopted \citep{discdiff_haefeli, continuous_time_graph_diffusion, SID}, with DiGress \citep{digress} and DeFoG \citep{DeFog} standing as two prominent representatives. However, transitioning abruptly between discrete states during denoising can drastically alter the structural properties of the instance; removing a single edge, for example, may disconnect the graph. Furthermore, GNNs employed as denoisers exhibit limited expressivity when operating on graphs with discrete node and/or edge attributes (see Appendix~\ref{ap:expressivity}). To address this limitation, discrete graph denoising methods typically require costly, handcrafted structural features.

In \emph{simplex-based denoising}, categorical variables are represented as points on the probability simplex, with one-hot encodings corresponding to its vertices. One strategy diffuses in $\sR^K$ and leverages distributional correspondences, such as gamma--Dirichlet or normal--logistic-normal, to map noisy samples back onto the simplex \citep{richemond2022, floto2023}. A second strategy generates noise \emph{directly} on the simplex \citep{avdeyev23a, DirichletFlowMating, simplex_diffusion_ssd}. In this work, we adopt the latter approach.

In graph generation, Beta diffusion \citep{BetaGraph} operates on binary data, and \citet{CatFlow} implement the Dirichlet flow-matching formulation as a baseline but report poor performance. No model has yet generalized successfully to the full probability simplex, suggesting that simplex-based graph generation is more challenging than it appears.

Yet, noise on the simplex presents notable advantages over discrete noise. (i) In general, continuous vectors contain more information than one-hot representations of the same dimension. (ii) Continuous noise on the simplex avoids discontinuities, most notably graph disconnection, and yields edge weights that naturally encode proximity, providing each node with information about its position relative to all others. (iii) Continuous vectors act as unique identifiers for nodes and edges, thereby alleviating the expressivity limitations of GNNs \citep{Loukas_WhatGNN} and limiting the need for additional handcrafted features.

\subsection{Non-Markovian Noising Processes}

Standard formulations of diffusion and flow-matching models are Markovian, such that $q_{t\mid t+h,1}(\rvx_t \mid \rvx_{t+h}, \rvx_1) = q_{t\mid t+h}(\rvx_t \mid \rvx_{t+h})$, where $h > 0$ denotes a small but not necessarily infinitesimal time interval such that $t + h \leq 1$. To our knowledge, DDIM \citep{DDIM} was the first diffusion model to adopt a non-Markovian noising scheme, defining the forward kernel $q_{t\mid t+h,1}$ to depend on both the preceding noisy state and the clean sample. 
Recently, a family of non-Markovian formulations defines the discrete noising process solely as a function of the noise level $\alpha_t$ and the clean data $\rvx_1$, such that $q_{t\mid t+h,1}(\rvx_t\mid \rvx_{t+h}, \rvx_1) = q_{t\mid 1}(\rvx_t\mid \rvx_1)$. During denoising, this independence assumption removes direct dependencies on previous noisy states, thereby mitigating error accumulation (``compounding denoising errors'') \citep{compounding_denoising_error}. To the best of our knowledge, our work is the first to extend this class of non-Markovian methods to simplex-based denoising; it can be interpreted as the simplex counterpart of Simple Iterative Denoising (SID) \citep{SID}.
\section{Noising on the Simplex}\label{sec:noising}

Noising processes defined on the probability simplex pose specific challenges. In particular, as shown by \citet{DirichletFlowMating}, specifying $q_t$ as an \emph{interpolant} between endpoint distributions $q_1$ and $q_0$, as is common in discrete diffusion and flow matching, restricts the support of $q_t$, leaving large regions of the simplex out of distribution and creating jump discontinuities in $q_t$. In this section, we formalize the setting, introduce an explicit, well-behaved parameterization of the noising path, and propose new theoretical tools for analyzing the noising dynamics. 

\subsection{Problem Setting}

\begin{figure} 
    \begin{small}
    \begin{center} 
    \begin{subfigure}[b]{0.18\textwidth}
        \includegraphics[width=\textwidth]{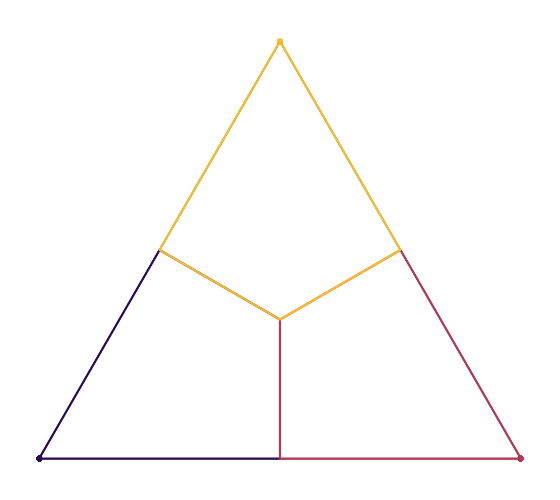} 
        \caption*{$t = 1$ }
    \end{subfigure}
    \begin{subfigure}[b]{0.18\textwidth}
        \includegraphics[width=\textwidth]{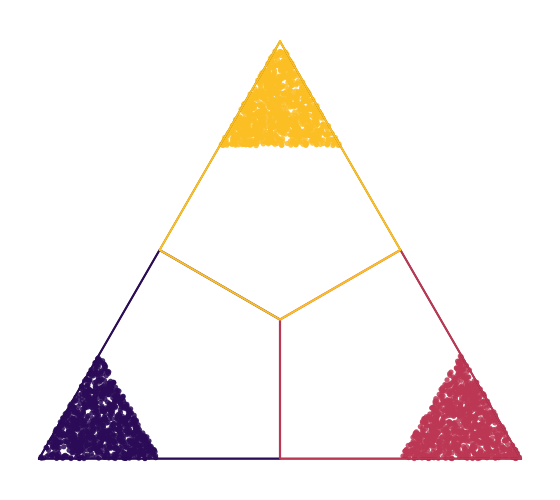} 
        \caption*{$t = 0.75$ }
    \end{subfigure}
    \begin{subfigure}[b]{0.18\textwidth}
        \includegraphics[width=\textwidth]{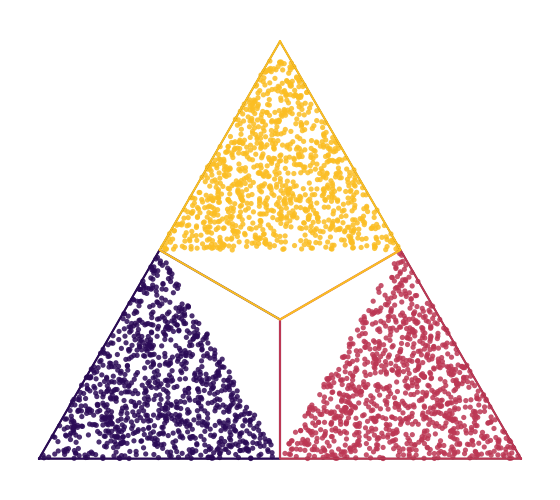} 
        \caption*{$t = 0.5$ }
    \end{subfigure}
    \begin{subfigure}[b]{0.18\textwidth}
        \includegraphics[width=\textwidth]{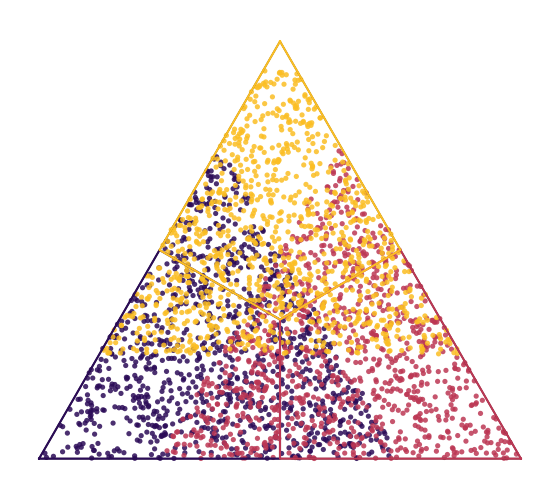} 
        \caption*{$t = 0.25$ }
    \end{subfigure}
    \begin{subfigure}[b]{0.18\textwidth}
        \includegraphics[width=\textwidth]{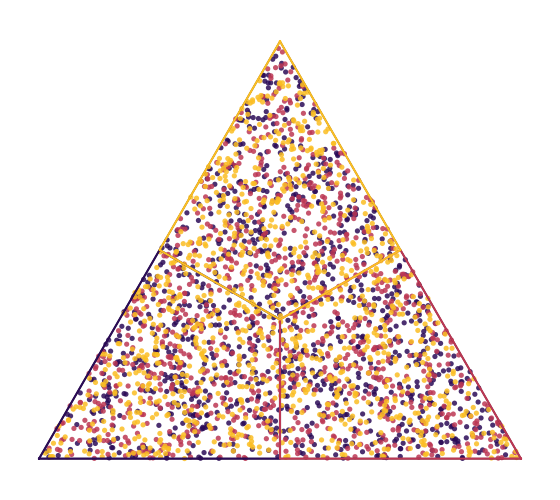} 
        \caption*{$t = 0$ }
    \end{subfigure}
    
    \begin{subfigure}[b]{0.18\textwidth}
        \includegraphics[width=\textwidth]{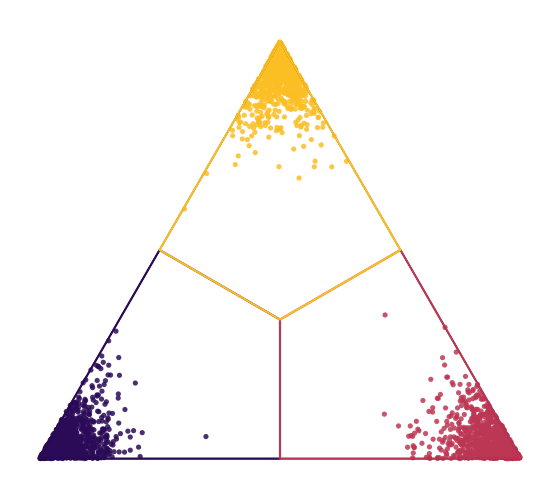} 
        \caption*{$t = 1$ }
    \end{subfigure}
    \begin{subfigure}[b]{0.18\textwidth}
        \includegraphics[width=\textwidth]{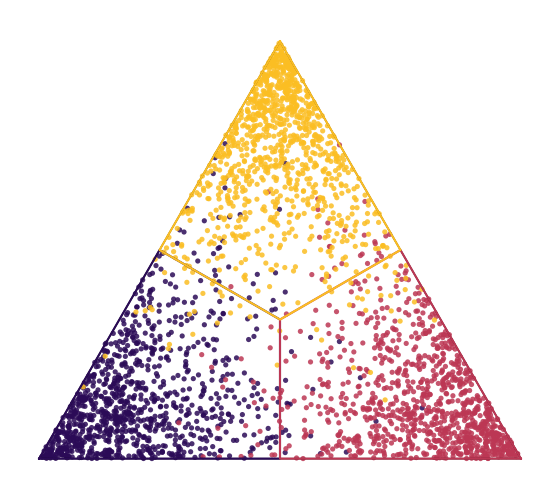} 
        \caption*{$t = 0.75$ }
    \end{subfigure}
    \begin{subfigure}[b]{0.18\textwidth}
        \includegraphics[width=\textwidth]{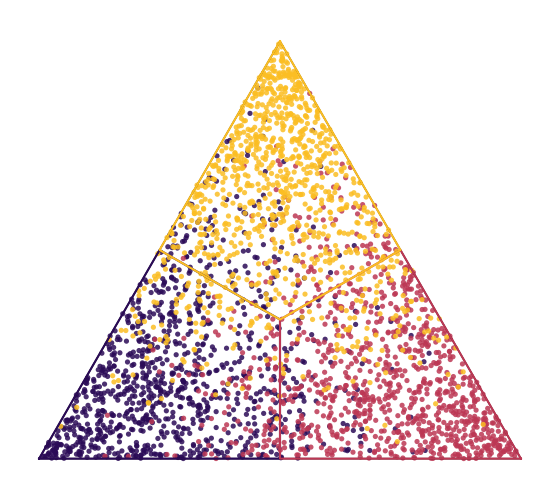} 
        \caption*{$t = 0.5$ }
    \end{subfigure}
    \begin{subfigure}[b]{0.18\textwidth}
        \includegraphics[width=\textwidth]{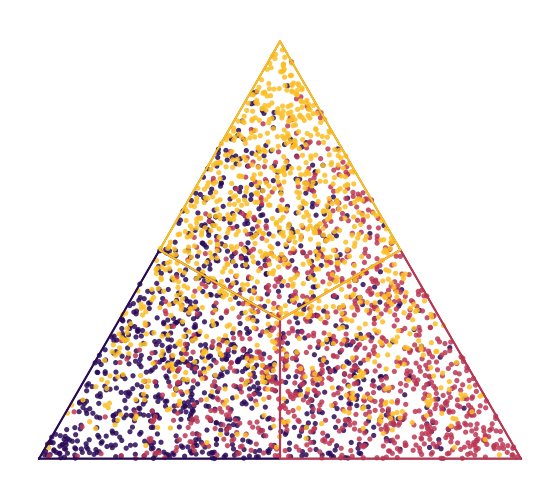} 
        \caption*{$t = 0.25$ }
    \end{subfigure}
    \begin{subfigure}[b]{0.18\textwidth}
        \includegraphics[width=\textwidth]{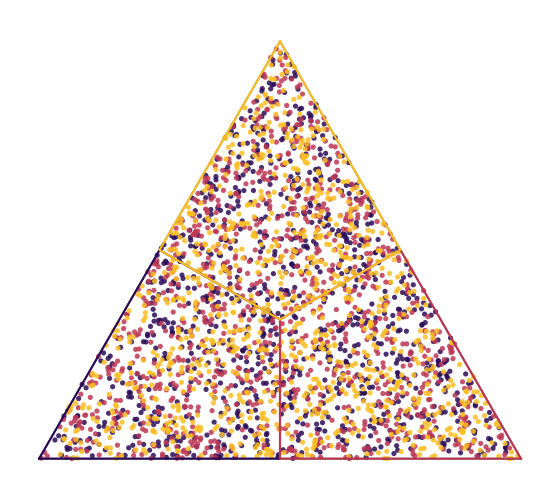} 
        \caption*{$t = 0$ }
    \end{subfigure}
    
    \end{center} 

    \caption{Upper row: Noising with a linear interpolant creates discontinuities and, for $t > 0.5$, all points remain in the Voronoi region of their original vertex.
    Lower row: Noising with an explicit parametric probability path avoids discontinuities.} 
    \label{fig:prob_path} 
    \end{small}
\end{figure}

Let $\sS_K$ denote the $(K-1)$-dimensional probability simplex,
$
\sS_K \,=\, \bigl\{\, \vx \in \sR^K \,\big|\, \vone_K^\top \vx = 1,\; x_i > 0 \;\forall i \,\bigr\}.
$
We represent categorical node/edge attributes as vertices of the simplex, i.e., $\rvx_1 \in \{\,\ve_i\,\}_{i \in [K]}$, where $\ve_i$ denotes the $i$th standard basis vector.

A noising process on the simplex is a family of conditional distributions $(q_t)_{t \in [0,1)}$ with $q_t(\,\cdot \mid \rvx_1)$ supported on $\sS_K$ for all $t \in [0,1)$, and boundary conditions
$
\lim_{t \to 1} q_t(\rvx_t \mid \rvx_1) = \delta(\rvx_t - \rvx_1),
\quad
q_0(\rvx_0 \mid \rvx_1) = q_0(\rvx_0),
$
i.e., the terminal distribution collapses to the clean vertex and the initial distribution is independent of $\rvx_1$. Such processes can be specified either implicitly via a stochastic interpolant or explicitly by choosing a parametric family on the simplex.

The interpolant construction is standard in discrete diffusion and flow matching, and also underlies SID \citep{SID}, which can be viewed as the discrete analogue of our approach. This construction sets
$\rvx_t = \alpha_t\rvx_1 + (1-\alpha_t)\rvx_0$, with $\rvx_1\sim p_{\text{data}}$, $\rvx_0 \sim q_0$, and $\alpha_t \in [0,1]$. 
However, this formulation exhibits what \citet{DirichletFlowMating} call pathological properties. In particular, within the interpolant framework, the posterior
$
P_{1 \mid t}(\rvx_1 \mid \rvx_t) \propto (\rvx_1) q_t(\rvx_t \mid \rvx_1)p_{\text{data}}
$
has support on at most $K-1$ vertices for all noise levels $\alpha_t > 1/K$, and becomes trivial, i.e., concentrated on a single vertex, for $\alpha_t > 0.5$. For all $t$, $q_t(\rvx_t \mid \rvx_1)$ is a uniform distribution over a subset of $\sS_K$, and zero elsewhere, producing support discontinuities. The upper row of Figure~\ref{fig:prob_path} illustrates such a process on the $2$-simplex. Such piecewise-constant behavior is ill-suited for approximating posteriors with differentiable neural networks. We avoid these pathologies by specifying an explicit, parametric probability path on the simplex.

Dirichlet distributions (and their mixtures) offer a natural, tractable choice on $\sS_K$. Recall the Dirichlet density function: 
$
\mathrm{Dir}(\vx;\bm{\alpha}) = \mathrm{B}(\bm{\alpha})^{-1} \prod_{i=1}^K x_i^{\alpha_i - 1},
$
where $\mathrm{B}$ is the multivariate beta function (see Equation~\ref{eq:beta}). For instance, a simple yet useful noising path is defined as 
$
q_t(\rvx_t \mid \rvx_1) = \mathrm{Dir}\bigl(\rvx_t; \vone + \alpha_t\rvx_1 \bigr),
$
with a non-decreasing noise schedule $\alpha_t \in \sR_{\ge 0}$ such that $\alpha_0 = 0$ and $\lim_{t \to 1} \alpha_t = \infty$, e.g., $\alpha_t = -a\log(1-t)$, where $a > 0$ is a hyperparameter. This way, $q_t$ has full support on $\sS_K$ for any $t \in [0,1)$, thereby avoiding the support restrictions of interpolant-based methods. The lower row of Figure~\ref{fig:prob_path} illustrates noising based on a parametric probability path. 

\subsection{Voronoi Probabilities}

The Voronoi regions of the simplex vertices provide an intuitive lens on how a noising process behaves. Let $\sV_k \subseteq \sS_K$ denote the Voronoi region of vertex $k$, defined as $\sV_k = \{\vx \in \sS_K \mid d(\vx,\ve_k) \leq d(\vx,\ve_j) \; \forall j \neq k\}$, with $d(\cdot, \cdot)$ the Euclidean distance. 
Geometrically, $\sV_k$ is the polytope of points whose nearest simplex vertex is 
$\ve_k$. Equivalently, we have $\vx \in \sV_k \iff x_k = \max_j x_j$. 

Given the conditional noising distribution $q_t(\rvx_t \mid \rvx_1)$, we define the \emph{Voronoi probability} for vertex $k$ as the conditional probability $P_{v_k}(\rvx_t \in \sV_k \mid \rvx_1 = \ve_k)$, i.e., the probability that the closest vertex to $\rvx_t$ is still its origin $\rvx_1$. Intuitively, this quantity summarizes how quickly the unconditional distribution $q_t(\rvx_t)$ ``mixes''.
For the stochastic-interpolant construction, for instance, one has $P_{v_i}(\rvx_t \in \sV_i \mid \rvx_1 = \ve_i) = 1$ for all $t$ such that $\alpha_t > 0.5$; consequently, the posterior over vertices collapses to the Voronoi indicator $P_{1 \mid t}(\rvx_1 = \ve_k \mid \rvx_t) = [\rvx_t \in \sV_k]$.

For the probability path $\mathrm{Dir}(\rvx_t; \vone + \alpha_t\rvx_1)$ introduced above, the Voronoi probability admits a closed-form expression:

\begin{proposition}\label{prop:voronoi}
    Assume $\rvx_t \mid \rvx_1 = \ve_i \sim \mathrm{Dir}(\vone + \alpha_t\ve_i)$. Then,
    \begin{equation}
       P_{v_i}(\rvx_t \in \sV_i \mid \rvx_1 = \ve_i) = \sum_{k=0}^{K-1} \frac{(-1)^k \binom{K-1}{k}}{(k+1)^{\alpha_t+1}}.
    \end{equation}
\end{proposition}

We present all proofs in Appendix~\ref{ap:proofs}.

We leverage Voronoi probabilities to design and calibrate explicit probability paths. In Appendix~\ref{ap:sched}, we evaluate our model across multiple values of the schedule hyperparameter $a$, demonstrating the usefulness of Voronoi probabilities for calibrating the noise schedule.
\section{Method}\label{sec:method}

In this section, we present our UNSID method. As with any other denoising model, it includes a noising process that progressively corrupts the data and a learned denoising process that operates backward.

We denote by $p := p_{\text{data}}$ an \emph{arbitrary} data distribution on $\gS^L$, where $\gS$ is a finite alphabet of size $K$ and $L$ is a finite number of dimensions, and by $\gX := \{\rvx_1 \in \gS^L : p(\rvx_1) > 0\}$ its support. 
Following Section~\ref{sec:noising}, we identify each symbol of $\gS$ with its one-hot encoding, i.e., $\gS = \{\ve_1, \dots, \ve_K\}$, so that clean instances $\rvx_1 \in \gS^L$ are vertices of the multi-simplex $\sS_K^L$. Throughout, $\rvx = \rvx^{(1:L)}$ denotes a full instance and $\rvx^{(i)}$ its $i$th dimension.
The multivariate forward (noising) process acts independently on each dimension $\rvx^{(i)}$, $i \in [L]$, noising instances on $\sS_K^L$.
In contrast, during the reverse process used at inference, dependencies across dimensions are captured by a learned denoiser that outputs element-wise logits conditioned on all dimensions $\rvx^{(1:L)}_t$. This setup follows prior work on discrete diffusion and flow matching \citep{discdiff_austin, digress, Discrete_FlowMatching_campbell, DirichletFlowMating, SID}. 

Our UNSID framework can be viewed as a simplex analogue of SID \citep{SID}. Extending SID to the continuous simplex setting is non-trivial and constitutes a significant methodological contribution of this work. We refer the reader to Appendix~\ref{ap:method_comparison} for a detailed comparison and a precise account of the contributions over SID.

\subsection{Noising}\label{sec:meth_noising}

Our noising mechanism relies on two key components: (i) a probability path that governs corruption at any continuous time $t \in [0,1)$, and (ii) a conditional-independence assumption across different times.

\paragraph{Probability path}
Since the forward process operates independently across dimensions, the distribution at time $t$ factorizes as
\begin{equation}
q_t(\rvx \mid \rvx_1) = \prod_{i=1}^L q_t\bigl(\rvx^{(i)} \mid \rvx_1^{(i)}\bigr).
\end{equation}

In principle, our method accommodates any path $(q_t)_{t\in[0, 1)}$. 
In practice, however, we require (a) the ability to easily sample from $q_t$ for all $t$, and (b) well-behaved noise distributions as described in Section~\ref{sec:noising}, in particular satisfying the overlap condition of Proposition~\ref{prop:conv}.
For instance, Dirichlet distributions (or mixtures thereof) represent a suitable choice, e.g., $q_t(\rvx^{(i)} \mid \rvx_1^{(i)}) = \mathrm{Dir}(\rvx^{(i)}; \vone + \alpha_t\rvx_1^{(i)})$, where $\alpha_t \in \sR_{\geq0}$ is a non-decreasing noise schedule with $\alpha_0=0$ and $\lim_{t\to 1}\alpha_t = \infty$. 
We further discuss the choice of the probability path in the context of graph generation in Appendix~\ref{ap:path_choice} and leave the exploration of alternative probability paths for future work.

\paragraph{Independence across time}
Unlike standard diffusion models such as DDPM \citep{DDPM}, DDIM \citep{DDIM}, or D3PM \citep{discdiff_austin}, our model does not define direct dependencies between $\rvx_t$ and $\rvx_{t+h}$. 
Instead, we assume conditional independence of intermediate noisy states given the clean input, as formalized in the following assumption: 
\begin{assumption}\label{ass:conditional_indep}
For any $s \neq t$, the noising process satisfies the following conditional-independence property:
\begin{equation}\label{eq:conditional_indep}
q_{t}(\rvx_t \mid \rvx_s, \rvx_1) = q_t(\rvx_t \mid \rvx_1).
\end{equation}
\end{assumption}
Because training typically draws a single noisy sample at a randomly chosen time $t$, this assumption has no practical effect on the training objective. 
It does, however, have major implications for the denoising process.

\subsection{Denoising}\label{sec:denoising}

Let $P_{1\mid t}(\rvx_{1} \mid \rvx_{t})$ denote the posterior distribution of the clean instance $\rvx_1 \in \gX$ given a noisy instance $\rvx_t$ at time $t$. We denote by $p_{t+h}(\rvx_{t+h} \mid \rvx_{t})$ a single denoising transition and obtain the following result.

\begin{proposition}\label{prop:denoising}
Given $P_{1\mid t}(\rvx_{1} \mid \rvx_{t})$ and Assumption~\ref{ass:conditional_indep}, the one-step denoising kernel satisfies
\begin{equation}\label{eq:denoising}
\begin{aligned}
    p_{t+h}(\rvx_{t+h} \mid \rvx_{t}) &=
    \sum_{\rvx_{1} \in \gX} q_{t+h}(\rvx_{t+h} \mid \rvx_{1})\, P_{1\mid t}(\rvx_{1} \mid \rvx_{t}) \\
    &= \mathbb{E}_{\rvx_{1} \sim P_{1\mid t}(\cdot \mid \rvx_{t})} \left[q_{t+h}(\rvx_{t+h} \mid \rvx_{1}) \right].
\end{aligned}
\end{equation}
\end{proposition}

Thus, each denoising step can be interpreted as (i) sampling a clean instance $\rvx_{1}$ from the posterior $P_{1\mid t}(\rvx_{1} \mid \rvx_{t})$, followed by (ii) re-noising this sample via $q_{t+h}(\cdot \mid \rvx_{1})$. Equivalently, we sample from the mixture of noising distributions weighted by the posterior. In practice, the posterior is approximated by a product of per-dimension distributions (Section~\ref{sec:param}), so both steps are performed independently for each dimension.
We emphasize that the simplicity of these formulations relies on the fact that 
$q_{t+h}(\rvx_{t+h} \mid \rvx_{1})$ remains on the simplex.
At inference time, we fix the number of function evaluations (NFE) to $T$ and use a uniform step size $h = 1/T$. The full reverse procedure consists of iterating the transition in Equation~\ref{eq:denoising} $T$ times.

\subsection{Convergence and Corrector Sampling}

Fixing the time index in Equation~\ref{eq:denoising} (i.e., setting $h = 0$) yields a Markov kernel at time $t$, whose convergence is characterized by the following proposition.

\begin{proposition}\label{prop:conv}
Let $t \in [0,1)$ and let $P_{1\mid t}$ be the true posterior. Assume the \emph{overlap condition}
\begin{equation}\label{eq:overlap}
\varepsilon_t \;:=\; \int_{\sS_K^L} \min_{\rvx_1 \in \gX} q_t(\rvx \mid \rvx_1)\, \mathrm{d}\rvx \;>\; 0.
\end{equation}
Then, for any initial distribution, the Markov chain with kernel
$
p_{t'}(\rvx_t' \mid \rvx_t) 
= \sum_{\rvx_1 \in \gX} P_{1\mid t}(\rvx_1 \mid \rvx_t)\, q_t(\rvx_t' \mid \rvx_1)
$
converges in total variation to its unique stationary distribution
\begin{equation}
\pi_t(\rvx_t) = \sum_{\rvx_1 \in \gX} p(\rvx_1)\, q_t(\rvx_t \mid \rvx_1).
\end{equation}
\end{proposition}
The overlap condition requires the noisy distributions $q_t(\cdot \mid \rvx_1)$ to share a common component of total mass $\varepsilon_t$. Since, for every $k$, the density of $\mathrm{Dir}(\vone + \alpha_t \ve_k)$ vanishes only on the boundary of the simplex, and the path factorizes across dimensions, we have $\varepsilon_t > 0$ for all $t < 1$.

Applying this kernel at a fixed time during sampling therefore yields a simple \emph{corrector} sampler, analogous in spirit to the corrector step of \citet{discrete_flow_matching_gat} for discrete flow matching. Thus, iterating denoising and corrector steps to stationarity recovers samples from $p(\rvx_1)$. 
In practice, the true posterior $P_{1\mid t}(\rvx_1 \mid \rvx_t)$ is not available and must be approximated by a parameterized model $P_{1\mid t}^{\theta}(\rvx_1 \mid \rvx_t)$. The resulting approximation error induces an error in the denoising transition, which is bounded as follows:

\begin{proposition}\label{prop:error_bound}
The KL divergence between the true and approximate one-step denoising distributions is upper-bounded by the KL divergence between the true and approximate posterior distributions:
\begin{equation}
D_{\mathrm{KL}}\left(p_{t+h}(\cdot \mid \rvx_t) \,\|\, p_{t+h}^{\theta}(\cdot \mid \rvx_t)\right)
\leq
D_{\mathrm{KL}}\left(P_{1\mid t}(\cdot \mid \rvx_t) \,\|\, P_{1\mid t}^{\theta}(\cdot \mid \rvx_t)\right).
\end{equation}
\end{proposition}
This proposition guarantees that the posterior approximation error is not amplified in the induced one-step denoising transition.

\subsection{Differences from and Advantages over Diffusion and Flow Matching}\label{sec:advantages}

While our procedure resembles standard diffusion and flow matching, there is a key distinction. In classical reverse processes, one typically enforces
$
\lim_{h \to 0} p_{t+h}(\rvx \mid \rvx_t) = \delta(\rvx - \rvx_t).
$
By Assumption~\ref{ass:conditional_indep}, our model does not impose this constraint. 
Intuitively, diffusion and flow matching update the sample by interpolating between the current noisy state and the denoiser's prediction, whereas our method directly resamples a less noisy state conditioned on the denoiser's prediction.

Unlike flow matching, our method contracts the error accumulated during sampling, as formalized below. We denote by $Q$ the Markov operator associated with the denoising kernel of Equation~\ref{eq:denoising} under the true posterior, acting on a distribution $\mu$ as $\mu Q := \int p_{t+h}(\cdot \mid \rvx_t)\, \mu(\rvx_t)\, \mathrm{d}\rvx_t$.

\begin{proposition}\label{prop:error_contraction}
Let $t + h < 1$ with $\varepsilon_{t+h} > 0$ (Equation~\ref{eq:overlap}), and let $\tilde\pi_t$ be any distribution on $\sS_K^L$. Let $\Phi$ be the flow map from $t$ to $t+h$ of an exact flow-matching ODE with marginals $(\pi_s)_s$, assumed to be a diffeomorphism. Then
\begin{align}
\|\tilde\pi_t Q - \pi_{t+h}\|_{\mathrm{TV}} &\le (1 - \varepsilon_{t+h})\, \|\tilde\pi_t - \pi_t\|_{\mathrm{TV}}, \label{eq:unsid-contraction}\\
\|\Phi_\# \tilde\pi_t - \pi_{t+h}\|_{\mathrm{TV}} &= \|\tilde\pi_t - \pi_t\|_{\mathrm{TV}}. \label{eq:fm-preservation}
\end{align}
\end{proposition}
Here, $\tilde\pi_t$ represents the law of the sampler at time $t$, including the error accumulated so far: a flow-matching step transports this error unchanged, whereas a UNSID step shrinks it by a factor of at least $1 - \varepsilon_{t+h}$.

Moreover, in diffusion and flow matching, if $\rvx_t$ lies in a region where the learned denoiser $P^\theta_{1\mid t}$ poorly approximates the true posterior $P_{1\mid t}$, or in a region highly sensitive to small perturbations (e.g., near equilibrium points), then the next iterate $\rvx_{t+h}$ remains close to $\rvx_t$, causing approximation errors to compound over time. In contrast, our method resamples from $P^\theta_{1\mid t}$ at each step, so $\rvx_{t+h}$ need not stay in the same local region. This reduces error compounding by allowing the trajectory to escape regions that are poorly approximated and/or sensitive to small perturbations.
Our experiments demonstrate that UNSID consistently outperforms diffusion-based models and Dirichlet Flow Matching by a large margin. 

In addition, our formulation is a substantial simplification compared with standard denoising methods such as diffusion and flow matching. Compared to diffusion, Assumption~\ref{ass:conditional_indep} yields the simplification $q_{t+h}(\rvx_{t+h}\mid \rvx_{1}, \rvx_{t}) = q_{t+h}(\rvx_{t+h}\mid \rvx_{1})$. Compared to flow matching, it removes the need to construct a vector field that realizes the chosen probability path and to evaluate its derivatives at every denoising step, operations that are computationally expensive (see the sampling-time comparison in Appendix~\ref{ap:time}). In contrast, our method is both simple and computationally efficient.
We therefore regard this simplification as an important contribution.

\subsection{Parameterization and Learning}\label{sec:param}

As in discrete diffusion, the posterior $P_{1\mid t}(\rvx_{1} \mid \rvx_{t})$ is intractable. We approximate it by a product of per-dimension categorical distributions, $P^\theta_{1\mid t}(\rvx_1 \mid \rvx_t) = \prod_{i=1}^L \mathrm{Cat}(\rvx_1^{(i)}; \bm{\omega}^{(i)})$, with $\bm{\omega}^{(1:L)} = \mathrm{softmax}\bigl(f_\theta(\rvx_t, t)\bigr)$, where the neural network $f_\theta$, referred to as the \emph{denoiser}, outputs element-wise logits. Our training objective matches the $x$-prediction of discrete diffusion and discrete flow matching methods, so any of their criteria can be adopted. To respect the symmetries of the graph, we instantiate $f_{\theta}$ as a GNN that is equivariant with respect to node permutations.

On graphs, in place of generic vectors $\rvx^{(i)} \in \{0,1\}^{K}$, we write $\rvv^{(i)} \in \{0,1\}^{d_v}$ for vectors encoding the categorical attributes of node $i$ ($d_v$ categories), and $\rva^{(i,j)} \in \{0,1\}^{d_a+1}$ for those encoding the categorical attributes of the pair $(i,j)$ ($d_a$ categories), with an additional category accounting for the absence of an edge.

Following \citet{digress}, we train our model by minimizing the following weighted negative log-likelihood:
\begin{equation}\label{eq:loss}
\begin{split}
 \gL_\theta = \mathbb{E} \biggl[ & \frac{\gamma}{n} \sum_i
 \left[-\log(P^\theta_{1\mid t}(\rvv_1^{(i)} \mid \gG_{t})) \right] \\
 & + \frac{1-\gamma}{n(n-1)}\sum_{i \neq j}
 \left[-\log(P^\theta_{1\mid t}(\rva_1^{(i, j)} \mid \gG_{t}))\right] \biggr],
\end{split}
\end{equation}
where the expectation is taken over $t \sim \gU(0, 1)$, $\gG_1 \sim p_{\text{data}}$, and $\gG_{t} \sim q_t(\gG_t \mid \gG_1)$, $n$ is the number of nodes, and $\gamma$ is a weighting factor between nodes and edges.

\subsection{Guidance}

While high-quality unconditional generation is a prerequisite, conditioning on graph-level properties is necessary for many downstream tasks. In drug discovery, for example, one often seeks molecules that are both synthetically accessible and highly active against specified targets. Denoising-based models typically steer samples via either \emph{classifier guidance} \citep{classifier_guidance} or \emph{classifier-free guidance} \citep{classifier_free_guidance}. Our framework supports both through direct adaptations of established techniques. Due to space limitations, we provide implementation details in Appendix~\ref{ap:guidance}. We demonstrate the effectiveness of our approach for property-conditioned molecular generation in Section~\ref{sec:eval}.
\section{Evaluation and Results}\label{sec:eval}

We evaluate our model on both molecular and unattributed graph datasets. 
For molecular data, we adopt two standard benchmark datasets: \texttt{QM9} and \texttt{ZINC250K}. For \texttt{QM9}, we use the more challenging version \emph{with explicit hydrogens} (\texttt{QM9H}), since performance on the hydrogen-free version is close to saturation. 
The \texttt{QM9H} dataset contains 133,885 molecules with up to 29 atoms of 5 types. The \texttt{ZINC250K} dataset contains 249,455 molecular graphs with up to 38 heavy atoms of 9 types.
For unattributed graphs, we evaluate on the synthetic \texttt{Planar} and Stochastic Block Model (\texttt{SBM}) datasets, comprising graphs of exactly 64 nodes and of up to 187 nodes, respectively, as well as on the real-world \texttt{Enzymes} dataset, with graphs of up to 125 nodes.
Visualizations of generated molecules and graphs are available in Appendix~\ref{ap:visual}.
Our (anonymous) code is available at \url{https://github.com/yoboget/UNSID}.

\subsection{Baselines}
We compare against two categories of baselines: (i) published models representing different denoising approaches, and (ii) our own implementations for controlled comparisons. Due to space limitations, we describe the baselines in Appendix~\ref{ap:baselines}. In general, we report results from each baseline's original paper, except on \texttt{Planar} and \texttt{SBM}, where we reran the experiments for baselines that reported results from a \emph{single} sampling run (see Appendix~\ref{ap:baselines} for details and further justification). 

To isolate the effect of our sampling formulation, we train three models with identical architectures and training hyperparameters for each experiment: a discrete denoiser, a simplex-based denoiser, and a continuous denoiser. The only notable difference is that the discrete denoiser includes the extra input features used by \textsc{DiGress}. For the discrete and the simplex-based denoisers, we evaluate two sampling procedures each. For the discrete denoiser, we use standard discrete diffusion (\textsc{DiscDiff}, as in \textsc{DiGress}) and its non-Markovian variant (\textsc{NM-DD}, as in \textsc{SID}). For the simplex-based denoiser, we implement the Dirichlet flow-matching sampler of \citet{DirichletFlowMating} (\textsc{DiriFM}), as well as our unrestrained simplex denoising (\textsc{UNSID}) sampler. The continuous denoiser is identical to the simplex-based denoiser, but its inputs lie in $\sR^{K\times L}$ instead of $\sS_K^L$. We refer to it as non-Markovian continuous diffusion (\textsc{NM-CD} (see Appendix~\ref{ap:continuous_diffusion} for details). Experimental and implementation details are provided in Appendix~\ref{ap:eval}.

\subsection{Results}
We report means over five independent sampling runs. Appendices~\ref{ap:eval} and~\ref{ap:results} provide the evaluation protocol, additional metrics, and standard deviations. We highlight in bold all results that fall within the training-set error margin. If none do, we bold the best result.

\paragraph{Molecule Generation} We evaluate generative performance on \texttt{QM9H} (Table~\ref{tab:qm9H} in Appendix~\ref{ap:eval}) and \texttt{ZINC250K} (Table~\ref{tab:zinc}) by measuring distributional similarity in chemical space via the Fr\'echet ChemNet Distance (FCD) \citep{fcd}, in subgraph structures via the neighborhood subgraph pairwise distance kernel MMD (NSPDK) \citep{nspdk}, reported $\times 10^3$), and chemical validity without correction. 
Our \textsc{UNSID} model surpasses all baselines, often by a large margin, on all metrics. In particular, it generates very few invalid molecules.

\begin{minipage}{\textwidth}
  \begin{minipage}[b]{0.49\textwidth}
    \centering
    \begin{sc}
    \begin{small}
    \captionof{table}{Results on \texttt{ZINC250K}.}
    \label{tab:zinc}
    \begin{tabular}{lccc}
    \toprule
          & Valid (\%)$\uparrow$ & FCD$\downarrow$ & NSPDK$\downarrow$ \\
    \midrule
        Train. &
        $100.00\phantom{0}$ & $1.13$ & $0.10$ \\
        DiGress &
        $94.99$ & $3.48$ & $2.1\phantom{0}$ \\
        GruM &
        $98.65$ & $2.26$ & $1.5\phantom{0}$ \\
        GraBFN &
        $99.22$ & $2.12$ & $1.3\phantom{0}$ \\
        SID &
        $99.50$ & $2.06$ & $2.01$ \\
    \midrule
        DiscDiff &
        $74.17$ & $4.78$ & $4.08$  \\
        NM-DD &
        $99.92$ & $2.65$ & $3.48$ \\
        NM-CD &
        $57.96$ & $9.51$ & $13.65\phantom{0}$ \\
        DiriFM &
        $97.32$ & $2.79$ & $1.92$ \\
    \midrule
        UNSID &
        $\bm{99.98}$ & $\bm{1.79}$ & $\bm{1.00}$ \\
    \bottomrule
    \end{tabular}
    \end{small}
    \end{sc}       
  \end{minipage}
  \hfill
  \begin{minipage}[b]{0.49\textwidth}
    \centering
    \includegraphics[width=0.8\linewidth]{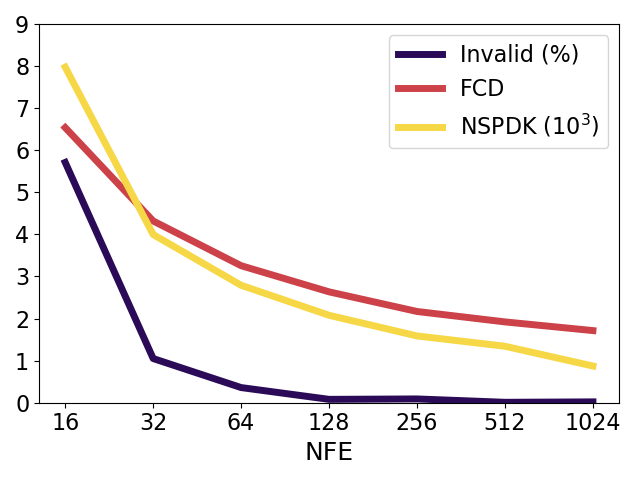}
    \captionof{figure}{\texttt{ZINC250K} results for various NFE.}
    \label{fig:nfe}
  \end{minipage}
\end{minipage}

\paragraph{Unattributed Graph Generation}
For unattributed graphs, we measure distributional similarity via Maximum Mean Discrepancy (MMD) over degree distributions, clustering coefficients, orbit counts, and spectral densities. For \texttt{Planar} and \texttt{SBM}, we follow the procedure of \citet{spectre}: we report MMDs $\times 10^3$, as well as validity, i.e., planarity for \texttt{Planar} and statistical consistency with the true stochastic block model for \texttt{SBM}. On \texttt{SBM}, we also report novelty, i.e., the fraction of unique generated graphs that are not isomorphic to those in the training set. On \texttt{Enzymes}, we follow \citet{gdss}. In particular, we compute MMDs using the Earth Mover's Distance rather than the total variation distance.

On \texttt{Planar} (Table~\ref{tab:planar_sbm}), our model achieves a perfect validity rate and results comparable to the best baselines on the other metrics. 
On \texttt{SBM}, we reach the highest validity after \textsc{DeFoG}, whose results, however, strongly suggest overfitting and dataset memorization and should be taken with caution (see Appendix~\ref{ap:sbm}). On the remaining metrics, our results are comparable to those of the strongest baselines.
On \texttt{Enzymes} (Table~\ref{tab:enzymes}), our model achieves the best orbit MMD and the best the mean ratio between the model's MMDs to the training-set MMDs, while remaining competitive on the other metrics.

\begin{table}[ht]
    \setlength{\tabcolsep}{2pt}
    \begin{scriptsize}
    \begin{center}
    \begin{sc}
    \caption{Results on \texttt{Planar} and \texttt{SBM}.}
    \begin{tabular}{lcccccc cccccc}
    \toprule
        & \multicolumn{5}{c}{\textbf{Planar}} & & \multicolumn{6}{c}{\textbf{SBM}} \\
        \cmidrule(lr){2-6} \cmidrule(lr){8-13}
        & Valid$\uparrow$ & Deg.$\downarrow$ & Clust.$\downarrow$ & Orbit$\downarrow$ & Spect.$\downarrow$
        & & Valid$\uparrow$ & Deg.$\downarrow$ & Clust.$\downarrow$ & Orbit$\downarrow$ & Spect.$\downarrow$ & Novel $\uparrow$ \\
    \midrule
        Train. &
        $100.0\phantom{0}$ & $\phantom{0}0.25$ & $36.6$ & $0.8$ & $6.6$ & &
        $93.5$ & $\phantom{0}1.57$ & $50.1$ & $37.0$ & $4.6$ & 100 \\
        DiGress &
        $45.5$ & $\phantom{0}0.71$ & $45.4$ & $\phantom{0}1.31$ & $8.4$ & &
        $51.0$ & $\phantom{0}\bm{1.28}$ & $51.5$ & $\bm{39.6}$ & $\bm{5.0}$& 100 \\
        GruM &
        $91.0$ & $\phantom{0}\bm{0.38}$ & $40.7$ & $\phantom{0}6.42$ & $\bm{7.9}$ & &
        $67.5$ & $\phantom{0}2.20$ & $\bm{49.9}$ & $\bm{40.4}$ & $\bm{5.1}$ & 100\\
        SID &
        $91.3$ & $5.9$ & $163.4\phantom{0}$ & $19.1\phantom{0}$ & $\bm{7.6}$ & &
        $63.5$ & $11.5\phantom{0}$ & $\bm{51.4}$ & $123.1\phantom{0}$ & $5.9$ & 100\\
        DeFoG &
        $99.5$ & $0.5$ & $50.1$ & $0.6$ & $\bm{7.2}$ &
         & $\bm{90.0}$ & \phantom{*}$0.6$* &  $51.7$ & $\phantom{0}55.6$ & $\bm{5.4}$ & \phantom{0}$90$ \\
    \midrule
        DiscDiff &
        $0.0$ & $56.1\phantom{0}$ & $294.0\phantom{0}$ & $1410.0\phantom{000}$ & $85.1\phantom{0}$ & &
        $\phantom{0}0.0$ & $\phantom{0}\bm{1.82}$ & $86.4$ & $125.7\phantom{0}$ & $11.3\phantom{0}$ & 100\\
        NM-DD &
        $98.0$ & $14.1\phantom{0}$ & $363.0\phantom{0}$ & $27.2\phantom{0}$ & $\bm{6.9}$ & &
        $60.5$ & $\phantom{0}4.38$ & $\bm{50.9}$ & $52.9$ & $5.7$ & 100 \\
        NM-CD &
        $\phantom{0}0.0$ & $1.2$ & $226.3\phantom{0}$ & $45.8\phantom{0}$ & $22.0\phantom{0}$ & &
        $57.5$ & $\phantom{0}\bm{1.89}$ & $\bm{50.3}$ & $\bm{41.1}$ & $\bm{5.4}$ & 100 \\
        DiriFM &
        $\phantom{0}0.0$ & $6.4$ & $196.6\phantom{0}$ & $45.0\phantom{0}$ & $21.8\phantom{0}$ & &
        $46.0$ & $\phantom{0}4.26$ & $53.0$ & $53.0$ & $\bm{5.0}$ & 100 \\
    \midrule
        UNSID &
        $\bm{100.0}\phantom{0}$ & $\phantom{0}\bm{0.36}$ & $\bm{39.9}$ & $\phantom{0}\bm{0.78}$ & $\bm{7.1}$ & &
        $78.5$ & $\phantom{0}\bm{1.74}$ & $\bm{49.9}$ & $52.1$ & $5.9$ & 100\\
    \bottomrule
    \end{tabular}
    \label{tab:planar_sbm}
    \end{sc}
    \end{center}
    * Significantly below the reference value. 
    \end{scriptsize} 
\end{table}

\subsection{Ablations and Efficiency}\label{sec:ablation}
 
We study the impact of key design choices, namely the unrestrained denoising formulation, the NFE, the noise schedule, and the initialization of the sampling noise, and we assess sampling efficiency.
First, \textsc{DiriFM} and \textsc{UNSID} use the same denoiser. The only difference lies in the sampling strategy. On almost all experiments and metrics, our unrestrained sampler outperforms Dirichlet flow matching, often by a large margin, demonstrating the empirical superiority of our sampling method.
Second, we ablate NFE on \texttt{ZINC250K} (Figure~\ref{fig:nfe} and Table~\ref{tab:nfe} in Appendix~\ref{ap:eval}). Our model attains state-of-the-art validity in as few as 64 steps, with competitive FCD and NSPDK. As expected, all metrics improve steadily with larger NFE.
Third, Appendix~\ref{ap:time} shows that our model is highly efficient at sampling: it is the fastest among all compared methods and achieves these results with fewer parameters than the strongest baselines.
Fourth, Appendix~\ref{ap:sched} demonstrates that Voronoi probabilities provide an effective tool for calibrating our noise schedule, and that our model remains robust under substantial variations of the schedule hyperparameter $a$.
Finally, we compare two noise initializations: the uniform prior $q_{0} = \mathrm{Dir}(\vone)$ and the marginal mixture $q_{0}^{\mathrm{marg}}$ with $\kappa = 2$ (see Appendix~\ref{ap:path_choice} for definitions). Marginal initialization yields a small but consistent improvement on all metrics (see Figure~\ref{fig:noise_compar} in Appendix~\ref{ap:eval}).

\begin{wraptable}{r}{8cm}
    \setlength{\tabcolsep}{2pt}
    \begin{sc}
    \begin{center}
    \begin{scriptsize}
    \centering
    \caption{Results on \texttt{Enzymes}.}
    \begin{tabular}{lccccc}
    \toprule
         & Deg.$\downarrow$ & Clust.$\downarrow$ &
         Orbit$\downarrow$ & Spect.$\downarrow$ &  Ratio$\downarrow$ \\
    \midrule    
       Train.  & $ 0.003$ & $0.019$ & $0.001$ & $0.0189$ & $1$ \\
       
       GraphVAE & $1.369$ & $0.629$ & $0.191$ & - & - \\
       GraphRNN & $0.017$ & $0.062$ & $0.046$ & - & - \\
       GDSS     & $0.026$ & $0.061$ & $0.009$ & - & - \\

    \midrule
        DiscDiff & $0.212$ & $0.631$ & $0.515$ & $0.0196$ & $155.0$\\
        NM-CD    & $0.084$ & $0.308$ & $0.125$ & $0.0202$ & $42.6$\\
        NM-DD   &  $0.014$ & $\bm{0.040}$ & $0.010$ & $0.0174$  & $4.3$ \\
        DiriFM  & $ \bm{0.007}$ & $0.101$ & $0.006$ & $0.0198$ & $3.7$  \\
        UNSID  & $ 0.010$ & $0.052$ & $\bm{0.003}$ & $\bm{0.0195}$ & $\bm{2.3}$  \\

    \bottomrule
    \end{tabular}\label{tab:enzymes}
    \vspace{-0.2cm}
    \end{scriptsize}
    \end{center}
    \end{sc}
\end{wraptable}

\subsection{Guidance}

We evaluate conditional generation toward a target molecular property using the Quantitative Estimate of Drug-likeness (QED) score, as implemented in \texttt{RDKit}. On \texttt{QM9H}, we apply classifier guidance to steer samples toward prescribed standardized QED values. Concretely, we draw 1,000 target values from the test set, generate 1,000 molecules conditioned on these targets, and compute the QED of each generated molecule with \texttt{RDKit}. We quantify guidance effectiveness by the mean absolute error (MAE) between the target and generated standardized QED values.
Results are reported in Table~\ref{tab:guidance} in Appendix~\ref{ap:eval}. Classifier guidance effectively steers generation toward the prescribed values, reducing the MAE from 1.06 (unconditional) to 0.42, while preserving the model's ability to generate valid molecules.
\section{Conclusion}

We introduced \emph{Unrestrained Simplex Denoising} (UNSID), a new denoising paradigm for categorical data that operates directly on the probability simplex and relaxes the Markovian constraints commonly imposed by diffusion and flow-matching methods. Theoretically, we established convergence and error-contraction guarantees and introduced Voronoi probabilities as a tool for analyzing and calibrating noising paths on the simplex. On molecular and unattributed graph benchmarks, UNSID matches or surpasses strong discrete diffusion and flow-matching baselines, reaches state-of-the-art validity with few denoising steps, and enables effective property-conditioned generation. Ablations demonstrate the benefits of the non-Markovian formulation, a favorable NFE--quality trade-off, and the utility of marginal-based initialization on the simplex. 

\paragraph{Limitations} Although UNSID could in principle extend to other forms of discrete data, we validate it only on graph generation in the present work. Its applicability to other discrete modalities, in particular sequences such as text, proteins, or DNA, remains to be established empirically and is left for future research.

\paragraph{Future work} This limitation also points to opportunities. As a new class of generative models, UNSID opens promising directions, such as exploring alternative probability paths and integrating hybrid discrete--continuous variables.

\begin{ack}
We acknowledge the financial support of the Swiss National Science Foundation within the LegoMol project (grant no. 207428). The computations were performed at the University of Geneva on ”Baobab” and ”Bamboo”
HPC clusters.
\end{ack}

\newpage

\bibliography{main}


\appendix

\section{Proofs}\label{ap:proofs}

\subsection{Preliminaries}

We recall the gamma function,
\begin{equation}
    \Gamma(a) = \int_0^\infty s^{a-1} e^{-s} \, \mathrm{d}s, \quad a \in \sR_{>0},
\end{equation}
the multivariate beta function,
\begin{equation}\label{eq:beta}
    \mathrm{B}(\bm{\alpha}) = \frac{\prod_{i=1}^K \Gamma(\alpha_i)}{\Gamma\bigl(\sum_{i=1}^K \alpha_i\bigr)},
\end{equation}
the density of the Dirichlet distribution at a point $\vx \in \sS_K$,
\begin{equation}
    \mathrm{Dir}(\vx; \bm{\alpha}) = \frac{1}{\mathrm{B}(\bm{\alpha})} \prod_{i=1}^{K} x_i^{\alpha_i - 1},
\end{equation}
and the density of the $\mathrm{Gamma}(a, b)$ distribution,
\begin{equation}\label{eq:gamma}
    f(x; a, b) = \frac{b^a}{\Gamma(a)} x^{a - 1} e^{-b x}, \quad x > 0,\; a > 0,\; b > 0.
\end{equation}

\subsection{Proposition~\ref{prop:voronoi}: Voronoi Probability}

Let $\rvx \sim \mathrm{Dir}(\vone + \alpha_t \ve_i)$ denote the noisy sample $\rvx_t$ given $\rvx_1 = \ve_i$, and let $a := \alpha_t + 1$; i.e., all Dirichlet parameters equal one except the $i$th, which equals $a$. We show that
\begin{equation}
    P_{v_i}(\rvx_t \in \sV_i \mid \rvx_1 = \ve_i) = \sum_{k=0}^{K-1} \frac{(-1)^k \binom{K-1}{k}}{(k+1)^{a}}.
\end{equation}

Recall that the Voronoi region of vertex $i$ is defined as
\begin{equation}
    \sV_i = \{\vx \in \sS_K \mid d(\vx,\ve_i) \leq d(\vx,\ve_j) \; \forall j \neq i\},
\end{equation}
with $d(\cdot, \cdot)$ the Euclidean distance. Since $d(\vx, \ve_j)^2 = \|\vx\|^2 - 2x_j + 1$, we have $d(\vx,\ve_i) \leq d(\vx,\ve_j) \iff x_i \geq x_j$, hence
\begin{equation}
    \vx \in \sV_i \iff x_i = \max_{j \in [K]} x_j,
\end{equation}
so that
\begin{equation}\label{eq:p_maxj}
    P_{v_i}(\rvx_t \in \sV_i \mid \rvx_1 = \ve_i) = P\Bigl(\evx_i = \max_{j \in [K]} \evx_j\Bigr).
\end{equation}

It is well known that, for any $b > 0$, if $g_1, \dots, g_K$ are independent with $g_j \sim \mathrm{Gamma}(\alpha_j, b)$, then
\begin{equation}
    [\evx_1, \ldots, \evx_K]^\top \sim \mathrm{Dir}(\alpha_1, \dots, \alpha_K), \qquad \text{where } \evx_j = \frac{g_j}{\sum_{l=1}^{K} g_l}.
\end{equation}
Since all coordinates share the same positive denominator, the normalization preserves their ordering:
\begin{equation}\label{eq:max_x_max_g}
    g_i = \max_{j \in [K]} g_j \iff \evx_i = \max_{j \in [K]} \evx_j.
\end{equation}
In the following, we take $b = 1$, with $g_i \sim \mathrm{Gamma}(a, 1)$ and $g_j \sim \mathrm{Gamma}(1, 1)$ for all $j \neq i$, all independent.

From Equations~\ref{eq:p_maxj} and~\ref{eq:max_x_max_g}, we get
\begin{equation}
    P_{v_i}(\rvx_t \in \sV_i \mid \rvx_1 = \ve_i) = P\bigl(g_i \geq g_j \;\; \forall j \neq i\bigr).
\end{equation}
The cumulative distribution function of $\mathrm{Gamma}(1, 1)$, i.e., the exponential distribution with unit rate, is $F(s) = 1 - e^{-s}$ for $s \geq 0$. Conditioning on $g_i = s$ and using the independence of the $g_j$, we have
\begin{equation}
    P\bigl(g_j \leq s \;\; \forall j \neq i \bigm| g_i = s\bigr) = (1 - e^{-s})^{K-1}.
\end{equation}
Integrating against the density of $g_i$ (Equation~\ref{eq:gamma} with $b = 1$) yields
\begin{equation}\label{eq:integral}
    P_{v_i}(\rvx_t \in \sV_i \mid \rvx_1 = \ve_i) = \frac{1}{\Gamma(a)} \int_0^\infty s^{a-1} e^{-s} (1 - e^{-s})^{K-1} \, \mathrm{d}s.
\end{equation}

By the binomial theorem,
\begin{equation}
    (1-e^{-s})^{K-1} = \sum_{k=0}^{K-1} (-1)^k \binom{K-1}{k} e^{-ks}.
\end{equation}
Substituting into Equation~\ref{eq:integral} and exchanging the finite sum and the integral, we obtain
\begin{equation}
    P_{v_i}(\rvx_t \in \sV_i \mid \rvx_1 = \ve_i) = \frac{1}{\Gamma(a)} \sum_{k=0}^{K-1} (-1)^k \binom{K-1}{k} \int_0^\infty s^{a-1} e^{-(k+1)s} \, \mathrm{d}s.
\end{equation}
With the change of variables $u = (k+1)s$, i.e., $s = u/(k+1)$ and $\mathrm{d}s = \mathrm{d}u/(k+1)$,

\begin{align}
    P_{v_i}(\rvx_t \in \sV_i \mid \rvx_1 = \ve_i)
    &= \frac{1}{\Gamma(a)}\sum_{k=0}^{K-1}(-1)^k\binom{K-1}{k}\int_0^\infty \frac{u^{a-1}}{(k+1)^{a-1}}e^{-u}\frac{1}{k+1}du  \\
    &= \frac{1}{\Gamma(a)}\sum_{k=0}^{K-1} \frac{(-1)^k\binom{K-1}{k}}{(k+1)^a}\int_0^\infty u^{a-1}e^{-u}du \\
    &= \frac{1}{\Gamma(a)}\sum_{k=0}^{K-1} \frac{(-1)^k\binom{K-1}{k}}{(k+1)^a} \Gamma(a) \\
    &=\sum_{k=0}^{K-1} \frac{(-1)^k\binom{K-1}{k}}{(k+1)^a} 
\end{align}

Hence,
\begin{equation}
    P_{v_i}(\rvx_t \in \sV_i \mid \rvx_1 = \ve_i) = \sum_{k=0}^{K-1} \frac{(-1)^k \binom{K-1}{k}}{(k+1)^{a}},
\end{equation}
which concludes the proof.

\subsection{Proposition~\ref{prop:denoising}: Denoising Kernel}

Given $P_{1\mid t}(\rvx_{1} \mid \rvx_{t})$ and Assumption~\ref{ass:conditional_indep}, the one-step denoising kernel satisfies
\begin{equation}
\begin{aligned}
p_{t+h}(\rvx_{t+h} \mid \rvx_{t})
&= \sum_{\rvx_{1} \in \gX} q_{t+h}(\rvx_{t+h} \mid \rvx_{1})\, P_{1\mid t}(\rvx_{1} \mid \rvx_{t}) \\
&= \mathbb{E}_{\rvx_{1} \sim P_{1\mid t}(\cdot \mid \rvx_{t})} \left[q_{t+h}(\rvx_{t+h} \mid \rvx_{1}) \right].
\end{aligned}
\end{equation}

By the law of total probability, we have
\begin{equation}
p_{t+h}(\rvx_{t+h} \mid \rvx_{t}) = \sum_{\rvx_{1} \in \gX} p_{t+h}(\rvx_{t+h} \mid \rvx_{t}, \rvx_{1})\, P_{1\mid t}(\rvx_{1} \mid \rvx_{t}).
\end{equation}
The true denoising process is given by the conditionals of the joint law of the noising process; in particular, $p_{t+h}(\rvx_{t+h} \mid \rvx_t, \rvx_1) = q_{t+h}(\rvx_{t+h} \mid \rvx_t, \rvx_1)$. By Assumption~\ref{ass:conditional_indep}, applied at time $t+h$ with $s = t$, this equals $q_{t+h}(\rvx_{t+h} \mid \rvx_1)$, which yields the first equality. The second equality rewrites the sum as an expectation over $\rvx_1 \sim P_{1\mid t}(\cdot \mid \rvx_t)$.

\subsection{Proposition~\ref{prop:error_bound}: Approximation Error Bound}

The KL divergence between the true and approximate one-step denoising distributions is upper-bounded by the KL divergence between the true and approximate posterior distributions:
\begin{equation}
D_{\mathrm{KL}}\bigl(p_{t+h}(\cdot \mid \rvx_t) \,\|\, p^{\theta}_{t+h}(\cdot \mid \rvx_t)\bigr)
\leq
D_{\mathrm{KL}}\bigl(P_{1\mid t}(\cdot \mid \rvx_t) \,\|\, P^{\theta}_{1\mid t}(\cdot \mid \rvx_t)\bigr).
\end{equation}

Here, $p^{\theta}_{t+h}$ denotes the denoising kernel of Equation~\ref{eq:denoising} with $P_{1\mid t}$ replaced by $P^{\theta}_{1\mid t}$. We first define the joint laws on $\gX \times \sS_K^L$
\begin{equation}
\rho(\rvx_1, \rvx_{t+h} \mid \rvx_t) = P_{1\mid t}(\rvx_1 \mid \rvx_t)\, q_{t+h}(\rvx_{t+h} \mid \rvx_1)
\end{equation}
and
\begin{equation}
\rho^{\theta}(\rvx_1, \rvx_{t+h} \mid \rvx_t) = P^{\theta}_{1\mid t}(\rvx_1 \mid \rvx_t)\, q_{t+h}(\rvx_{t+h} \mid \rvx_1).
\end{equation}
By Proposition~\ref{prop:denoising}, their marginals in $\rvx_{t+h}$ are $p_{t+h}(\cdot \mid \rvx_t)$ and $p^{\theta}_{t+h}(\cdot \mid \rvx_t)$, respectively. Moreover, since both joint laws share the same kernel $q_{t+h}(\cdot \mid \rvx_1)$, it cancels in the log-ratio:
\begin{align}
D_{\mathrm{KL}}(\rho \,\|\, \rho^{\theta})
&= \sum_{\rvx_1 \in \gX} \int \rho(\rvx_1, \rvx_{t+h} \mid \rvx_t) \log \frac{\rho(\rvx_1, \rvx_{t+h} \mid \rvx_t)}{\rho^{\theta}(\rvx_1, \rvx_{t+h} \mid \rvx_t)} \, \mathrm{d}\rvx_{t+h} \\
&= \sum_{\rvx_1 \in \gX} \int P_{1\mid t}(\rvx_1 \mid \rvx_t)\, q_{t+h}(\rvx_{t+h} \mid \rvx_1) \log \frac{P_{1\mid t}(\rvx_1 \mid \rvx_t)}{P^{\theta}_{1\mid t}(\rvx_1 \mid \rvx_t)} \, \mathrm{d}\rvx_{t+h} \\
&= \sum_{\rvx_1 \in \gX} P_{1\mid t}(\rvx_1 \mid \rvx_t) \log \frac{P_{1\mid t}(\rvx_1 \mid \rvx_t)}{P^{\theta}_{1\mid t}(\rvx_1 \mid \rvx_t)} \int q_{t+h}(\rvx_{t+h} \mid \rvx_1) \, \mathrm{d}\rvx_{t+h} \\
&= D_{\mathrm{KL}}\bigl(P_{1\mid t}(\cdot \mid \rvx_t) \,\|\, P^{\theta}_{1\mid t}(\cdot \mid \rvx_t)\bigr),
\end{align}
since $q_{t+h}(\cdot \mid \rvx_1)$ integrates to $1$ for every $\rvx_1$.

The log-sum inequality yields, for each fixed $\rvx_{t+h}$,
\begin{equation}
\begin{split}
&\Bigl( \sum_{\rvx_1 \in \gX} \rho(\rvx_1, \rvx_{t+h} \mid \rvx_t) \Bigr)
\log \frac{\sum_{\rvx_1 \in \gX} \rho(\rvx_1, \rvx_{t+h} \mid \rvx_t)}{\sum_{\rvx_1 \in \gX} \rho^{\theta}(\rvx_1, \rvx_{t+h} \mid \rvx_t)} \\
&\qquad \leq \sum_{\rvx_1 \in \gX} \rho(\rvx_1, \rvx_{t+h} \mid \rvx_t) \log \frac{\rho(\rvx_1, \rvx_{t+h} \mid \rvx_t)}{\rho^{\theta}(\rvx_1, \rvx_{t+h} \mid \rvx_t)}.
\end{split}
\end{equation}
Using the definition of the marginals, this becomes
\begin{equation}
p_{t+h}(\rvx_{t+h} \mid \rvx_t) \log \frac{p_{t+h}(\rvx_{t+h} \mid \rvx_t)}{p^{\theta}_{t+h}(\rvx_{t+h} \mid \rvx_t)}
\leq
\sum_{\rvx_1 \in \gX} \rho(\rvx_1, \rvx_{t+h} \mid \rvx_t) \log \frac{\rho(\rvx_1, \rvx_{t+h} \mid \rvx_t)}{\rho^{\theta}(\rvx_1, \rvx_{t+h} \mid \rvx_t)}.
\end{equation}
Integrating over $\rvx_{t+h}$ gives
\begin{equation}
D_{\mathrm{KL}}\bigl(p_{t+h}(\cdot \mid \rvx_t) \,\|\, p^{\theta}_{t+h}(\cdot \mid \rvx_t)\bigr)
\leq D_{\mathrm{KL}}(\rho \,\|\, \rho^{\theta}).
\end{equation}
Combining this with the previous identity yields
\begin{equation*}
D_{\mathrm{KL}}\bigl(p_{t+h}(\cdot \mid \rvx_t) \,\|\, p^{\theta}_{t+h}(\cdot \mid \rvx_t)\bigr)
\leq
D_{\mathrm{KL}}\bigl(P_{1\mid t}(\cdot \mid \rvx_t) \,\|\, P^{\theta}_{1\mid t}(\cdot \mid \rvx_t)\bigr).
\end{equation*}

\subsection{Proposition~\ref{prop:conv}: Convergence}

Fix $t \in [0,1)$. Throughout this proof, we write $q := q_t$, $\varepsilon := \varepsilon_t$, and $\pi := \pi_t$, where $\pi(\rvx) = \sum_{\rvx_1 \in \gX} p(\rvx_1)\, q(\rvx \mid \rvx_1)$ is the marginal of the noising process at time $t$. Assume:
\begin{enumerate}
    \item[\textup{(A1)}] \textbf{(Overlapping noising densities)} The noising densities share a positive amount of probability mass:
    \begin{equation}
        \varepsilon \;:=\; \int_{\sS_K^L} \min_{\rvx_1 \in \gX}\, q(\rvx \mid \rvx_1)\, \mathrm{d}\rvx \;>\; 0.
    \end{equation}
    \item[\textup{(A2)}] \textbf{(Exact posterior)} $P_{1\mid t}$ is the Bayes posterior of the noising process: for every $\rvx_1 \in \gX$ and every $\rvx_t$ with $\pi(\rvx_t) > 0$,
    \begin{equation}
    \label{eq:exact-posterior}
        P_{1\mid t}(\rvx_1 \mid \rvx_t)
        \;=\;
        \frac{p(\rvx_1)\, q(\rvx_t \mid \rvx_1)}{\pi(\rvx_t)}.
    \end{equation}
    Where $\pi(\rvx_t) = 0$, $P_{1\mid t}(\cdot \mid \rvx_t)$ may be any distribution on $\gX$.
\end{enumerate}
Define the Markov kernel on $\sS_K^L$ that denoises $\rvx_t$ and then renoises the prediction at the same noise level $t$, independently of $\rvx_t$:
\begin{equation}
\label{eq:kernel-def}
    p_{t'\mid t}(\rvx_t' \mid \rvx_t)
    \;=\;
    \sum_{\rvx_1 \in \gX} P_{1\mid t}(\rvx_1 \mid \rvx_t)\, q(\rvx_t' \mid \rvx_1),
\end{equation}
and let $Q_t$ denote the associated Markov operator, acting on a distribution $\mu$ as $(\mu Q_t)(\rvx_t') := \int \mu(\rvx_t)\, p_{t'\mid t}(\rvx_t' \mid \rvx_t)\, \mathrm{d}\rvx_t$. Then $\pi$ is the unique stationary distribution of $p_{t'\mid t}$, and for every initial distribution $\mu_0$ on $\sS_K^L$,\footnote{We use the convention $\|\mu - \mu'\|_{\mathrm{TV}} = \sup_A |\mu(A) - \mu'(A)|$, which lies in $[0,1]$.}
\begin{equation}
\label{eq:geometric-rate}
    \bigl\| \mu_0 Q_t^n - \pi \bigr\|_{\mathrm{TV}}
    \;\le\;
    (1 - \varepsilon)^n,
    \qquad n \ge 0.
\end{equation}
In particular, the law of the iterates converges to $\pi$ in total variation from any initialization, at a geometric rate.

We (1) check that Equation~\ref{eq:kernel-def} is a valid Markov kernel, (2) show that $\pi$ is stationary, (3) show that every transition lies above a common floor of mass $\varepsilon$, and (4) deduce geometric convergence and uniqueness.

\medskip
\noindent\textbf{1. $p_{t'\mid t}$ is a valid Markov kernel.}
For fixed $\rvx_t$, the posterior weights $P_{1\mid t}(\cdot \mid \rvx_t)$ are nonnegative and sum to one over $\gX$. The kernel is therefore nonnegative, and
\begin{equation}
    \int_{\sS_K^L} p_{t'\mid t}(\rvx_t' \mid \rvx_t)\, \mathrm{d}\rvx_t'
    \;=\; \sum_{\rvx_1 \in \gX} P_{1\mid t}(\rvx_1 \mid \rvx_t) \int_{\sS_K^L} q(\rvx_t' \mid \rvx_1)\, \mathrm{d}\rvx_t'
    \;=\; \sum_{\rvx_1 \in \gX} P_{1\mid t}(\rvx_1 \mid \rvx_t)
    \;=\; 1 .
\end{equation}

\medskip
\noindent\textbf{2. $\pi$ is stationary.}
We first note the Bayes identity
\begin{equation}
\label{eq:bayes-identity}
    \pi(\rvx_t)\, P_{1\mid t}(\rvx_1 \mid \rvx_t) \;=\; p(\rvx_1)\, q(\rvx_t \mid \rvx_1)
    \qquad \text{for all } \rvx_t \in \sS_K^L,\ \rvx_1 \in \gX .
\end{equation}
Where $\pi(\rvx_t) > 0$, this is Equation~\ref{eq:exact-posterior} multiplied by $\pi(\rvx_t)$. Where $\pi(\rvx_t) = 0$, both sides vanish, since $\pi(\rvx_t) = 0$ implies $p(\rvx_1)\, q(\rvx_t \mid \rvx_1) = 0$ for all $\rvx_1 \in \gX$.

Using Equation~\ref{eq:kernel-def}, exchanging the finite sum and the integral, and applying Equation~\ref{eq:bayes-identity},
\begin{align}
    (\pi Q_t)(\rvx_t')
    &= \int_{\sS_K^L} \pi(\rvx_t)\, p_{t'\mid t}(\rvx_t' \mid \rvx_t)\, \mathrm{d}\rvx_t \\
    &= \sum_{\rvx_1 \in \gX} q(\rvx_t' \mid \rvx_1) \int_{\sS_K^L} \pi(\rvx_t)\, P_{1\mid t}(\rvx_1 \mid \rvx_t)\, \mathrm{d}\rvx_t \\
    &= \sum_{\rvx_1 \in \gX} q(\rvx_t' \mid \rvx_1)\, p(\rvx_1) \int_{\sS_K^L} q(\rvx_t \mid \rvx_1)\, \mathrm{d}\rvx_t \\
    &= \sum_{\rvx_1 \in \gX} p(\rvx_1)\, q(\rvx_t' \mid \rvx_1)
    \;=\; \pi(\rvx_t') .
\end{align}
Hence $\pi Q_t = \pi$: the marginal $\pi$ is stationary for $p_{t'\mid t}$.

\medskip
\noindent\textbf{3. Doeblin minorization.}
Let
\begin{equation}
\label{eq:floor}
    g(\rvx) \;:=\; \min_{\rvx_1 \in \gX}\, q(\rvx \mid \rvx_1),
    \qquad \text{so that} \qquad
    \varepsilon \;=\; \int_{\sS_K^L} g(\rvx)\, \mathrm{d}\rvx
\end{equation}
by Equation~\ref{eq:overlap}. The function $g$ is the pointwise lower envelope of the noising densities, and $\varepsilon$ is the probability mass they all share. We have $\varepsilon > 0$ by \textup{(A1)}, and $\varepsilon \le 1$ since $g \le q(\cdot \mid \rvx_1)$.

Whichever $\rvx_1$ is sampled, the renoising density lies above $g$, and since the posterior weights sum to one (Step~1), so does their mixture: for all $\rvx_t, \rvx_t'$,
\begin{equation}
    p_{t'\mid t}(\rvx_t' \mid \rvx_t)
    \;=\; \sum_{\rvx_1 \in \gX} P_{1\mid t}(\rvx_1 \mid \rvx_t)\, q(\rvx_t' \mid \rvx_1)
    \;\ge\; g(\rvx_t') .
\end{equation}
Writing $\nu := g / \varepsilon$ for the normalized floor, this is the uniform Doeblin minorization
\begin{equation}
\label{eq:doeblin}
    p_{t'\mid t}(\rvx_t' \mid \rvx_t) \;\ge\; \varepsilon\, \nu(\rvx_t')
    \qquad \text{for all } \rvx_t, \rvx_t' .
\end{equation}

\medskip
\noindent\textbf{4. Geometric convergence and uniqueness.}
If $\varepsilon = 1$, then each $q(\cdot \mid \rvx_1)$ coincides with $g$, so $p_{t'\mid t}(\cdot \mid \rvx_t) = \pi$ for every $\rvx_t$ and the chain reaches $\pi$ in one step. Otherwise, by Equation~\ref{eq:doeblin}, each transition can be written as a mixture
\begin{equation}
\label{eq:doeblin-th}
    p_{t'\mid t}(\cdot \mid \rvx_t) \;=\; \varepsilon\, \nu \;+\; (1 - \varepsilon)\, r(\cdot \mid \rvx_t),
    \qquad
    r(\cdot \mid \rvx_t) := \frac{p_{t'\mid t}(\cdot \mid \rvx_t) - \varepsilon\, \nu}{1 - \varepsilon},
\end{equation}
where $r$ is again a transition density: it is nonnegative by Equation~\ref{eq:doeblin} and integrates to one. In words, at every step the chain ignores its current state with probability $\varepsilon$ and draws a fresh sample from $\nu$.

Now compare the chain started from two distributions $\mu$ and $\mu'$, and let $R$ denote the Markov operator of $r$. The fresh-sample component $\varepsilon \nu$ is the same for both, so it cancels after one step:
\begin{equation}
    \mu Q_t - \mu' Q_t \;=\; (1 - \varepsilon)\, (\mu R - \mu' R).
\end{equation}
Since applying a Markov kernel never increases the total variation distance,
\begin{equation}
\label{eq:contraction}
    \|\mu Q_t - \mu' Q_t\|_{\mathrm{TV}} \;\le\; (1 - \varepsilon)\, \|\mu - \mu'\|_{\mathrm{TV}} .
\end{equation}
That is, each step shrinks the distance between any two distributions by a factor of at least $1 - \varepsilon$.

Taking $\mu = \mu_0$ and $\mu' = \pi$, using $\pi Q_t = \pi$ (Step~2), and iterating $n$ times gives
\begin{equation}\label{eq:total_variation}
    \|\mu_0 Q_t^n - \pi\|_{\mathrm{TV}}
    \;\le\; (1 - \varepsilon)^n\, \|\mu_0 - \pi\|_{\mathrm{TV}}
    \;\le\; (1 - \varepsilon)^n ,
\end{equation}
which is Equation~\ref{eq:geometric-rate}. 

Finally, if $\pi'$ is another stationary distribution, then $\|\pi' - \pi\|_{\mathrm{TV}} = \|\pi' Q_t^n - \pi Q_t^n\|_{\mathrm{TV}} \le (1 - \varepsilon)^n \to 0$, so $\pi' = \pi$, which proves uniqueness.

\subsection{Proposition~\ref{prop:error_contraction}: UNSID Contracts Errors, Flow Matching Preserves Them}

Let $p$ and $\gX$ be as in Section~\ref{sec:method}. For each $t \in [0,1)$ and $\rvx_1 \in \gX$, let $q_t(\cdot \mid \rvx_1)$ be the noising density at time $t$ on $\sS_K^L$, with marginal
\begin{equation}
    \pi_t(\rvx) \;:=\; \sum_{\rvx_1 \in \gX} p(\rvx_1)\, q_t(\rvx \mid \rvx_1).
\end{equation}
Fix a time $t$ and a step size $h > 0$ with $t + h < 1$, and let $\tilde\pi_t$ be \emph{any} distribution on $\sS_K^L$, representing the law of the sampler at time $t$, including the error accumulated so far.
\begin{enumerate}
    \item[\textup{(a)}] \textbf{UNSID contracts the error.} Let $P_{1\mid t}$ be the true posterior, let
    \begin{equation}
        p_{t+h}(\rvx \mid \rvx_t) \;=\; \sum_{\rvx_1 \in \gX} P_{1\mid t}(\rvx_1 \mid \rvx_t)\, q_{t+h}(\rvx \mid \rvx_1)
    \end{equation}
    be the UNSID denoising kernel (Equation~\ref{eq:denoising}), and let $Q$ be the associated Markov operator. Assume that the noising densities at time $t+h$ overlap:
    \begin{equation}
    \label{eq:overlap-th}
        \varepsilon_{t+h} \;:=\; \int_{\sS_K^L} \min_{\rvx_1 \in \gX}\, q_{t+h}(\rvx \mid \rvx_1)\, \mathrm{d}\rvx \;>\; 0.
    \end{equation}
    Then
    \begin{equation}
    \label{eq:unsid-contraction2}
        \|\tilde\pi_t Q - \pi_{t+h}\|_{\mathrm{TV}}
        \;\le\; (1 - \varepsilon_{t+h})\, \|\tilde\pi_t - \pi_t\|_{\mathrm{TV}}.
    \end{equation}
    \item[\textup{(b)}] \textbf{Flow matching preserves the error.} Let $\Phi : \sS_K^L \to \sS_K^L$ be the flow-matching step from $t$ to $t+h$. Assume that $\Phi$ is a measurable bijection with measurable inverse and that it transports the true marginals, $\Phi_\# \pi_t = \pi_{t+h}$. Then
    \begin{equation}
    \label{eq:fm-preservation2}
        \|\Phi_\# \tilde\pi_t - \pi_{t+h}\|_{\mathrm{TV}}
        \;=\;
        \|\tilde\pi_t - \pi_t\|_{\mathrm{TV}} .
    \end{equation}
\end{enumerate}

\medskip
\noindent\textbf{(a) UNSID contracts the error.}
We first show that $\pi_t Q = \pi_{t+h}$. By the Bayes identity at time $t$ (Equation~\ref{eq:bayes-identity}), $\int_{\sS_K^L} \pi_t(\rvx_t)\, P_{1\mid t}(\rvx_1 \mid \rvx_t)\, \mathrm{d}\rvx_t = p(\rvx_1)$, hence
\begin{align}
    (\pi_t Q)(\rvx)
    &=\; \sum_{\rvx_1 \in \gX} q_{t+h}(\rvx \mid \rvx_1) \int_{\sS_K^L} \pi_t(\rvx_t)\, P_{1\mid t}(\rvx_1 \mid \rvx_t)\, \mathrm{d}\rvx_t \\
    &=\; \sum_{\rvx_1 \in \gX} p(\rvx_1)\, q_{t+h}(\rvx \mid \rvx_1) \\
    &=\; \pi_{t+h}(\rvx).    
\end{align}
Next, Steps~3 and~4 of the proof of Proposition~\ref{prop:conv} only use the facts that the posterior weights sum to one and that the renoising densities overlap. Applied with $q_{t+h}$ in place of $q_t$, they yield the minorization $p_{t+h}(\cdot \mid \rvx_t) \ge \varepsilon_{t+h}\, \nu_{t+h}$, where $\nu_{t+h} := \min_{\rvx_1 \in \gX} q_{t+h}(\cdot \mid \rvx_1) / \varepsilon_{t+h}$. They therefore also yield the contraction of Equation~\ref{eq:contraction}, with $Q$ and $\varepsilon_{t+h}$ in place of $Q_t$ and $\varepsilon$. Taking $\mu = \tilde\pi_t$ and $\mu' = \pi_t$,
\begin{equation}
    \|\tilde\pi_t Q - \pi_{t+h}\|_{\mathrm{TV}}
    \;=\; \|\tilde\pi_t Q - \pi_t Q\|_{\mathrm{TV}}
    \;\le\; (1 - \varepsilon_{t+h})\, \|\tilde\pi_t - \pi_t\|_{\mathrm{TV}}.
\end{equation}

\medskip
\noindent\textbf{(b) Flow matching preserves the error.}
For any measurable set $A$, $(\Phi_\# \tilde\pi_t)(A) - (\Phi_\# \pi_t)(A) = \tilde\pi_t(B) - \pi_t(B)$ with $B = \Phi^{-1}(A)$. Because $\Phi$ is a measurable bijection with measurable inverse, $A \mapsto \Phi^{-1}(A)$ is a one-to-one correspondence between measurable sets: every measurable $B$ arises this way, from $A = \Phi(B)$. Taking the supremum over $A$ is therefore the same as taking it over $B$:
\begin{equation}
    \|\Phi_\# \tilde\pi_t - \Phi_\# \pi_t\|_{\mathrm{TV}}
    \;=\; \sup_{A} \bigl|\tilde\pi_t(\Phi^{-1}(A)) - \pi_t(\Phi^{-1}(A))\bigr|
    \;=\; \sup_{B} \bigl|\tilde\pi_t(B) - \pi_t(B)\bigr|
    \;=\; \|\tilde\pi_t - \pi_t\|_{\mathrm{TV}} .
\end{equation}
Using $\Phi_\# \pi_t = \pi_{t+h}$ gives Equation~\ref{eq:fm-preservation2}.
\section{Probability Path and Prior Choices}\label{ap:path_choice}

A natural choice for the noise prior is the uniform distribution over the simplex,
\( q_{0}(\rvx) = \mathrm{Dir}(\vone) \).
However, graphs are typically sparse, with highly imbalanced edge distributions. In discrete diffusion, \citet{digress} has shown that the empirical marginal distribution is theoretically optimal and empirically effective. 

Motivated by this result, we consider a marginal-weighted Dirichlet mixture prior:
\(
q_{0}^{\mathrm{marg}}(\rvx) = \sum_{k=1}^{K} m_{k}\mathrm{Dir}(\rvx; \vone + \kappa\ve_{k}),
\)
where \(\vm = (m_{1},\ldots,m_{K})\) denotes the empirical marginal over categories, \(\ve_{k}\) is the \(k\)th standard basis vector, and \(\kappa > 0\) is a user-specified concentration parameter. 

We can directly parametrize the prior as such, but we can also approximate this prior using a model trained with the probability path
\(
q_t^\star(\rvx_t \mid \rvx_1) := \mathrm{Dir}(\rvx_t; \vone + \alpha_t \rvx_1).
\)
In that case, we have dimension-wise:
\(
\mathbb{E}_{\rvx_1 \sim p_{\text{data}}}\left[ q_t^\star(\rvx_t \mid \rvx_1) \right] = q^{\mathrm{marg}}_0(\rvx_0)
\quad \iff \quad \alpha_t = \kappa.
\)
This identity does not directly extend to the \(L\)-dimensional joint, since in general the \(L\) dimensions are not independent. Nevertheless, when \(q_{t}^{\star}\) is sufficiently close to the stationary distribution \(q_{0}^{\star}\), we can leverage the following approximation:
\begin{equation}
q_t^\star\bigl(\rvx^{(1:L)}_t \mid \rvx_1^{(1:L)}\bigr) \approx
\prod_{l=1}^L q_t^\star\bigl(\rvx^{(l)}_t \mid \rvx_1^{(l)}\bigr)
= q^{\mathrm{marg}}_{0}(\rvx^{(1:L)}),
\end{equation}
with $\alpha_t = \kappa$. 

Operationally, we can thus initialize inference with 
\(
\rvx_0^{(1:L)} \sim q^{\mathrm{marg}}_{0},
\)
and then run the reverse process along the path \(q_t^\star\) with \(\alpha_0 = \kappa\). 
We show in Section \ref{sec:eval} that this initialization improves generative performance.
\section{Supplementary Discussion}

\subsection{Extended Related Work on Graph Generation}\label{ap:related_works}

A central challenge in generative graph modeling arises from node permutations: a graph with $n$ nodes can be represented by up to $n!$ different adjacency matrices. 
This has motivated two dominant families of methods: \emph{sequential} (autoregressive) approaches, which generate graphs step by step by adding nodes, edges, or higher-order motifs according to a fixed or learned ordering \citep{graphrnn, graphaf, graphdf, gran, graphgen, grapharm, pard}, and \emph{permutation-equivariant} models, which enforce equivariance by design. Permutation-equivariant architectures have been instantiated within several generative frameworks, including variational autoencoders \citep{graphvae}, generative adversarial networks \citep{molgan, gggan, spectre}, normalizing flows \citep{graphnvp, moflow, gnf}, and vector-quantized autoencoders \citep{dgae, glad}.

Recently, \emph{denoising models} coupled with equivariant GNNs have become the dominant approach, as they have proven highly effective. Continuous-space methods include score-based diffusion \citep{edp-gnn, gdss}, diffusion bridges \citep{drum}, and flow matching \citep{CatFlow}. Discrete formulations, such as diffusion in discrete time \citep{discdiff_haefeli, digress} or in continuous time \citep{continuous_time_graph_diffusion}, flow matching \citep{DeFog}, and non-Markovian discrete diffusion \citep{SID}, also report strong results. However, these approaches do not perform \emph{smooth} diffusion on the probability simplex: they either add noise to graphs in an unconstrained Euclidean space or operate as jump processes over discrete states.

Beta diffusion \citep{BetaGraph} applies continuous noise on the $1$-simplex via the beta distribution and handles $K$-categorical variables by encoding them as $K$ binary variables. \citet{CatFlow} implement Dirichlet flow matching for graphs as a baseline but report poor performance (consistent with our \textsc{DiriFM} results), which leads them to relax the noising process to $\sR^K$. More recently, the Graph Bayesian Flow Network \citep{graphBFN} proposes a diffusion mechanism that adds Gaussian noise in $\sR^K$, followed by a projection onto the simplex. 

Orthogonally, a recent line of work targets scalability. A key limitation of standard models is their reliance on dense representations and all-pairs computations, which impedes scaling to large graphs. To address this, prior work has proposed scalable denoising architectures and sampling strategies \citep{sparsediff, EDGE, higen, graphle, head}. Notably, \citet{sparsediff} and \citet{head} introduce methods that enable permutation-equivariant models to scale to large graphs. These techniques are complementary to our approach and could be used to scale \textsc{UNSID}; we leave this exploration to future work.

\subsection{Denoiser Expressivity in Discrete and Continuous Settings}\label{ap:expressivity}

GNNs have limited expressivity because of inherent graph symmetries \citep{Morris2019}. These limitations, however, disappear once each node has a unique identifier. In particular, \citet{Loukas_WhatGNN} demonstrates that sufficiently deep message-passing architectures become universal when nodes can be uniquely distinguished. Continuous noise naturally provides such distinguishing information, effectively acting as a unique identifier for each node. Consequently, injecting continuous noise enhances the expressive power of GNN-based denoisers.

In practice, discrete denoisers often require additional handcrafted features to overcome symmetry-induced limitations. Common examples include spectral embeddings derived from Laplacian eigenvectors \citep{digress} and relative random-walk probability features \citep{DeFog}. These features must be recomputed for every noisy graph, during both training and sampling, adding non-negligible computational overhead.

A typical criticism of continuous models is that adding continuous noise to edge attributes may collapse the graph into a fully connected weighted structure, ostensibly erasing the discrete adjacency information. However, the graph operations underlying the aforementioned extra features---such as spectral decompositions and random-walk statistics---remain well defined on weighted graphs. Thus, continuous perturbations do not destroy the structure required for these computations. Moreover, since continuous noise serves as an intrinsic node identifier, it eliminates the need for such auxiliary features altogether.

\subsection{Comparison with Simple Iterative Denoising}\label{ap:method_comparison}

Our \textsc{UNSID} framework can be viewed as a simplex analogue of Simple Iterative Denoising (SID) \citep{SID}. Below, we outline the key similarities and differences.

\paragraph{Noising.}
In SID, the corruption process follows discrete diffusion for categorical data \citep{discdiff_austin}, where intermediate noisy distributions linearly interpolate between a point mass at the clean data and a stationary prior:
\begin{equation}\label{eq:noising}
q_t(\rvx \mid \rvx_1) = \alpha_t \delta_{\rvx_1}(\rvx) + (1-\alpha_t) q_0(\rvx).
\end{equation}
As discussed in Section~\ref{sec:noising}, interpolation-based constructions are ill-suited to the continuous simplex setting. Accordingly, \textsc{UNSID} adopts an explicit probability path on the simplex.

\paragraph{Independence Assumption.}
Both SID and \textsc{UNSID} rely on the same conditional-independence assumption (Assumption~\ref{ass:conditional_indep}), which is central to the formulation.

\paragraph{Denoising.}
The denoising rules in \textsc{UNSID} follow directly from the conditional-independence assumption and are closely related in spirit to SID. A notable distinction is our original hybrid construction, which couples a \emph{continuous} kernel $q_t$ on the simplex with a \emph{discrete} posterior $P_{1\mid t}$. 

\paragraph{Convergence.}
Proposition~\ref{prop:conv} and its proof are original to this work. SID provides no analogous convergence result. We regard this result as a core theoretical contribution.

\paragraph{Advantages over Diffusion and Flow Matching.}
Although there are conceptual similarities at a high level, including the connection to compounding denoising errors, the underlying intuition differs substantially between the continuous and discrete settings. In particular, the observation that $\rvx_{t+h}$ remains close to $\rvx_t$, and the reasoning that follows from it, hold only in the continuous case and do not translate directly to discrete models.

\paragraph{Parameterization and Learning.}
Both approaches employ standard parameterizations and training objectives common to discrete diffusion and flow-matching methods.

\paragraph{Probability Paths and Priors.}
Our choices and analysis of probability paths and priors are specific to continuous simplex noise and have no direct counterpart in SID.

\paragraph{Guidance.}
As explained in the main text, guidance in \textsc{UNSID} is obtained via direct adaptations of standard diffusion techniques (classifier and classifier-free guidance). SID does not include a guidance mechanism.

Beyond this point-by-point comparison, our primary contribution is integrating these components into a simple, efficient framework for graph generation that achieves state-of-the-art performance on three of the five datasets considered, while remaining competitive with the strongest baselines on the other two.

\subsection{Comments on the \texttt{SBM} Experiments}\label{ap:sbm}

MMD-based metrics do not penalize dataset memorization: a model that reproduces training graphs obtains near-perfect validity and MMDs close to the training-set reference, and a model that reproduces test graphs obtains near-zero MMDs. On \texttt{SBM}, two independent observations point in this direction for \textsc{DeFoG}:

\begin{itemize}
    \item \textbf{Novelty.} \texttt{SBM} graphs have on the order of 100 nodes, i.e., thousands of node pairs, so the probability that two independently sampled graphs are isomorphic is negligible. A novelty of 90\% therefore cannot be explained by chance and indicates that about 10\% of the samples reproduce training graphs.
    \item \textbf{Reference MMD.} The MMD between the training and test splits is the natural reference level for a model that generalizes. \textsc{DeFoG} reports a degree MMD of 0.6, less than half of this reference (1.57), which is difficult to reconcile with sampling from the data distribution and may indicate overfitting to the evaluation split.
\end{itemize}

Taken together, these observations suggest that the validity and MMD results of \textsc{DeFoG} on \texttt{SBM} should be interpreted with caution. 

\subsection{Use of LLMs}

We used large language models (LLMs) for editing, consistency checks, identifying typographical errors, and formatting tables. LLMs were also used to check and correct the proofs and to improve their readability. No code, analyses, or results were generated by LLMs.
\section{Guidance}\label{ap:guidance}

Classifier guidance and classifier-free guidance are two standard mechanisms for steering generation toward desired properties. Classifier guidance augments an \emph{unconditional} generative model with the gradient of a separately trained classifier (or regressor) for the target property. Classifier-free guidance does not require an auxiliary classifier, but it does require both a \emph{conditional} and an \emph{unconditional} generative model, typically obtained by training a single network with randomly masked conditioning vectors.

Below, we first present the key ideas and then provide the derivations.

\subsection{Key Ideas}

Let $y$ denote the target property, let $p(y \mid \rvx)$ be the conditional distribution of $y$ given a noisy state $\rvx$, and let $p_{t+h\mid t}(\rvx_{t+h} \mid \rvx_t)$ denote the unconditional reverse transition from time $t$ to $t+h$. We modulate the strength of conditioning with a scale $\omega > 0$ via
\begin{equation}
    p^\omega_{t+h\mid t}(\rvx_{t+h} \mid \rvx_t, y) \;\propto\; p(y \mid \rvx_{t+h})^\omega\, p_{t+h\mid t}(\rvx_{t+h} \mid \rvx_t).
\end{equation}

\paragraph{Classifier guidance.}
Here, $p(y \mid \rvx)$ is approximated by a classifier $p_\phi(y \mid \rvx)$ trained at all noise levels. During sampling, one combines a reverse step with an update in the direction of the classifier gradient $\nabla_{\rvx} \log p_\phi(y \mid \rvx)\rvert_{\rvx = \rvx_t}$.

\paragraph{Classifier-free guidance.}
In this approach, no external classifier is needed; instead, one uses a conditional reverse model $p_{t+h\mid t}(\rvx_{t+h} \mid \rvx_t, y)$ and an unconditional one $p_{t+h\mid t}(\rvx_{t+h} \mid \rvx_t)$, typically produced by a single network via masking. A standard identity yields the log-linear interpolation
\begin{equation}
    \log p^\omega_{t+h\mid t}(\rvx_{t+h} \mid \rvx_t, y)
    = \omega \log p_{t+h\mid t}(\rvx_{t+h} \mid \rvx_t, y)
    + (1-\omega) \log p_{t+h\mid t}(\rvx_{t+h} \mid \rvx_t) + C,
\end{equation}
where $C$ does not depend on $\rvx_{t+h}$.

\subsection{Derivations}

Assume that, given $\rvx_{t+h}$, the noisier state $\rvx_t$ carries no additional information about $y$, i.e., $q(\rvx_t \mid \rvx_{t+h}, y) = q(\rvx_t \mid \rvx_{t+h})$. By Bayes' rule and this assumption,
\begin{equation}
    p(y \mid \rvx_t, \rvx_{t+h})
    = \frac{q(\rvx_t \mid \rvx_{t+h}, y)\, p(y \mid \rvx_{t+h})}{q(\rvx_t \mid \rvx_{t+h})}
    = p(y \mid \rvx_{t+h}).
\end{equation}
Consequently,
\begin{equation}
    p_{t+h\mid t}(\rvx_{t+h} \mid \rvx_t, y)
    = \frac{p(y \mid \rvx_t, \rvx_{t+h})\, p_{t+h\mid t}(\rvx_{t+h} \mid \rvx_t)}{p(y \mid \rvx_t)}
    \;\propto\; p(y \mid \rvx_{t+h})\, p_{t+h\mid t}(\rvx_{t+h} \mid \rvx_t).
\end{equation}
Introducing the scale $\omega > 0$ controls the strength of the guidance, with $\omega > 1$ sharpening it:
\begin{equation}\label{eq:guidance_propto}
    p^\omega_{t+h\mid t}(\rvx_{t+h} \mid \rvx_t, y) \;\propto\; p(y \mid \rvx_{t+h})^\omega\, p_{t+h\mid t}(\rvx_{t+h} \mid \rvx_t).
\end{equation}

The above assumption holds exactly for Markovian noising processes. Under Assumption~\ref{ass:conditional_indep}, however, $\rvx_t$ and $\rvx_{t+h}$ are conditionally independent noisy views of $\rvx_1$, so $\rvx_t$ may carry additional information about $y$. In other words, Equation~\ref{eq:guidance_propto} is exact for Markovian models and serves as an approximation for UNSID.

\subsubsection{Classifier-Free Guidance}

Applying Bayes' rule to $p(y \mid \rvx_{t+h}) = p(y \mid \rvx_t, \rvx_{t+h})$ in Equation~\ref{eq:guidance_propto}, we obtain
\begin{align}
    p^\omega_{t+h\mid t}(\rvx_{t+h} \mid \rvx_t, y)
    &\propto \left(\frac{p_{t+h\mid t}(\rvx_{t+h} \mid \rvx_t, y)\, p(y \mid \rvx_t)}{p_{t+h\mid t}(\rvx_{t+h} \mid \rvx_t)}\right)^{\omega} p_{t+h\mid t}(\rvx_{t+h} \mid \rvx_t) \\
    &= p_{t+h\mid t}(\rvx_{t+h} \mid \rvx_t, y)^{\omega}\, p_{t+h\mid t}(\rvx_{t+h} \mid \rvx_t)^{1-\omega}\, p(y \mid \rvx_t)^{\omega}.
\end{align}
Since $p(y \mid \rvx_t)^\omega$ does not depend on $\rvx_{t+h}$, taking the logarithm and absorbing this term into the constant yields
\begin{equation}
    \log p^\omega_{t+h\mid t}(\rvx_{t+h} \mid \rvx_t, y)
    = \omega \log p_{t+h\mid t}(\rvx_{t+h} \mid \rvx_t, y)
    + (1-\omega) \log p_{t+h\mid t}(\rvx_{t+h} \mid \rvx_t) + C.
\end{equation}

\subsubsection{Classifier Guidance}

For classifier guidance, we start from Equation~\ref{eq:guidance_propto}, approximate $p(y \mid \rvx_{t+h})$ by the classifier $p_\phi(y \mid \rvx_{t+h})$, and linearize its logarithm around $\rvx_t$ with a first-order Taylor expansion:
\begin{equation}
    \log p_\phi(y \mid \rvx_{t+h})
    \approx \log p_\phi(y \mid \rvx_t) + (\rvx_{t+h} - \rvx_t)^\top \vg_t
    = \rvx_{t+h}^\top \vg_t + C
    = \sum_{i=1}^{L} \bigl(\rvx_{t+h}^{(i)}\bigr)^\top \vg_t^{(i)} + C,
\end{equation}
where $\vg_t := \nabla_{\rvx} \log p_\phi(y \mid \rvx)\rvert_{\rvx = \rvx_t}$, $\vg_t^{(i)}$ is its block for dimension $i$, and $C$ does not depend on $\rvx_{t+h}$. Substituting into Equation~\ref{eq:guidance_propto} gives
\begin{equation}\label{eq:classifier_guidance}
    \log p^\omega_{t+h\mid t}(\rvx_{t+h} \mid \rvx_t, y)
    \approx \log p_{t+h\mid t}(\rvx_{t+h} \mid \rvx_t) + \omega\, \rvx_{t+h}^\top \vg_t + C.
\end{equation}
This shows that classifier guidance tilts the unconditional reverse kernel in the direction of the classifier gradient, toward states with a higher likelihood of the target property.
\section{Technical Report}\label{ap:technical}

\subsection{GNN Architecture}

Our denoiser uses the same architecture as SID \citep{SID}; we reproduce its description below for completeness. The denoisers are GNNs inspired by the general, powerful, scalable (GPS) graph Transformer. A single layer is described as
\begin{align}
    \tilde{\mX}^{\ell}, \tilde{\mE}^{\ell} &= \text{MPNN}(\mX^{\ell}, \mE^{\ell}), \\
    \mX^{\ell+1} &= \text{MultiheadAttention}(\tilde{\mX}^{\ell} + \mX^{\ell}) + \tilde{\mX}^{\ell}, \\
    \mE^{\ell+1} &= \tilde{\mE}^{\ell} + \mE^{\ell},
\end{align}
where $\mX^{\ell}$ and $\mE^{\ell}$ are the node and edge hidden representations at the input of the $l$th layer. The MultiheadAttention layer is the classical multi-head attention layer of \citet{attention_is_all}, and MPNN is a message-passing neural network layer described below.

The MPNN operates on node and edge representations as follows:
\begin{align}
    \vh^{\ell}_{i, j} &= \text{ReLU}\bigl(\mW^{\ell}_{\text{src}}\vx^{\ell}_i + \mW^{\ell}_{\text{trg}}\vx^{\ell}_j + \mW^{\ell}_{\text{edge}}\va^{\ell}_{i, j}\bigr), \\
    \va^{\ell+1}_{i, j} &= \text{LayerNorm}\bigl(f_{\text{edge}}(\vh^{\ell}_{i, j})\bigr), \\
    \vx^{\ell+1}_{i} &= \text{LayerNorm}\Bigl(\vx^{\ell}_{i} + \sum_{j \in \mathcal{N}(i)} f_{\text{node}}(\vh^{\ell}_{i, j})\Bigr),
\end{align}
with $\mW^{\ell}_{\text{src}}$, $\mW^{\ell}_{\text{trg}}$, and $\mW^{\ell}_{\text{edge}}$ denoting trainable weight matrices, and $f_{\text{node}}$ and $f_{\text{edge}}$ being small neural networks.

The node hidden states $\vx_i$ and the outputs of $f_{\text{node}}$ are of dimension $d_h$, a tunable hyperparameter (see Table~\ref{tab:hyperparam_denoising_fix}). The edge hidden states $\va_{i,j}$, intermediate messages $\vh^{\ell}_{i,j}$, and the outputs of $f_{\text{edge}}$ have dimension $d_h/2$ or $d_h/4$.

\paragraph{Inputs and Outputs}
At the input, we concatenate the node attributes, extra features, and time step into node features, copying graph-level information (e.g., the time step or graph size) to each node. The node and edge input vectors are then projected to their respective hidden dimensions, $d_h$ for nodes and $d_h/4$ for edges.

Similarly, the outputs of the final layer are projected to their respective dimensions, $d_v$ for nodes and $d_a + 1$ for edges. To enforce edge symmetry, we replace each edge output $\va_{i,j}$ by $(\va_{i,j} + \va_{j,i})/2$. Finally, we ensure that the outputs can be interpreted as probabilities by applying a softmax or sigmoid function, as appropriate.

\subsection{Probability Path and Noise Schedule}

\begin{figure}[H]
    \begin{center}
     \includegraphics[width=0.4\textwidth]{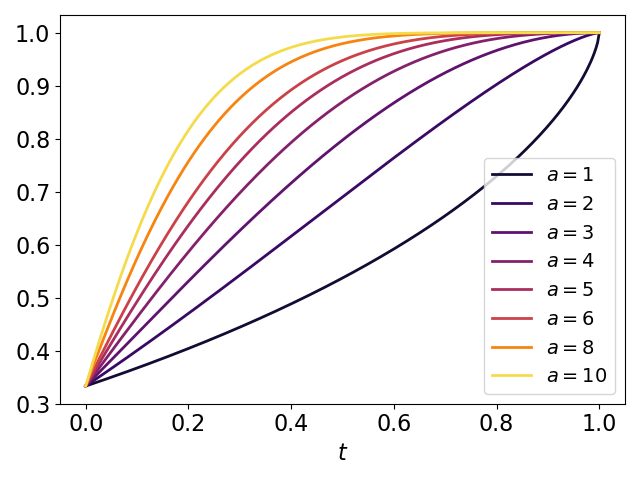}
    \end{center}
    \caption{
        Voronoi probabilities over time for $\rvx_t \sim \mathrm{Dir}(\vone + \alpha_t\rvx_1)$ on $\sS_3$, with $\alpha_t = -a\log(1-t)$, for various values of $a$. For $a = 1$, $P_{v_k}$ remains low for most of the trajectory and only increases sharply as $t \to 1$; for $a = 10$, $P_{v_k}$ is already close to $1$ by $t = 0.6$. Suitable choices of $a$ therefore likely lie between $1$ and $10$.}
    \label{fig:voronoi}
\end{figure}

For all experiments, we use a probability path parameterized via the Dirichlet distribution, $\mathrm{Dir}(\rvx_t; \vone + \alpha_t\rvx_1)$, with $\alpha_t = -a\log(1-t)$. The hyperparameter $a$ controls the noising dynamics (see Figure~\ref{fig:voronoi}). We use $a = 3$ in all our experiments, except on \texttt{SBM}, where we use $a = 2$.

\subsection{Hyperparameters}

\begin{table}[H]
    \centering
    \caption{Hyperparameters.}
    \begin{tabular}{lcccc}
        \toprule
        & \texttt{Planar} & \texttt{SBM} & \texttt{QM9H} & \texttt{ZINC250K} \\
        \midrule
        GNN layers   & 8 & 8 & 8 & 8 \\
        Hidden layers in MLPs   & 2 & 2 & 2 & 2 \\
        Node representation size & 256 & 256 & 256 & 256 \\
        Edge representation size  & 64 & 128 & 128 & 128 \\
        Denoising steps & 128 & 512 & 1024 & 1024 \\
        Learning rate &  0.0002 &  0.0005 &  0.0005 &  0.0005 \\
        Optimizer & AdamW & AdamW & AdamW & AdamW \\ 
        \bottomrule
    \end{tabular}
    \label{tab:hyperparam_denoising_fix}
\end{table}

For discrete diffusion (\textsc{DiscDiff} and \textsc{NM-DD}), we use the cosine schedule \citep{cosine_sched} and the marginal noise distribution. 

\subsection{Extra Features}

For discrete diffusion (Markovian and non-Markovian, i.e., \textsc{DiscDiff} and \textsc{NM-DD}), we follow common practice \citep{digress, DeFog, head} and enhance the discrete graph representation with synthetic extra node features. For \textsc{UNSID} and Dirichlet flow matching, we add only the graph size. 

We use the following extra features: spectral features (\texttt{ZINC250K}, \texttt{Planar}, \texttt{SBM}), graph size, molecular features (\texttt{ZINC250K}), cycle information (\texttt{Planar}), and relative random-walk probabilities (RRWPs; \texttt{QM9H}). All these features are concatenated to the input node attributes; the RRWPs are additionally concatenated to the edge attributes.

\paragraph{Spectral features}
We use the eigenvectors associated with the $k$ lowest eigenvalues of the graph Laplacian. Additionally, we concatenate the corresponding $k$ lowest eigenvalues to each node.

\paragraph{Graph size encoding}
The graph size is encoded as the ratio between the size of the current graph and that of the largest graph in the dataset, $n/n_{\text{max}}$. This value is concatenated to all nodes in the graph.

\paragraph{Molecular features}
For \texttt{ZINC250K}, we use the charge and valency of each atom as additional features.

\paragraph{Cycles}
Following \citet{digress}, we count the number of cycles of size 3, 4, and 5 that each node is part of, and use these counts as features.

\paragraph{Relative random-walk probabilities}
RRWPs are the probabilities of reaching a target node from a source node in a given number of steps. We compute them for all numbers of steps from $1$ to $k$.

\subsection{Evaluation}\label{ap:eval}

\subsubsection{Molecule Generation}
For \texttt{ZINC250K}, we adopt the evaluation procedure of \citet{drum} (from which we also take baseline results), originally established by \citet{gdss}. We assess performance using three standard metrics: (i) the Fr\'echet ChemNet Distance (FCD) \citep{fcd}, which measures similarity in chemical feature space; (ii) the Neighborhood Subgraph Pairwise Distance Kernel (NSPDK) \citep{nspdk}, which evaluates graph-structural similarity; and (iii) validity, the fraction of chemically valid molecules without any post hoc correction or resampling. We report means over five sampling runs, each generating 10{,}000 molecules. For completeness, we also provide standard deviations, as well as uniqueness and novelty statistics, in Appendix~\ref{ap:results}.

\paragraph{QM9H}
For \texttt{QM9H}, we generally follow the above procedure. For evaluation, we rely on SMILES with explicit hydrogen atoms to compute validity, uniqueness, and FCD. Under this evaluation procedure, a small fraction of the molecules in the dataset are themselves invalid. We use the kekulized version of the dataset (i.e., with three bond types: single, double, and triple). Some concurrent implementations use a variant of the dataset with explicit aromatic bonds; in such cases, evaluation metrics, in particular validity and FCD, are not directly comparable to ours.

\subsubsection{Unattributed Synthetic Graph Generation}
For unattributed graphs, we follow the procedure of \citet{spectre}: an 80/20 train--test split, with 20\% of the training set used for validation. We measure distributional similarity via MMD over degree distributions, clustering coefficients, orbit counts, and spectral densities, and we additionally report validity. For \texttt{Planar}, a valid graph must be planar and connected. For \texttt{SBM}, validity indicates consistency with the data-generating stochastic block model (intra-community edge density $0.3$, inter-community edge density $0.005$). We omit uniqueness because all models reach $100\%$ on both datasets, and we report novelty only for \texttt{SBM}, since it is $100\%$ for all models on \texttt{Planar}. Uniqueness is the fraction of distinct graphs among the generated samples; novelty is the fraction of unique graphs not present in the training set. We report means over five runs, each generating 40 graphs.

\subsubsection{Real-World Graph Generation}
\texttt{Enzymes} is a dataset of 587 protein graphs representing the tertiary structures of enzymes from the BRENDA database. The graphs have between 10 and 125 nodes. For consistency, we follow \citet{gdss} in using the Earth Mover's Distance (EMD) to compute the MMDs. We use 20\% of the data as the test set and the remaining 80\% as the training set, from which we additionally reserve 20\% for validation. MMDs are computed between the test set and a batch of generated graphs of the same size.

\subsection{Non-Markovian Continuous Diffusion}\label{ap:continuous_diffusion}

To further assess the role of the simplex parameterization itself, we implemented an additional baseline that uses the same non-Markovian resampling rule, but with a parameterization in unconstrained Euclidean space rather than on the simplex. Concretely, we define the probability path as
\begin{equation}
    \rvx_t = \alpha_t (2\rvx_1 - \vone) + (1-\alpha_t)\rvx_0,
\end{equation}
with $\rvx_1 \sim p_{\text{data}}$ and $\rvx_0 \sim \mathcal{N}(\vzero, \mI)$. We use a linear schedule, $\alpha_t = t$, which we found more effective. As in \textsc{UNSID}, we learn $P_{1\mid t}^\theta(\rvx_1 \mid \rvx_t)$ and denoise with Equation~\ref{eq:denoising}, where $q_{t+h}(\cdot \mid \rvx_1)$ is now the Gaussian path above.

\subsection{Baselines}\label{ap:baselines}

We report results for five representative methods: \textsc{DiGress} \citep{digress} implements standard discrete diffusion; \textsc{GruM} \citep{drum} performs denoising in a continuous space; \textsc{GraBFN} \citep{graphBFN} is a Bayesian-flow-network-based generative model; \textsc{SID} \citep{SID} is a non-Markovian discrete diffusion method; and \textsc{DeFoG} \citep{DeFog} is a discrete flow-matching model. In general, we report the numbers from the original papers, except for the \texttt{ZINC250K} results of \textsc{DiGress}, for which we use the values provided by \citet{drum}, as \textsc{DiGress} does not report results on this dataset. Finally, we also report metrics computed on random samples of the training set, each containing the same number of graphs as the generated samples (\textsc{Train.}).

For the \texttt{Planar} and \texttt{SBM} datasets, we reran the experiments for baselines that reported results from a \emph{single} sampling run. Below, we motivate this choice and describe our reproduction procedure.

\subsubsection{Motivation for Multiple Sampling Runs}

On unattributed graph datasets such as \texttt{Planar} and \texttt{SBM}, the test sets are relatively small, and sampling-based metrics exhibit high variance across runs. Moreover, evaluating many models, checkpoints, or seeds increases the risk of overfitting to the test set. Multiple sampling runs are therefore strongly recommended, and the corresponding variability should be assessed via standard deviations.

Several baselines report results from a \emph{single} sampling run, including \textsc{GruM}, \textsc{DiGress}, and \textsc{GraBFN}. These reported results even outperform samples drawn directly from the training set, which suggests overfitting to the test set. For this reason, we reran the experiments for \textsc{GruM} and \textsc{DiGress}. Unfortunately, the official \textsc{GraBFN} repository is currently empty, which prevented us from reproducing this baseline; we therefore do not report it on \texttt{Planar} and \texttt{SBM}.

\subsubsection{Procedure for Result Reproduction}

We were unable to reproduce the published results of either \textsc{GruM} or \textsc{DiGress}. Below, we describe the exact procedure we followed.

\paragraph{\textsc{GruM}.}
We used the pretrained models available in the official repository:
\url{https://github.com/harryjo97/GruM/tree/master/GruM_2D}.
We performed five sampling runs, varying only the sampling seed, and computed metrics using the code provided by the authors. We note that the default seed yields substantially better performance than the others.

\paragraph{\textsc{DiGress}.}
Since no pretrained models are currently available, we retrained \textsc{DiGress} using the official repository (\url{https://github.com/cvignac/DiGress/}) and the configurations specified in the corresponding YAML files. As specified in these configurations, we evaluated the model every 400 epochs and computed metrics on the validation set. For stability, we increased the number of graphs sampled at each evaluation during training to 40. We trained for 100{,}000 epochs on \texttt{Planar} and 20{,}000 epochs on \texttt{SBM}, and selected the checkpoint with the highest validation validity for final evaluation. Note that we fixed a minor bug in \verb|src/datasets/spectre_dataset.py|, where preprocessed graphs were added twice to their respective sets (lines 93 and 100).
\section{Additional Results}\label{ap:results}

In this section, we provide the following additional results:
\begin{itemize}
    \item model sizes and sampling times (Appendix~\ref{ap:time});
    \item results on \texttt{QM9H} (Appendix~\ref{ap:qm9h});
    \item ablations on the NFE, the noise schedule, and the prior (Appendix~\ref{ap:ablation});
    \item guidance (Appendix~\ref{ap:guidance_results});
    \item detailed results with standard deviations on \texttt{Planar}, \texttt{SBM}, \texttt{ZINC250K}, and \texttt{Enzymes} (Appendix~\ref{ap:detailed}).
\end{itemize}

\subsection{Model Sizes and Sampling Times}\label{ap:time}

We report the number of parameters of each model, as well as the wall-clock time required for 100 sampling iterations when generating 40 \texttt{Planar} graphs and 40 \texttt{SBM} graphs. All experiments were conducted on a single server with one GPU (NVIDIA GeForce RTX 3090) and 128 CPU cores.

Our model is the fastest on both datasets. Since Dirichlet flow matching (\textsc{DiriFM}) and \textsc{UNSID} share the same denoiser architecture, differences in sampling speed arise solely from the sampling procedure itself. Likewise, our implementation of SID (\textsc{NM-DD}) uses the same architecture as \textsc{UNSID}; the additional overhead of \textsc{NM-DD} is primarily due to the computation of the extra structural features required by the discrete denoiser.

\begin{table}[H]
    \centering
    \caption{Model sizes and sampling times (in seconds).}
    \label{tab:time_and_size}
    \begin{small}
    \begin{sc}
    \begin{tabular}{lcccccc}
    \toprule
        & \multicolumn{2}{c}{\textbf{Planar}} & \multicolumn{2}{c}{\textbf{SBM}} & \textbf{QM9H} & \textbf{ZINC250K} \\
        \cmidrule(lr){2-3} \cmidrule(lr){4-5} \cmidrule(lr){6-6} \cmidrule(lr){7-7}
        & Time$\downarrow$ & Params & Time$\downarrow$ & Params & Params & Params \\
    \midrule
        DiGress & $16.9 \pm 0.0$ & $8.9$M & $96.9 \pm 0.1$ & $7.1$M & $3.6$M & $8.2$M \\
        GruM    & $29.3 \pm 0.0$ & $7.1$M & $84.2 \pm 0.0$ & $7.1$M & -      & $8.2$M \\
    \midrule
        NM-DD   & $\phantom{0}9.9 \pm 0.4$ & $5.4$M & $84.1 \pm 0.3$ & $6.2$M & $6.2$M & $6.2$M \\
        DiriFM  & $\phantom{0}9.4 \pm 0.1$ & $5.4$M & $89.0 \pm 0.3$ & $6.2$M & $6.2$M & $6.2$M \\
    \midrule
        UNSID   & $\phantom{0}6.3 \pm 0.0$ & $5.4$M & $72.6 \pm 0.5$ & $6.2$M & $6.2$M & $6.2$M \\
    \bottomrule
    \end{tabular}
    \end{sc}
    \end{small}
\end{table}

For \textsc{DiGress} on \texttt{ZINC250K}, whose results we take from \citet{drum}, we assume that \citet{drum} used the same architecture as for \textsc{GruM}.

\subsection{Results on \texttt{QM9H}}\label{ap:qm9h}

\textsc{UNSID} outperforms all baselines on all metrics except uniqueness, where it remains competitive. For \textsc{DiGress}, we report the results from the original paper, which does not report FCD and NSPDK.

\begin{table}[H]
    \setlength{\tabcolsep}{2pt}
    \centering
    \caption{Results on \texttt{QM9H}. NSPDK is reported $\times 10^3$.}
    \label{tab:qm9H}
    \begin{small}
    \begin{sc}
    \begin{tabular}{lcccc}
    \toprule
        & Valid (\%)$\uparrow$ & FCD$\downarrow$ & NSPDK$\downarrow$ & Unique (\%)$\uparrow$ \\
    \midrule
        Train.  & $98.90 \pm 0.05$ & $0.062 \pm 0.002$ & $0.121 \pm 0.016$ & $99.81 \pm 0.04$ \\
        DiGress & $95.4\phantom{0} \pm 1.1\phantom{0}$ & - & - & $97.6\phantom{0} \pm 0.4\phantom{0}$ \\
    \midrule
        DiscDiff & $22.29 \pm 0.62$ & $4.246 \pm 0.454$ & $41.932 \pm 0.741\phantom{0}$ & $73.72 \pm 0.56$ \\
        NM-DD    & $97.97 \pm 0.07$ & $0.366 \pm 0.021$ & $1.149 \pm 0.047$ & $95.23 \pm 0.17$ \\
        NM-CD    & $31.29 \pm 0.15$ & $1.688 \pm 0.031$ & $22.675 \pm 0.099\phantom{0}$ & $91.45 \pm 0.27$ \\
        DiriFM   & $92.20 \pm 0.19$ & $0.356 \pm 0.029$ & $0.495 \pm 0.010$ & $97.53 \pm 0.13$ \\
    \midrule
        UNSID    & $98.87 \pm 0.07$ & $0.152 \pm 0.011$ & $0.487 \pm 0.068$ & $96.33 \pm 0.14$ \\
    \bottomrule
    \end{tabular}
    \end{sc}
    \end{small}
\end{table}

\subsection{Ablations}\label{ap:ablation}

We conduct all ablation studies on \texttt{ZINC250K}, as it is by far the largest dataset and has the largest test set, which yields more precise and robust evaluation metrics. This dataset is therefore best suited to detecting small effects.

\subsubsection{NFE}

When ablating the NFE, we observe that our model achieves state-of-the-art validity with as few as 64 denoising steps. FCD and NSPDK keep improving with more steps: our model surpasses the best baseline FCD from 512 steps on, and the best baseline NSPDK at 1024 steps. This confirms the efficiency and overall effectiveness of our approach. Mismatch with the main table comes from the single run evaluation used for this ablation.

\begin{table}[H]
    \centering
    \caption{Effect of the NFE on \texttt{ZINC250K}. NSPDK is reported $\times 10^3$.}
    \label{tab:nfe}
    \begin{small}
    \begin{sc}
    \begin{tabular}{lccc}
    \toprule
        NFE & Invalid (\%)$\downarrow$ & FCD$\downarrow$ & NSPDK$\downarrow$ \\
    \midrule
        $16$   & $5.71$ & $6.54$ & $7.97$ \\
        $32$   & $1.05$ & $4.32$ & $4.00$ \\
        $64$   & $0.36$ & $3.26$ & $2.79$ \\
        $128$  & $0.08$ & $2.63$ & $2.08$ \\
        $256$  & $0.09$ & $2.17$ & $1.59$ \\
        $512$  & $0.01$ & $1.92$ & $1.34$ \\
        $1024$ & $0.02$ & $1.71$ & $0.87$ \\
    \bottomrule
    \end{tabular}
    \end{sc}
    \end{small}
\end{table}

\subsubsection{Noise Schedule}\label{ap:sched}

We assess the effect of the noise schedule $\alpha_t = -a\log(1-t)$, which parameterizes our probability path $\mathrm{Dir}(\rvx_t; \vone + \alpha_t\rvx_1)$, by evaluating our model across a range of values of the hyperparameter $a$.

\begin{figure}[H]
    \begin{center}
    \includegraphics[width=0.45\textwidth]{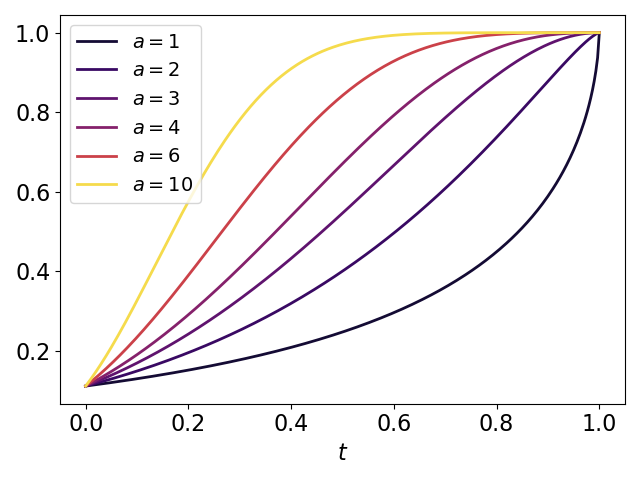}
    \includegraphics[width=0.45\textwidth]{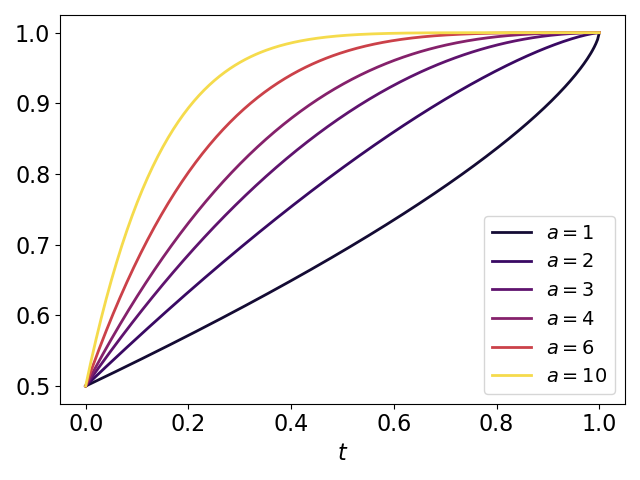}
    \end{center}
    \caption{Voronoi probabilities for various values of $a$ in the schedule $\alpha_t = -a\log(1-t)$, for $K = 9$ (left) and $K = 2$ (right).}
    \label{fig:noise_sched}
\end{figure}

Figure~\ref{fig:noise_sched} shows the corresponding Voronoi probabilities for each tested value of $a$. Since the Voronoi probability depends on the number of categories $K$, we report it separately for node attributes ($K = 9$) and edge attributes ($K = 4$).

Figure~\ref{fig:noise_sched} shows that low values of $a$ (in particular $a = 1$) induce a probability path that becomes too sharp near the end of the trajectory (i.e., as $t \to 1$), which makes the denoising task overly difficult in the final steps. Conversely, for large values of $a$, the probability path becomes too flat toward the end of the process, leaving the denoiser with little useful signal to correct.

The results in Table~\ref{tab:schedule} align with these expectations. When $a = 1$, performance degrades substantially. For large values of $a$, performance also decreases, but only marginally. Overall, the model is robust to variations in $a$: all tested values $a \geq 2$ outperform the baselines in FCD and NSPDK.

We draw two main conclusions from this experiment: (1) the Voronoi-probability analysis provides a reliable tool for calibrating the noise schedule, and (2) our model is robust to a wide range of schedules.

\begin{table}[H]
    \setlength{\tabcolsep}{2pt}
    \centering
    \caption{Results on \texttt{ZINC250K} for various values of the schedule hyperparameter $a$. NSPDK is reported $\times 10^3$.}
    \label{tab:schedule}
    \begin{small}
    \begin{sc}
    \begin{tabular}{lccccc}
    \toprule
        $a$ & Valid (\%)$\uparrow$ & FCD$\downarrow$ & NSPDK$\downarrow$ & Unique (\%)$\uparrow$ & Novel (\%)$\uparrow$ \\
    \midrule
        $1$  & $98.87 \pm 0.16$ & $4.13 \pm 0.08$ & $2.61 \pm 0.11$ & $99.86 \pm 0.01$ & $99.99 \pm 0.01$ \\
        $2$  & $99.89 \pm 0.03$ & $1.89 \pm 0.01$ & $0.98 \pm 0.06$ & $99.98 \pm 0.00$ & $99.96 \pm 0.01$ \\
        $3$  & $99.98 \pm 0.01$ & $1.79 \pm 0.03$ & $1.00 \pm 0.06$ & $100.00 \pm 0.00\phantom{0}$ & $99.97 \pm 0.01$ \\
        $4$  & $99.94 \pm 0.02$ & $1.82 \pm 0.05$ & $1.05 \pm 0.04$ & $99.99 \pm 0.00$ & $99.98 \pm 0.01$ \\
        $6$  & $99.92 \pm 0.02$ & $1.66 \pm 0.03$ & $0.82 \pm 0.03$ & $99.99 \pm 0.01$ & $99.97 \pm 0.02$ \\
        $10$ & $99.91 \pm 0.00$ & $1.95 \pm 0.03$ & $1.25 \pm 0.07$ & $99.99 \pm 0.01$ & $99.99 \pm 0.00$ \\
    \bottomrule
    \end{tabular}
    \end{sc}
    \end{small}
\end{table}

\subsubsection{Prior Choice}

\begin{figure}[H]
    \begin{center}
    \includegraphics[width=0.5\textwidth]{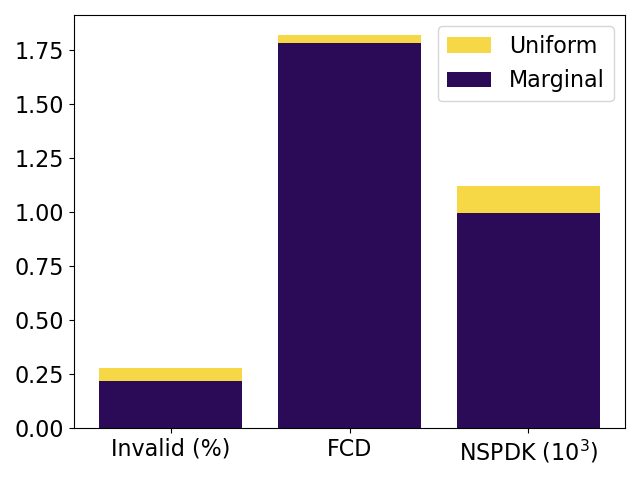}
    \end{center}
    \caption{Comparison between the uniform prior $\mathrm{Dir}(\vone)$ and the marginal-weighted mixture of Dirichlet distributions $q_0^{\mathrm{marg}}$ ($\kappa = 2$) on \texttt{ZINC250K}.}
    \label{fig:noise_compar}
\end{figure}

Compared with the uniform prior, the marginal-weighted mixture of Dirichlet distributions slightly but consistently improves sampling performance across all metrics (Figure~\ref{fig:noise_compar}).

\subsection{Guidance}\label{ap:guidance_results}

The sole purpose of this experiment is to show that our model supports classifier guidance without affecting its ability to generate valid molecules. As described in Section~\ref{sec:eval}, we target standardized QED values on \texttt{QM9H} and compute the MAE in standardized units.

\begin{table}[H]
    \centering
    \caption{Effect of classifier guidance on \texttt{QM9H}. The MAE between target and generated QED values is computed in standardized units.}
    \label{tab:guidance}
    \begin{small}
    \begin{sc}
    \begin{tabular}{lcc}
    \toprule
        & Valid (\%)$\uparrow$ & MAE$\downarrow$ \\
    \midrule
        Unconditional       & $99.6$ & $1.06$ \\
        Classifier guidance & $99.5$ & $0.42$ \\
    \bottomrule
    \end{tabular}
    \end{sc}
    \end{small}
\end{table}

\subsection{Detailed Results}\label{ap:detailed}

We report means and standard deviations over five sampling runs on \texttt{Planar} (Table~\ref{tab:planar_ap}), \texttt{SBM} (Table~\ref{tab:sbm_ap}), \texttt{ZINC250K} (Table~\ref{tab:zinc_ap}), and \texttt{Enzymes} (Table~\ref{tab:ap:enzymes}).

\begin{table}[H]
    \setlength{\tabcolsep}{2pt}
    \centering
    \caption{Results on \texttt{Planar}. MMDs are reported $\times 10^3$.}
    \label{tab:planar_ap}
    \begin{small}
    \begin{sc}
    \begin{tabular}{lccccc}
    \toprule
        & Valid (\%)$\uparrow$ & Deg.$\downarrow$ & Clust.$\downarrow$ & Orbit$\downarrow$ & Spect.$\downarrow$ \\
    \midrule
        Train.  & $100.0 \pm 0.0$ & $0.25 \pm 0.18$ & $36.6 \pm 3.5$ & $0.78 \pm 0.37$ & $6.62 \pm 1.34$ \\
        DiGress & $45.5 \pm 7.0$ & $0.71 \pm 0.40$ & $\phantom{0}45.4 \pm 13.6$ & $1.31 \pm 0.78$ & $8.40 \pm 1.16$ \\
        GruM    & $91.0 \pm 2.6$ & $0.38 \pm 0.15$ & $40.7 \pm 4.8$ & $6.42 \pm 1.96$ & $7.87 \pm 1.17$ \\
        SID     & $91.3 \pm 4.1$ & $5.93 \pm 1.26$ & $163.4 \pm 31.9$ & $19.1\phantom{0} \pm 4.14\phantom{0}$ & $7.62 \pm 1.34$ \\
        DeFoG   & $99.5 \pm 1.0$ & $0.5\phantom{0} \pm 0.2\phantom{0}$ & $\phantom{0}50.1 \pm 14.9$ & $0.6\phantom{0} \pm 0.4\phantom{0}$ & $7.2\phantom{0} \pm 1.1\phantom{0}$ \\
    \midrule
        DiscDiff & $\phantom{0}0.0 \pm 0.0$ & $56.1\phantom{0} \pm 3.70\phantom{0}$ & $294.0 \pm 3.4\phantom{0}$ & $1410.0\phantom{0} \pm 44.5\phantom{000}$ & $85.1\phantom{0} \pm 1.6\phantom{00}$ \\
        NM-DD    & $98.0 \pm 1.0$ & $14.1\phantom{0} \pm 2.58\phantom{0}$ & $363.0 \pm 45.6$ & $27.2\phantom{0} \pm 5.6\phantom{00}$ & $6.9\phantom{0} \pm 0.8\phantom{0}$ \\
        NM-CD    & $\phantom{0}0.0 \pm 0.0$ & $1.23 \pm 0.50$ & $226.3 \pm 8.4\phantom{0}$ & $45.77 \pm 6.20\phantom{0}$ & $21.96 \pm 0.74\phantom{0}$ \\
        DiriFM   & $\phantom{0}0.0 \pm 0.0$ & $6.44 \pm 1.42$ & $196.6 \pm 22.5$ & $45.04 \pm 5.83\phantom{0}$ & $21.76 \pm 1.30\phantom{0}$ \\
    \midrule
        UNSID    & $100.0 \pm 0.0$ & $0.36 \pm 0.24$ & $39.9 \pm 8.7$ & $0.78 \pm 0.49$ & $7.12 \pm 0.45$ \\
    \bottomrule
    \end{tabular}
    \end{sc}
    \end{small}
\end{table}

\begin{table}[H]
    \setlength{\tabcolsep}{2pt}
    \centering
    \caption{Results on \texttt{SBM}. MMDs are reported $\times 10^3$.}
    \label{tab:sbm_ap}
    \begin{small}
    \begin{sc}
    \begin{tabular}{lcccccc}
    \toprule
        & Valid (\%)$\uparrow$ & Deg.$\downarrow$ & Clust.$\downarrow$ & Orbit$\downarrow$ & Spect.$\downarrow$ & Novel (\%)$\uparrow$ \\
    \midrule
        Train.  & $93.50 \pm 2.00$ & $1.57 \pm 0.55$ & $50.11 \pm 0.51$ & $37.0 \pm 10.7$ & $4.58 \pm 0.49$ & $100$ \\
        DiGress & $51.00 \pm 9.62$ & $1.28 \pm 0.48$ & $51.49 \pm 1.31$ & $39.6 \pm 7.7$ & $5.04 \pm 0.63$ & $100.0 \pm 0.0$ \\
        GruM    & $67.54 \pm 2.98$ & $2.20 \pm 0.76$ & $49.88 \pm 0.62$ & $40.4 \pm 5.6$ & $5.06 \pm 0.72$ & $100.0 \pm 0.0$ \\
        SID     & $63.5\phantom{0} \pm 3.7\phantom{0}$ & $11.5\phantom{0} \pm 2.7\phantom{0}$ & $51.4 \pm 1.5$ & $123.1 \pm 5.4\phantom{0}$ & $5.93 \pm 1.18$ & $100.0 \pm 0.0$ \\
        DeFoG   & $90.0\phantom{0} \pm 5.2\phantom{0}$ & $0.6\phantom{0} \pm 2.3\phantom{0}$ & $51.7 \pm 1.2$ & $\phantom{0}55.6 \pm 73.9$ & $5.40 \pm 1.20$ & $90.0 \pm 5.1$ \\
    \midrule
        DiscDiff & $\phantom{0}0.00 \pm 0.00$ & $1.82 \pm 0.88$ & $86.38 \pm 6.37$ & $125.7 \pm 3.3\phantom{0}$ & $11.28 \pm 1.06\phantom{0}$ & $100.0 \pm 0.0$ \\
        NM-DD    & $60.50 \pm 4.30$ & $4.38 \pm 1.65$ & $50.92 \pm 0.89$ & $52.9 \pm 6.3$ & $5.70 \pm 0.30$ & $100.0 \pm 0.0$ \\
        NM-CD    & $57.50 \pm 5.40$ & $1.89 \pm 0.50$ & $50.29 \pm 0.45$ & $41.08 \pm 1.00$ & $5.38 \pm 1.30$ & $100.0 \pm 0.0$ \\
        DiriFM   & $46.00 \pm 4.36$ & $4.26 \pm 0.54$ & $53.04 \pm 1.05$ & $53.0 \pm 8.9$ & $5.02 \pm 1.14$ & $100.0 \pm 0.0$ \\
    \midrule
        UNSID    & $78.50 \pm 4.64$ & $1.74 \pm 0.60$ & $49.94 \pm 1.07$ & $52.1 \pm 1.1$ & $5.90 \pm 1.05$ & $100.0 \pm 0.0$ \\
    \bottomrule
    \end{tabular}
    \end{sc}
    \end{small}
\end{table}

\begin{table}[H]
    \setlength{\tabcolsep}{2pt}
    \centering
    \caption{Results on \texttt{ZINC250K}. NSPDK is reported $\times 10^3$.}
    \label{tab:zinc_ap}
    \begin{small}
    \begin{sc}
    \begin{tabular}{lccccc}
    \toprule
        & Valid (\%)$\uparrow$ & FCD$\downarrow$ & NSPDK$\downarrow$ & Unique (\%)$\uparrow$ & Novel (\%)$\uparrow$ \\
    \midrule
        Train.  & $100.00 \pm 0.00$ & $1.13 \pm 0.01$ & $0.10 \pm 0.00$ & $99.97 \pm 0.02$ & - \\
        DiGress & $94.99$ & $3.48$ & $2.1$ & - & - \\
        GruM    & $98.65$ & $2.26$ & $1.5$ & - & - \\
        GraBFN  & $99.22$ & $2.12$ & $1.3$ & - & - \\
        SID     & $99.50 \pm 0.06$ & $2.06 \pm 0.05$ & $2.01 \pm 0.01$ & $99.84 \pm 0.02$ & $99.97 \pm 0.00$ \\
    \midrule
        DiscDiff & $74.17 \pm 0.65$ & $4.78 \pm 0.14$ & $4.08 \pm 0.13$ & $100.00 \pm 0.00\phantom{0}$ & $100.00 \pm 0.00\phantom{0}$ \\
        NM-DD    & $99.92 \pm 0.01$ & $2.65 \pm 0.06$ & $3.48 \pm 0.06$ & $99.88 \pm 0.04$ & $99.95 \pm 0.03$ \\
        NM-CD    & $57.96 \pm 0.16$ & $9.51 \pm 0.04$ & $13.65 \pm 0.07\phantom{0}$ & $99.92 \pm 0.02$ & $99.99 \pm 0.00$ \\
        DiriFM   & $97.32 \pm 0.12$ & $2.79 \pm 0.01$ & $1.92 \pm 0.07$ & $99.99 \pm 0.01$ & $99.99 \pm 0.01$ \\
    \midrule
        UNSID    & $99.98 \pm 0.01$ & $1.79 \pm 0.03$ & $1.00 \pm 0.06$ & $100.00 \pm 0.00\phantom{0}$ & $99.97 \pm 0.01$ \\
    \bottomrule
    \end{tabular}
    \end{sc}
    \end{small}
\end{table}

\begin{table}[H]
    \setlength{\tabcolsep}{2pt}
    \centering
    \caption{Results on \texttt{Enzymes}. MMDs are reported $\times 10^3$.}
    \label{tab:ap:enzymes}
    \begin{small}
    \begin{sc}
    \begin{tabular}{lcccc}
    \toprule
        & Deg.$\downarrow$ & Clust.$\downarrow$ & Orbit$\downarrow$ & Spect.$\downarrow$ \\
    \midrule
        Train.   & $3.3 \pm 0.3$ & $19.2 \pm 4.2$ & $0.9 \pm 0.3$ & $18.9 \pm 0.1$ \\
        GraphVAE & $1369 \pm 20$ & $629 \pm 5$ & $191 \pm 20$ & - \\
        GraphRNN & $17 \pm 7$ & $62 \pm 20$ & $46 \pm 31$ & - \\
        GDSS     & $26 \pm 8$ & $61 \pm 10$ & $9 \pm 5$ & - \\
    \midrule
        DiscDiff & $212.7 \pm 11.8$ & $631.1 \pm 45.7$ & $515.5 \pm 54.1$ & $19.6 \pm 0.2$ \\
        NM-DD    & $13.6 \pm 4.1$ & $39.9 \pm 5.2$ & $10.4 \pm 0.9$ & $17.4 \pm 0.0$ \\
        NM-CD    & $84.5 \pm 14.8$ & $307.9 \pm 29.9$ & $125.4 \pm 16.0$ & $20.2 \pm 0.2$ \\
        DiriFM   & $7.4 \pm 2.4$ & $100.6 \pm 17.5$ & $6.5 \pm 2.7$ & $19.8 \pm 0.2$ \\
    \midrule
        UNSID    & $10.2 \pm 4.2$ & $52.0 \pm 15.9$ & $2.7 \pm 1.1$ & $19.5 \pm 0.0$ \\
    \bottomrule
    \end{tabular}
    \end{sc}
    \end{small}
\end{table}
\section{Visualizations}\label{ap:visual}

We present here some visualizations of the generated and reference graphs.

\begin{figure}[H]
    \centering
    \caption{\texttt{Qm9H}}
    \begin{sc}
    \begin{tabular}{|ccc|ccc|}
    \hline
      \multicolumn{3}{|c|}{Generated molecules}   &
      \multicolumn{3}{|c|}{Real molecules}
     \\
    \hline
           
        \includegraphics[width=0.12\textwidth]{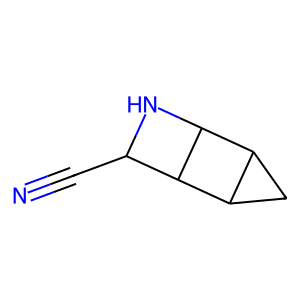} &
        \includegraphics[width=0.12\textwidth]{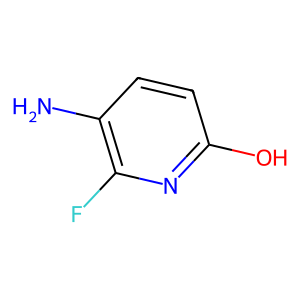} &
        \includegraphics[width=0.12\textwidth]{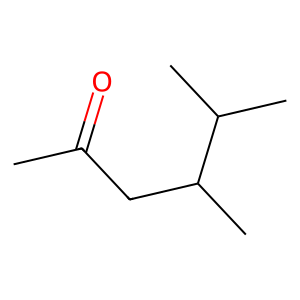} &
        \includegraphics[width=0.12\textwidth]{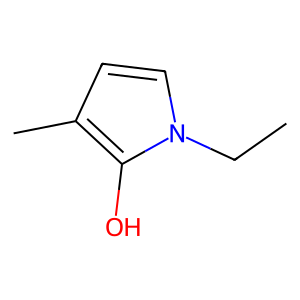} &
        \includegraphics[width=0.12\textwidth]{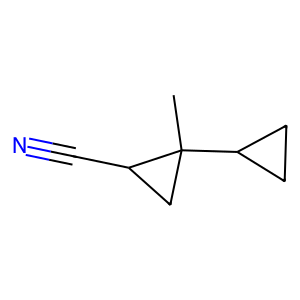}&
        \includegraphics[width=0.12\textwidth]{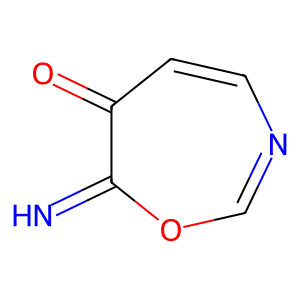} 
        \\
        \includegraphics[width=0.12\textwidth]{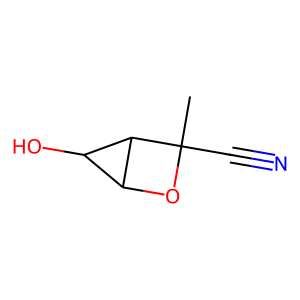} &
        \includegraphics[width=0.12\textwidth]{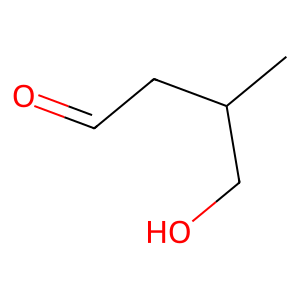} &
        \includegraphics[width=0.12\textwidth]{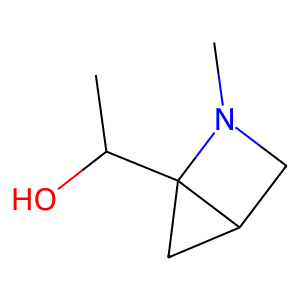} &
        \includegraphics[width=0.12\textwidth]{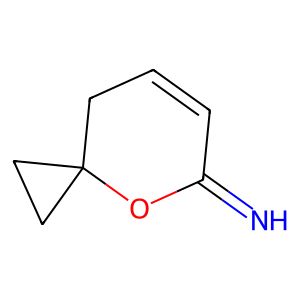} &
        \includegraphics[width=0.12\textwidth]{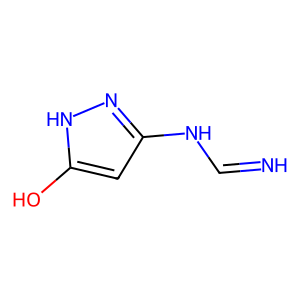}&
        \includegraphics[width=0.12\textwidth]{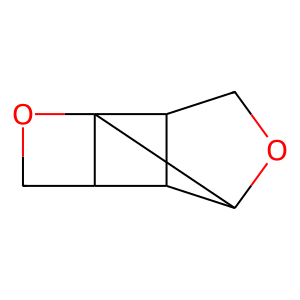} 
        \\
        \includegraphics[width=0.12\textwidth]{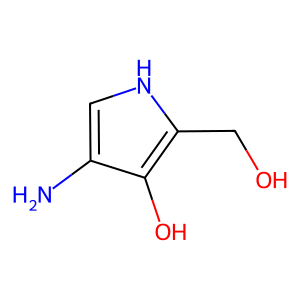} &
        \includegraphics[width=0.12\textwidth]{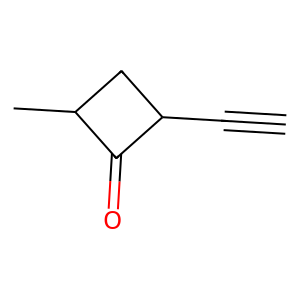} &
        \includegraphics[width=0.12\textwidth]{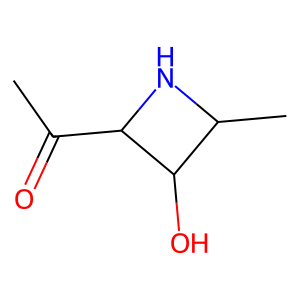} &
        \includegraphics[width=0.12\textwidth]{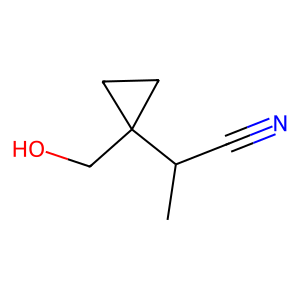} &
        \includegraphics[width=0.12\textwidth]{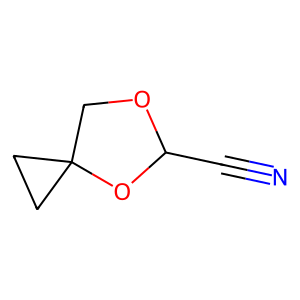}&
        \includegraphics[width=0.12\textwidth]{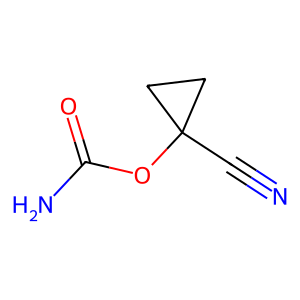} 
        \\
        \includegraphics[width=0.12\textwidth]{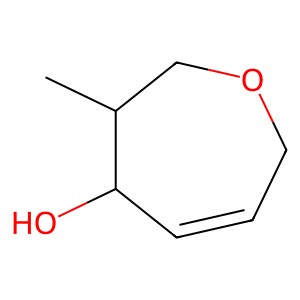} &
        \includegraphics[width=0.12\textwidth]{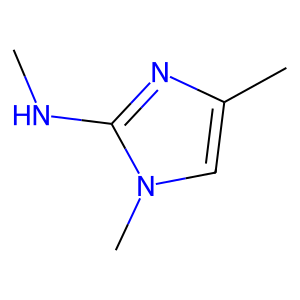} &
        \includegraphics[width=0.12\textwidth]{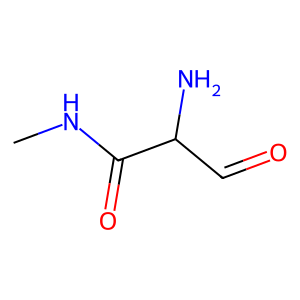} &
        \includegraphics[width=0.12\textwidth]{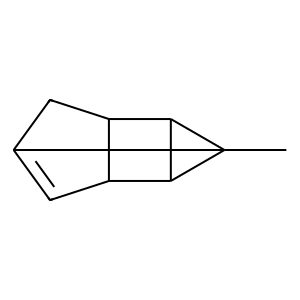} &
        \includegraphics[width=0.12\textwidth]{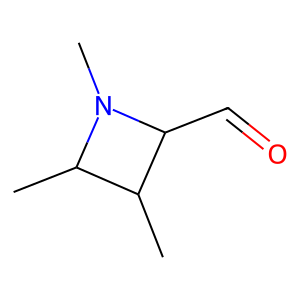}&
        \includegraphics[width=0.12\textwidth]{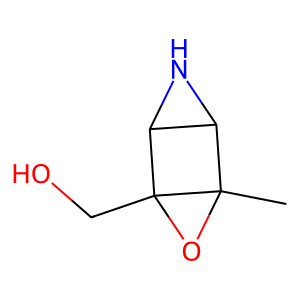} 
        \\
        \hline
        \end{tabular}
        \end{sc}
\end{figure}

\begin{figure}[ht!]
    \centering
    \caption{\texttt{ZINC250K}}
    \begin{sc}
    \begin{tabular}{|ccc|ccc|}
    \hline
      \multicolumn{3}{|c|}{Generated molecules}   &
      \multicolumn{3}{|c|}{Real molecules}
     \\
    \hline
           
        \includegraphics[width=0.12\textwidth]{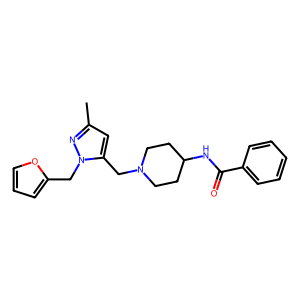} &
        \includegraphics[width=0.12\textwidth]{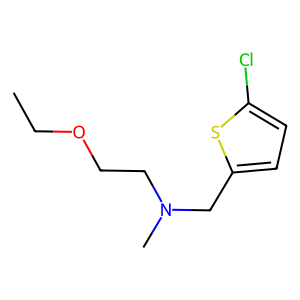} &
        \includegraphics[width=0.12\textwidth]{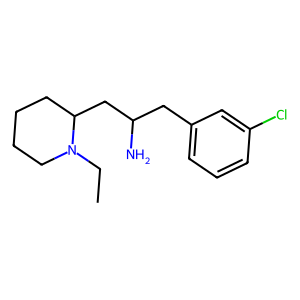} &
        \includegraphics[width=0.12\textwidth]{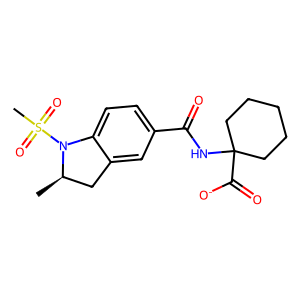} &
        \includegraphics[width=0.12\textwidth]{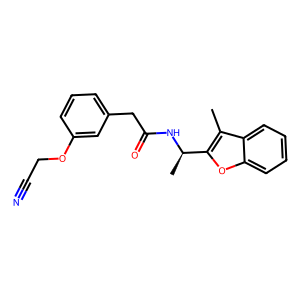}&
        \includegraphics[width=0.12\textwidth]{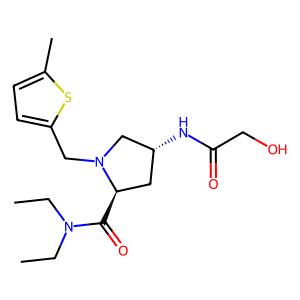} 
        \\
        \includegraphics[width=0.12\textwidth]{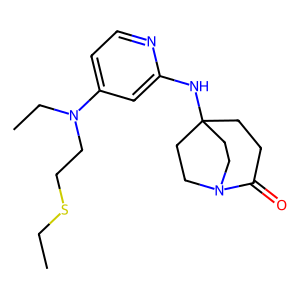} &
        \includegraphics[width=0.12\textwidth]{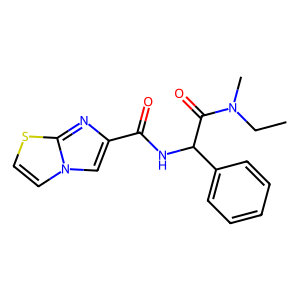} &
        \includegraphics[width=0.12\textwidth]{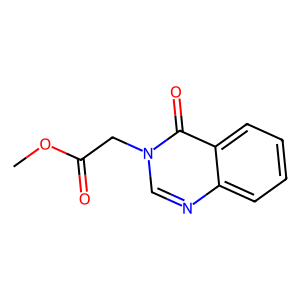} &
        \includegraphics[width=0.12\textwidth]{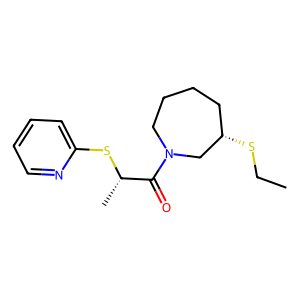} &
        \includegraphics[width=0.12\textwidth]{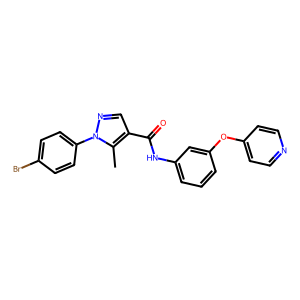}&
        \includegraphics[width=0.12\textwidth]{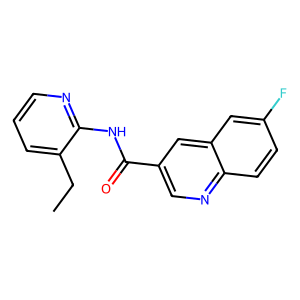} 
        \\
        \includegraphics[width=0.12\textwidth]{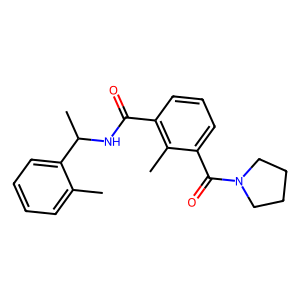} &
        \includegraphics[width=0.12\textwidth]{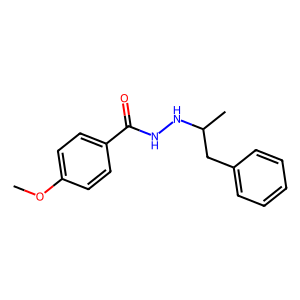} &
        \includegraphics[width=0.12\textwidth]{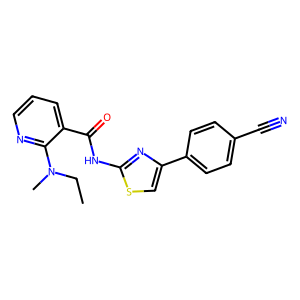} &
        \includegraphics[width=0.12\textwidth]{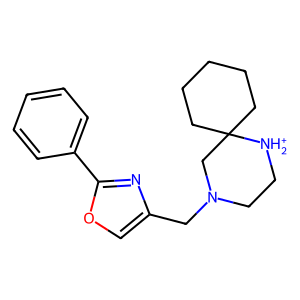} &
        \includegraphics[width=0.12\textwidth]{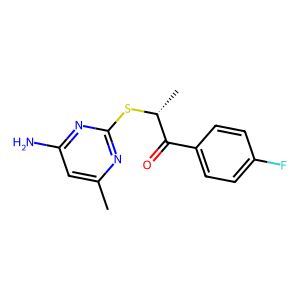}&
        \includegraphics[width=0.12\textwidth]{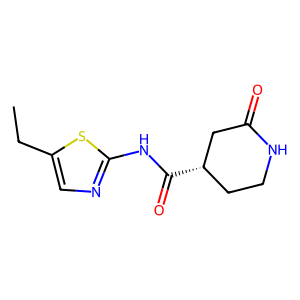} 
        \\
        \includegraphics[width=0.12\textwidth]{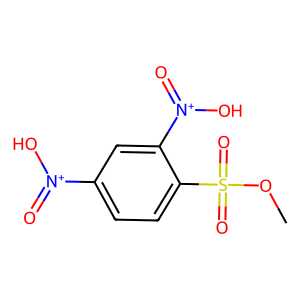} &
        \includegraphics[width=0.12\textwidth]{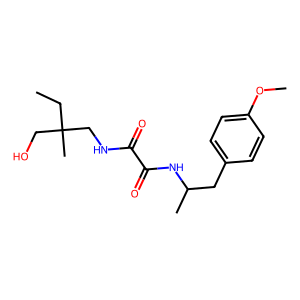} &
        \includegraphics[width=0.12\textwidth]{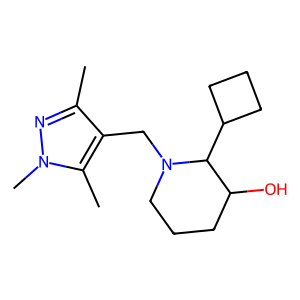} &
        \includegraphics[width=0.12\textwidth]{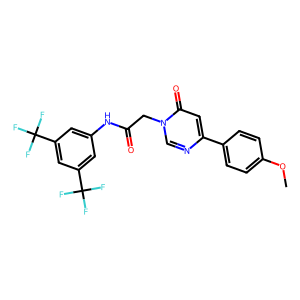} &
        \includegraphics[width=0.12\textwidth]{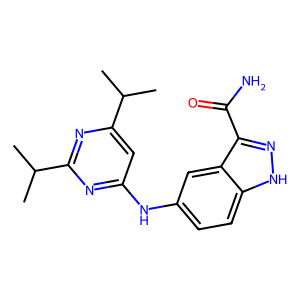}&
        \includegraphics[width=0.12\textwidth]{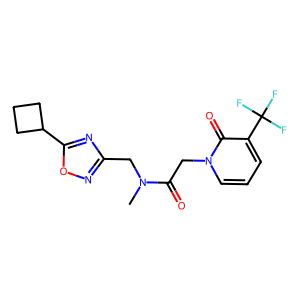} 
        \\
        \hline
        \end{tabular}
        \end{sc}
\end{figure}

\begin{figure}[ht!]
    \centering
    \caption{\texttt{Planar}}
    \begin{sc}
    \begin{tabular}{|ccc|ccc|}
    \hline
      \multicolumn{3}{|c|}{Generated graphs}   &
      \multicolumn{3}{|c|}{Real graphs}
     \\
    \hline
           
        \includegraphics[width=0.15\textwidth]{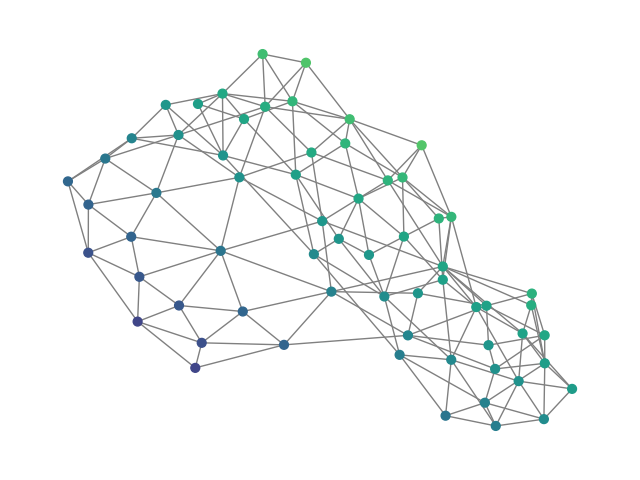} &
        \includegraphics[width=0.15\textwidth]{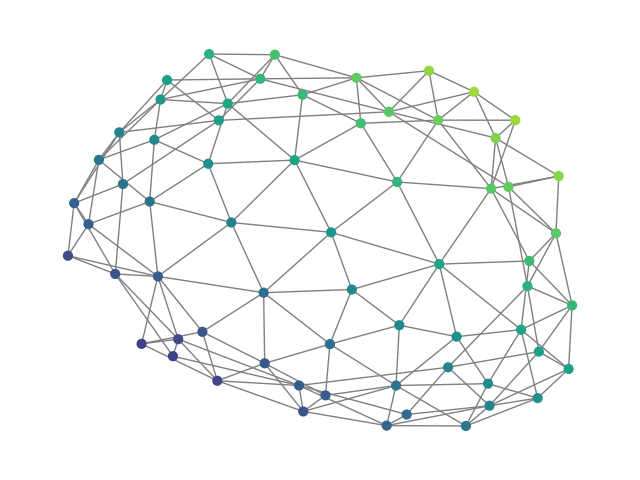} &
        \includegraphics[width=0.15\textwidth]{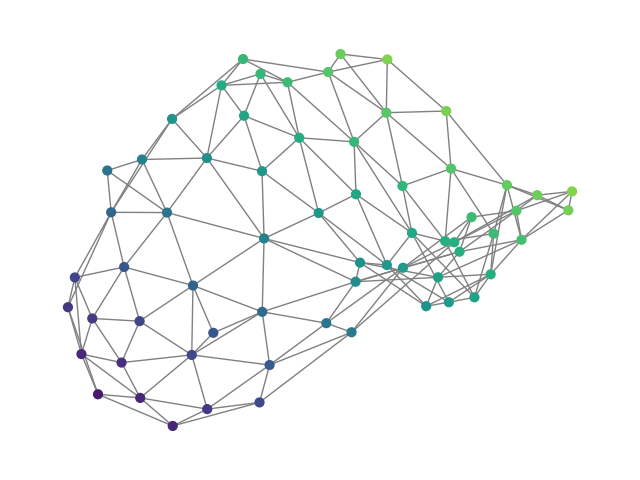} &
        \includegraphics[width=0.15\textwidth]{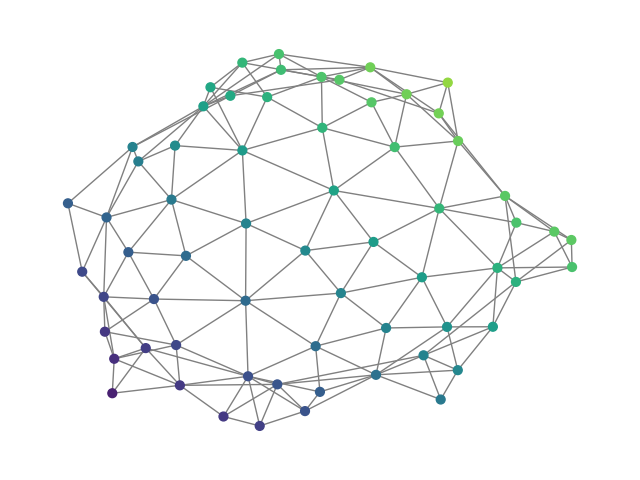} &
        \includegraphics[width=0.15\textwidth]{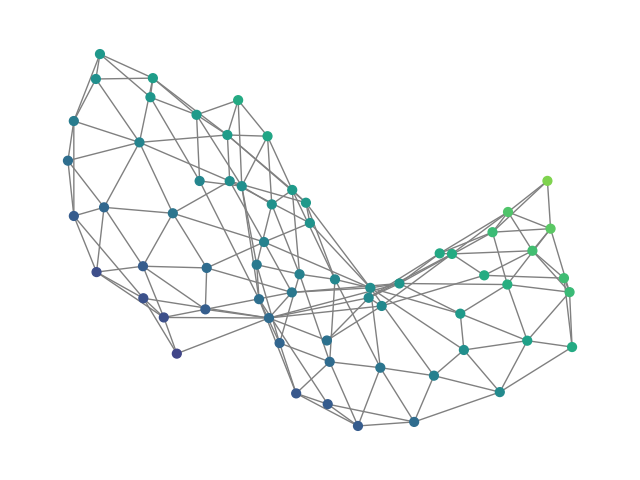}&
        \includegraphics[width=0.15\textwidth]{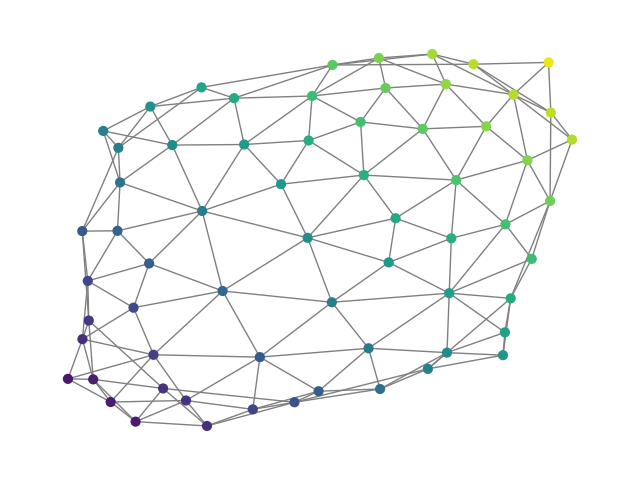} 
        \\
        \includegraphics[width=0.15\textwidth]{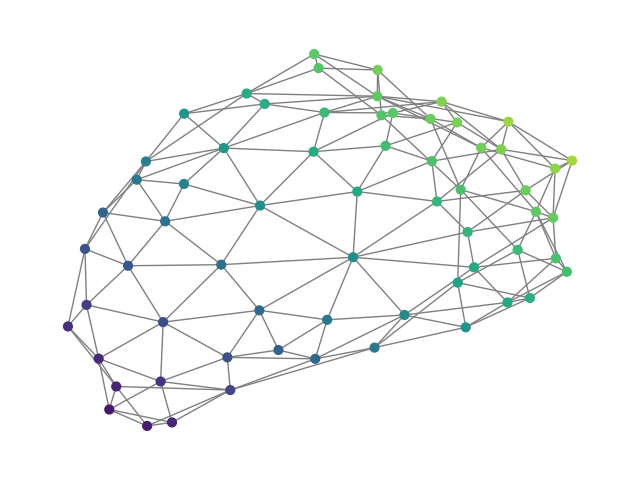} &
        \includegraphics[width=0.15\textwidth]{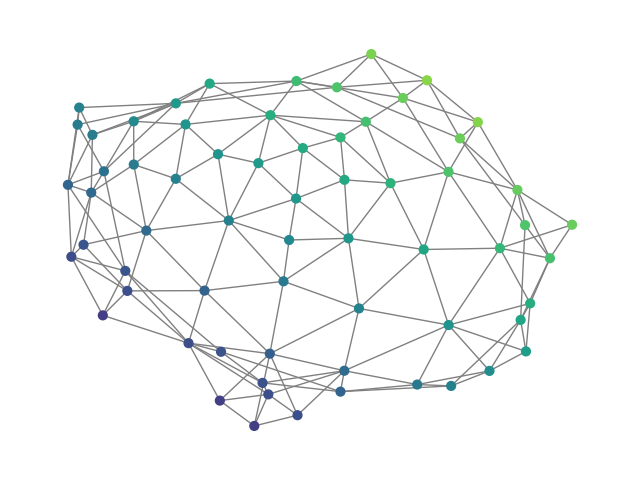} &
        \includegraphics[width=0.15\textwidth]{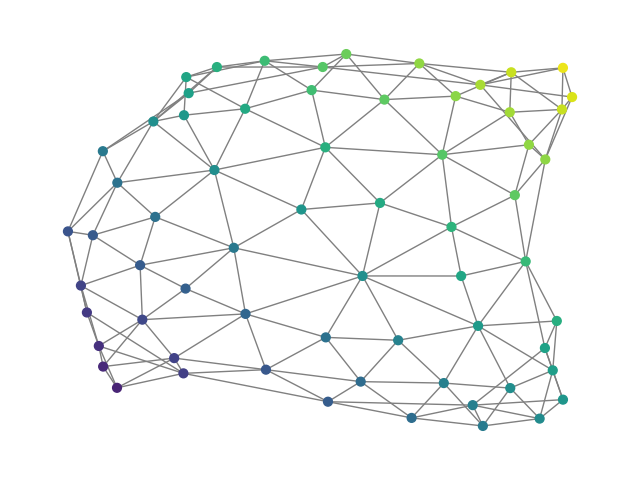} &
        \includegraphics[width=0.15\textwidth]{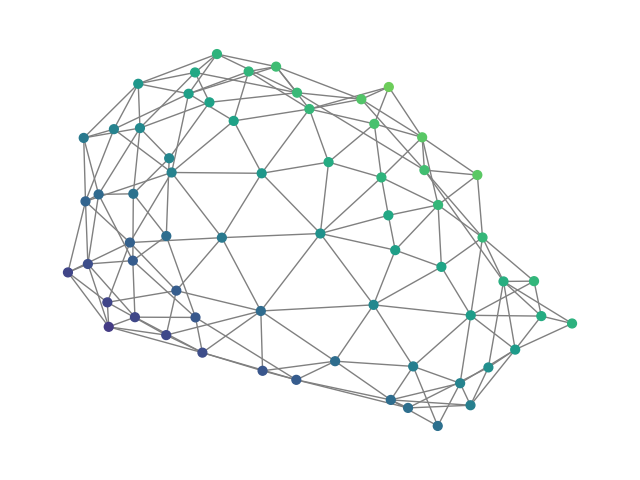} &
        \includegraphics[width=0.15\textwidth]{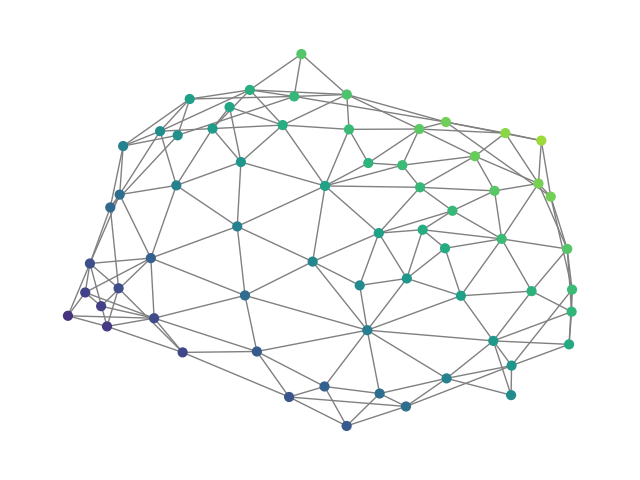}&
        \includegraphics[width=0.15\textwidth]{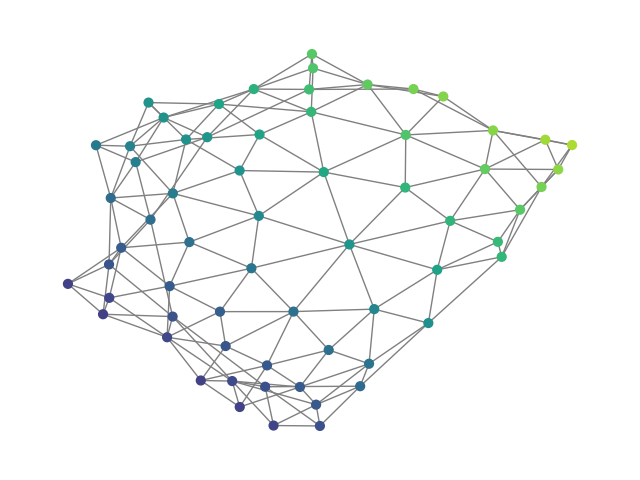} 
        \\
        \includegraphics[width=0.15\textwidth]{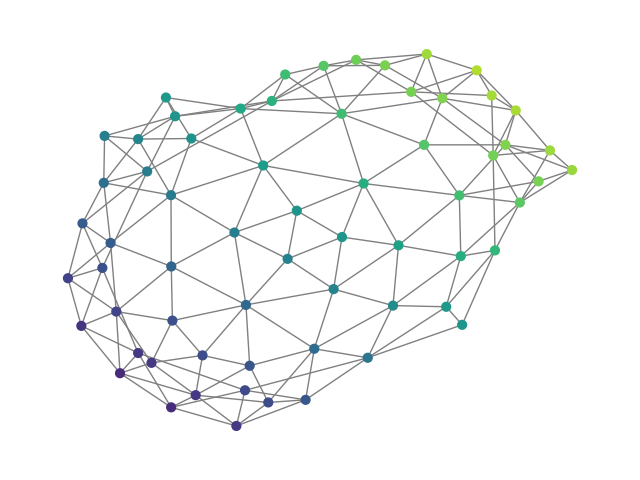} &
        \includegraphics[width=0.15\textwidth]{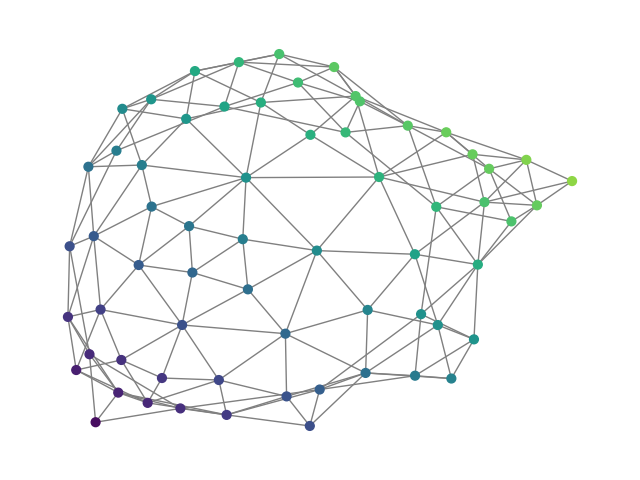} &
        \includegraphics[width=0.15\textwidth]{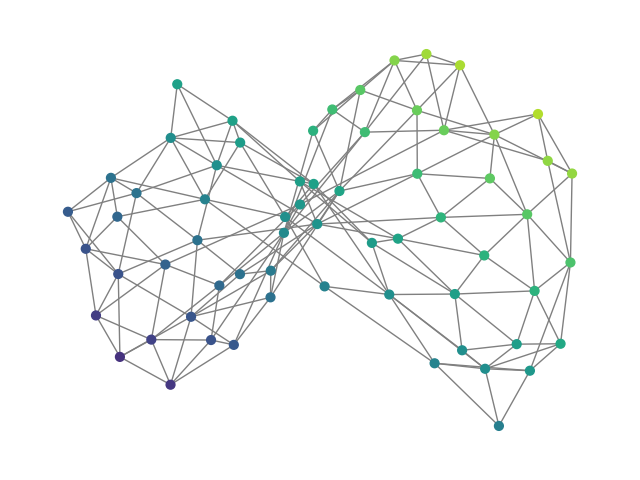} &
        \includegraphics[width=0.15\textwidth]{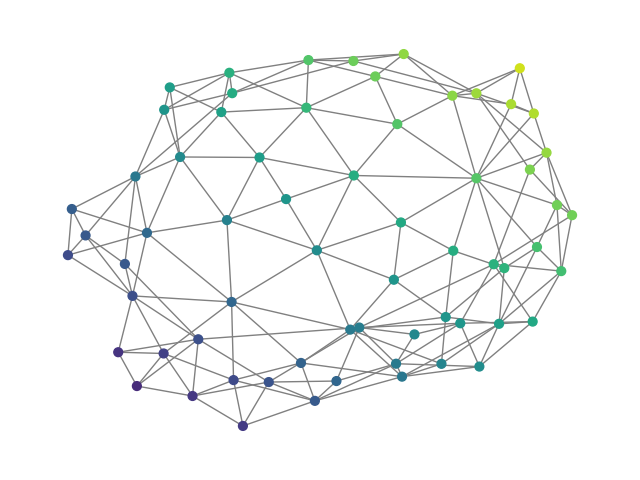} &
        \includegraphics[width=0.15\textwidth]{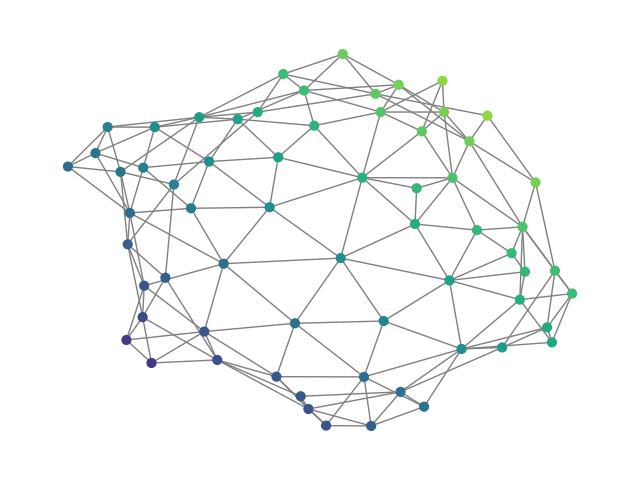}&
        \includegraphics[width=0.15\textwidth]{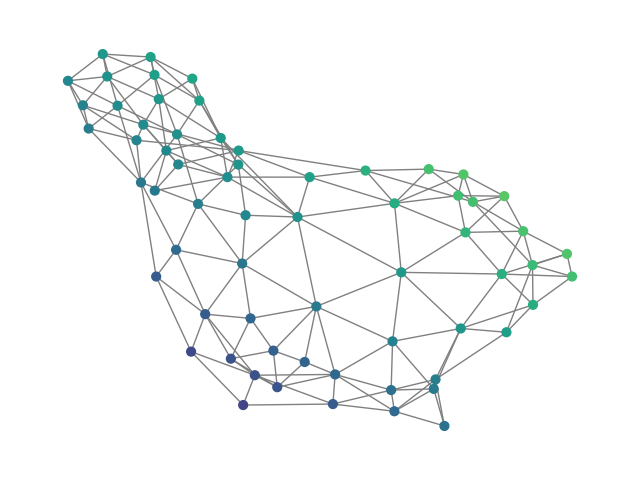} 
        \\
        \includegraphics[width=0.15\textwidth]{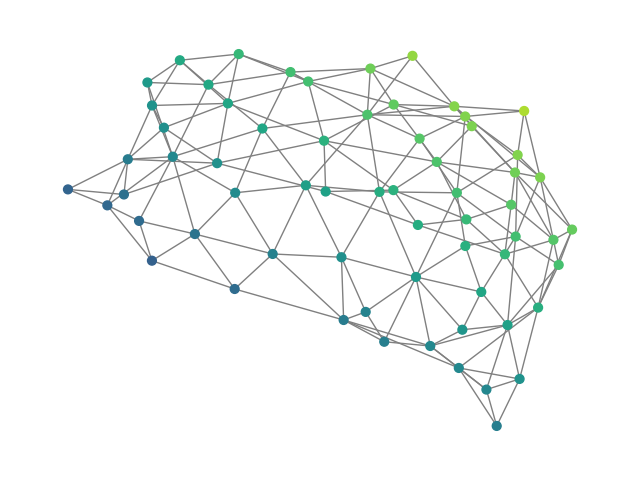} &
        \includegraphics[width=0.15\textwidth]{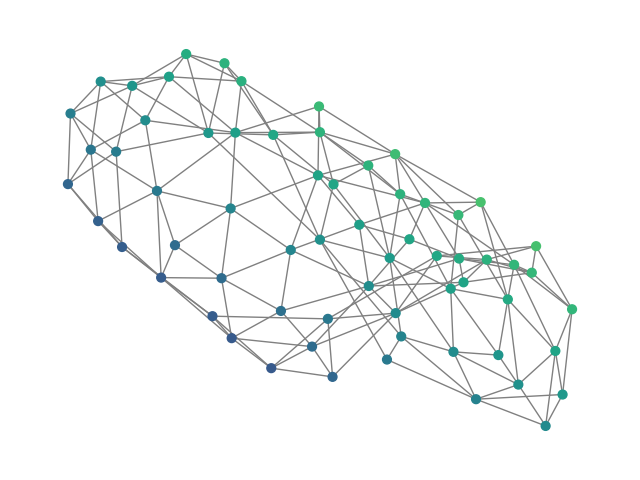} &
        \includegraphics[width=0.15\textwidth]{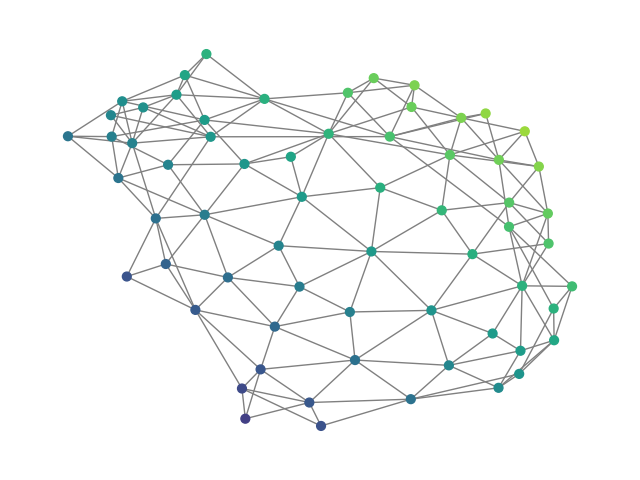} &
        \includegraphics[width=0.15\textwidth]{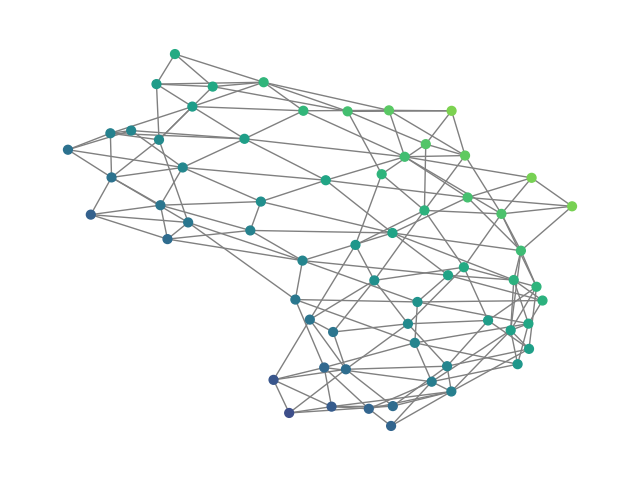} &
        \includegraphics[width=0.15\textwidth]{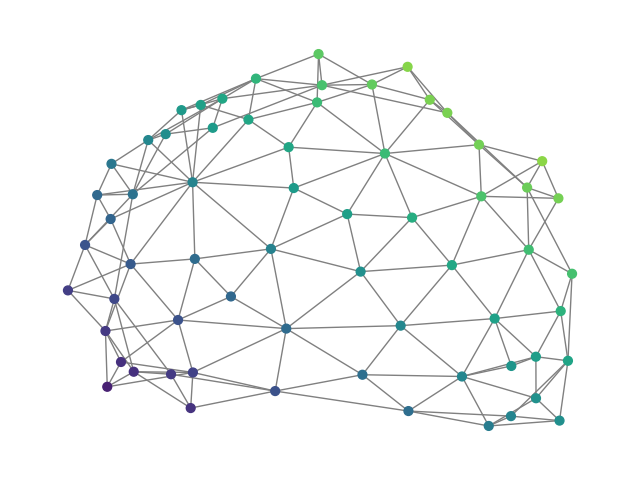}&
        \includegraphics[width=0.15\textwidth]{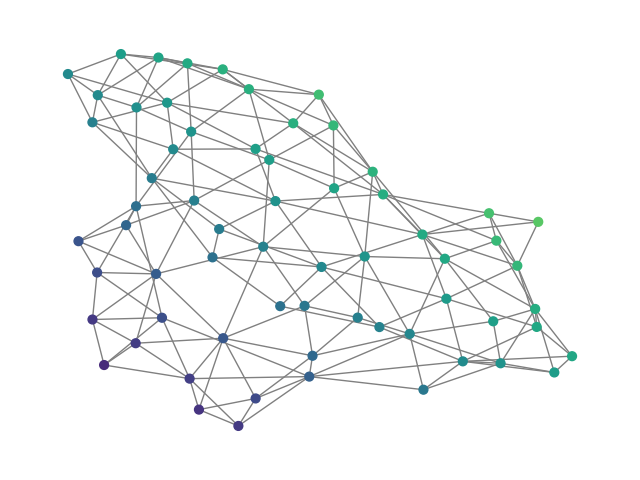} 
        \\
        \hline
        \end{tabular}
        \end{sc}
\end{figure}

\begin{figure}[ht!]
    \centering
    \caption{\texttt{Stochstic Block Model}}
    \begin{sc}
    \begin{tabular}{|ccc|ccc|}
    \hline
      \multicolumn{3}{|c|}{Generated graphs}   &
      \multicolumn{3}{|c|}{Real graphs}
     \\
    \hline
           
        \includegraphics[width=0.15\textwidth]{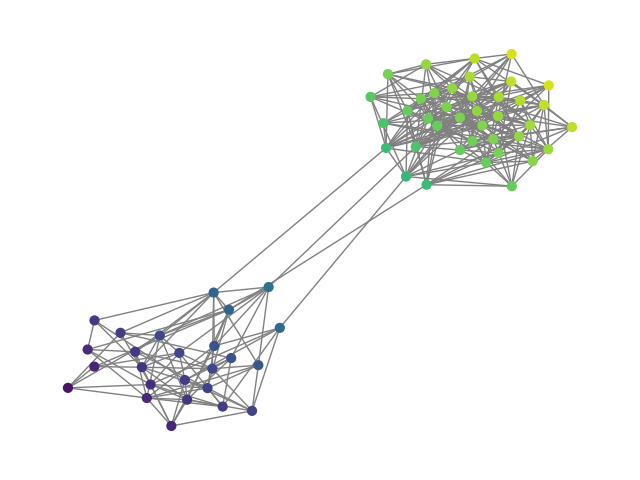} &
        \includegraphics[width=0.15\textwidth]{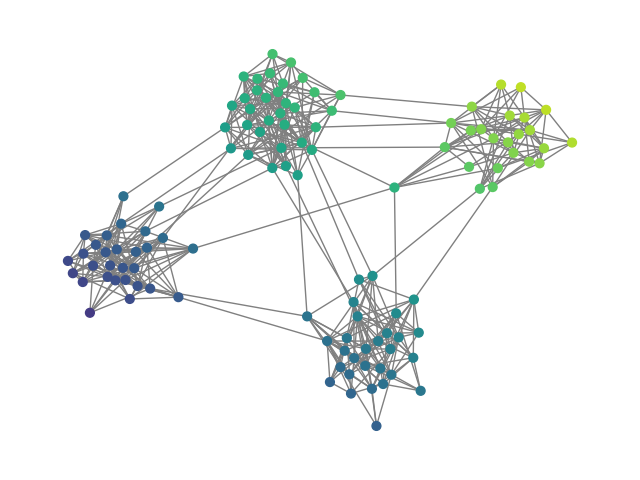} &
        \includegraphics[width=0.15\textwidth]{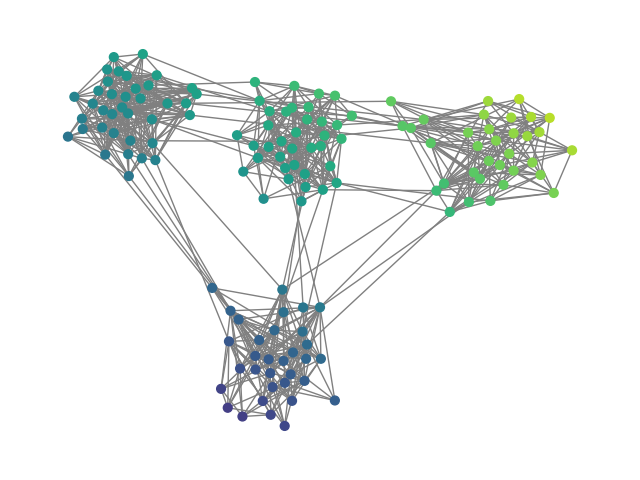} &
        \includegraphics[width=0.15\textwidth]{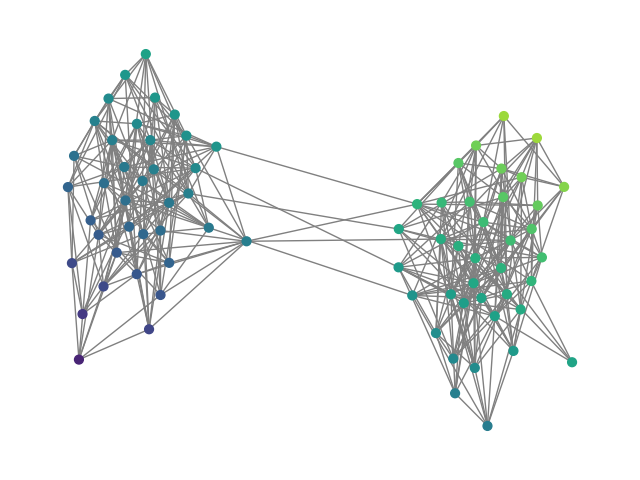} &
        \includegraphics[width=0.15\textwidth]{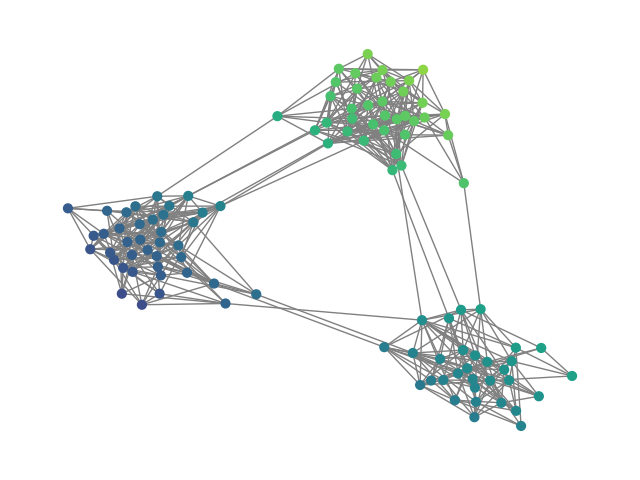}&
        \includegraphics[width=0.15\textwidth]{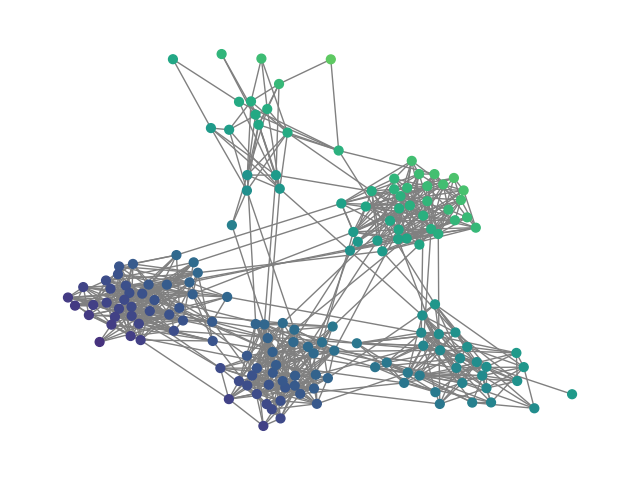} 
        \\
        \includegraphics[width=0.15\textwidth]{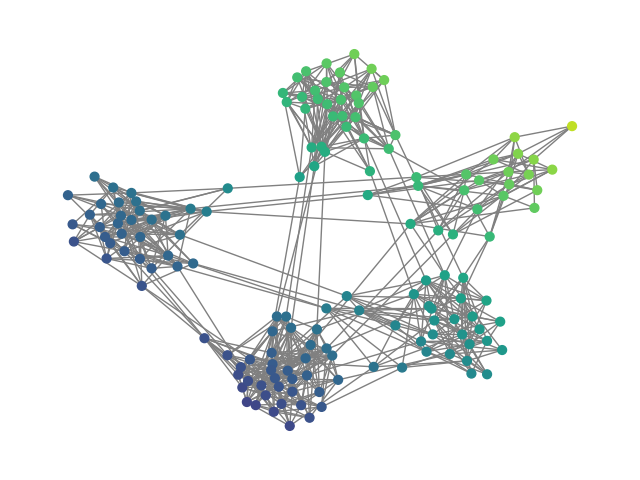} &
        \includegraphics[width=0.15\textwidth]{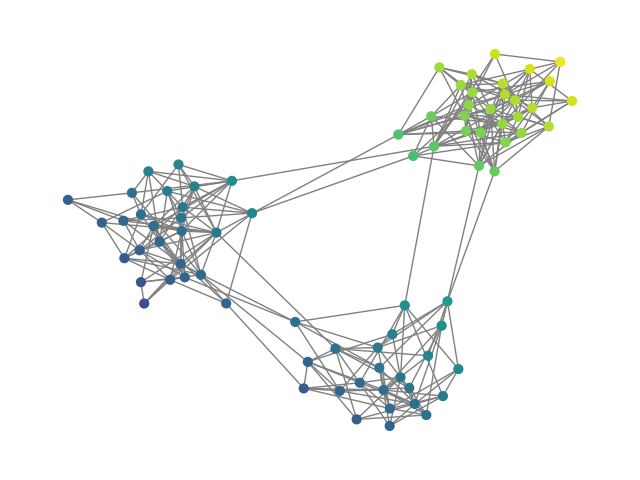} &
        \includegraphics[width=0.15\textwidth]{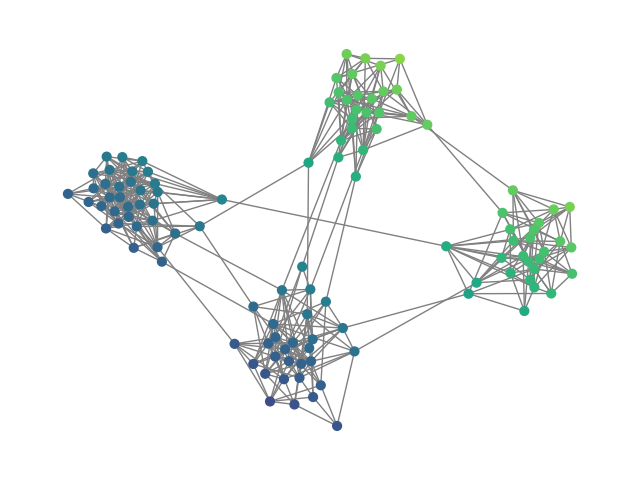} &
        \includegraphics[width=0.15\textwidth]{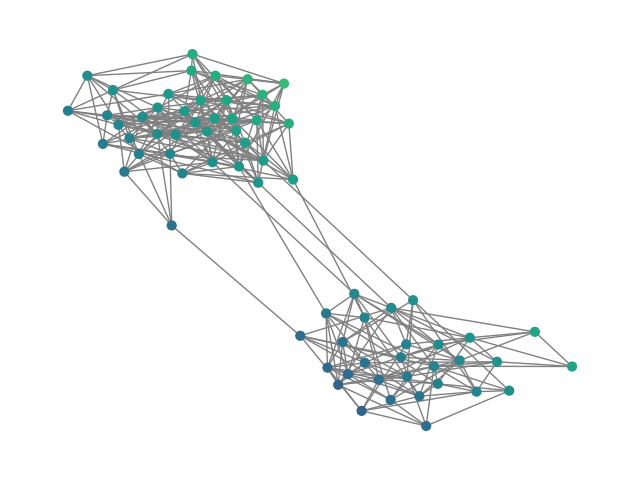} &
        \includegraphics[width=0.15\textwidth]{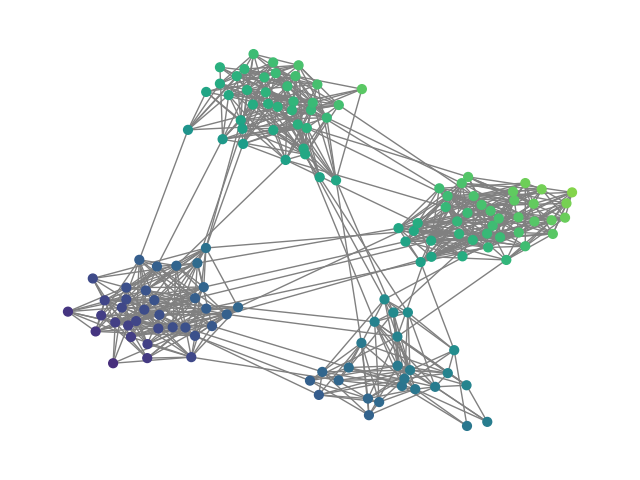}&
        \includegraphics[width=0.15\textwidth]{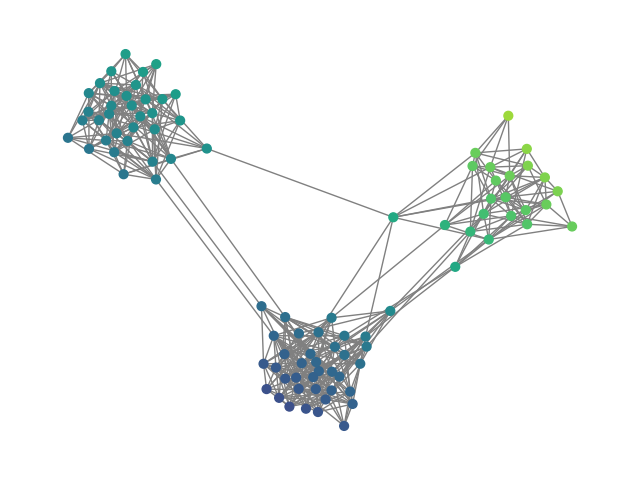} 
        \\
        \includegraphics[width=0.15\textwidth]{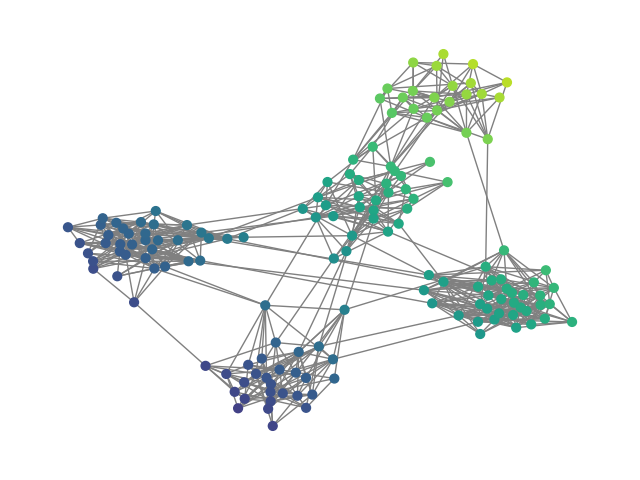} &
        \includegraphics[width=0.15\textwidth]{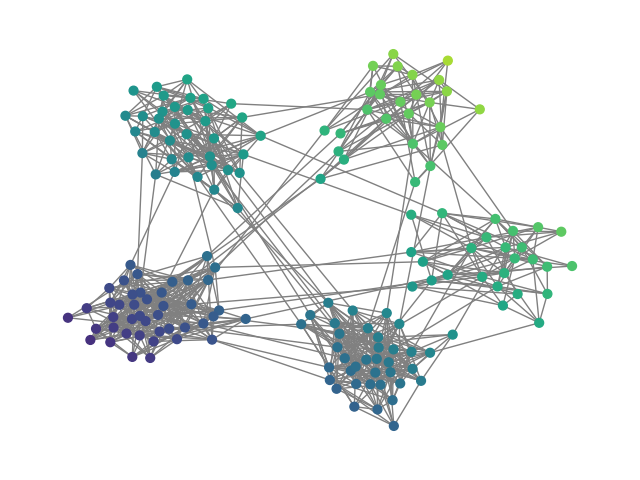} &
        \includegraphics[width=0.15\textwidth]{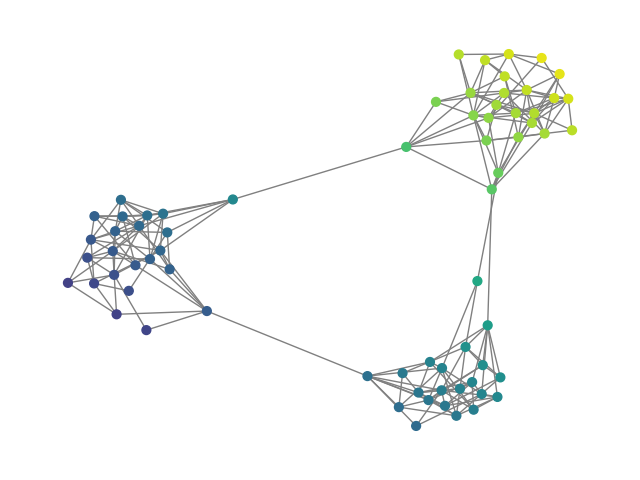} &
        \includegraphics[width=0.15\textwidth]{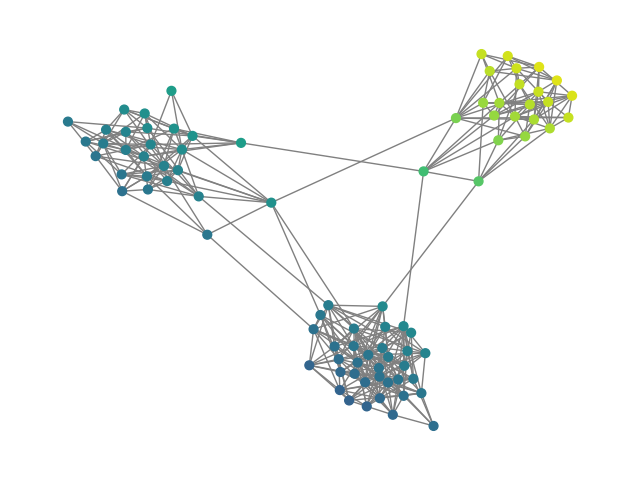} &
        \includegraphics[width=0.15\textwidth]{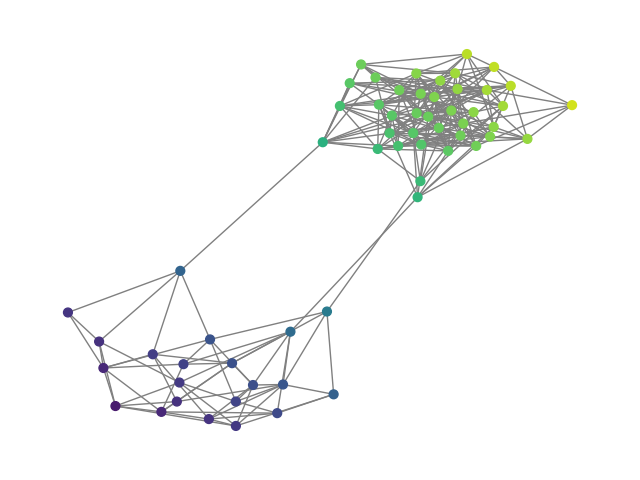}&
        \includegraphics[width=0.15\textwidth]{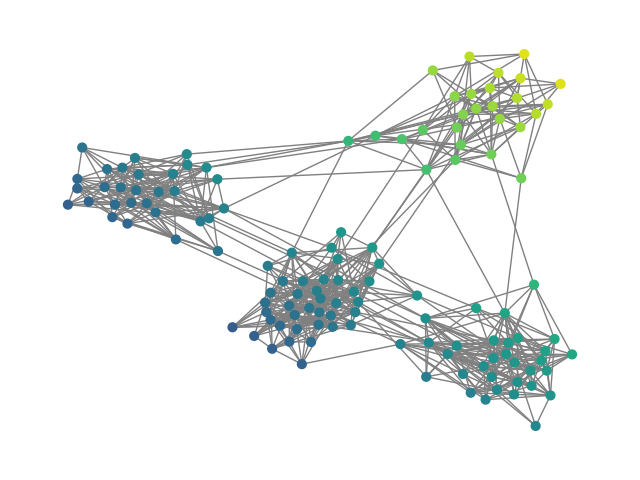} 
        \\
        \includegraphics[width=0.15\textwidth]{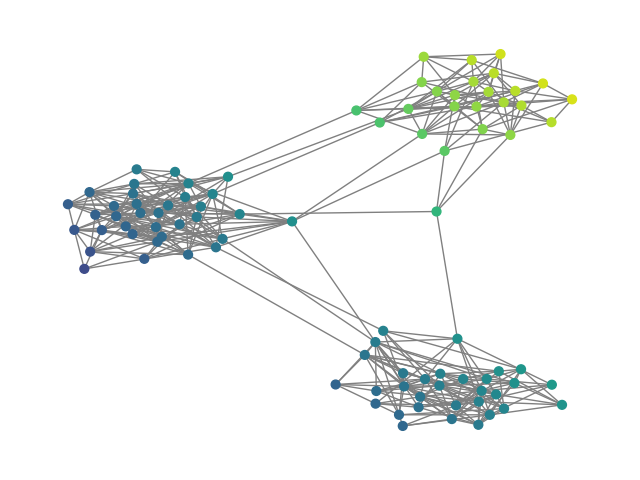} &
        \includegraphics[width=0.15\textwidth]{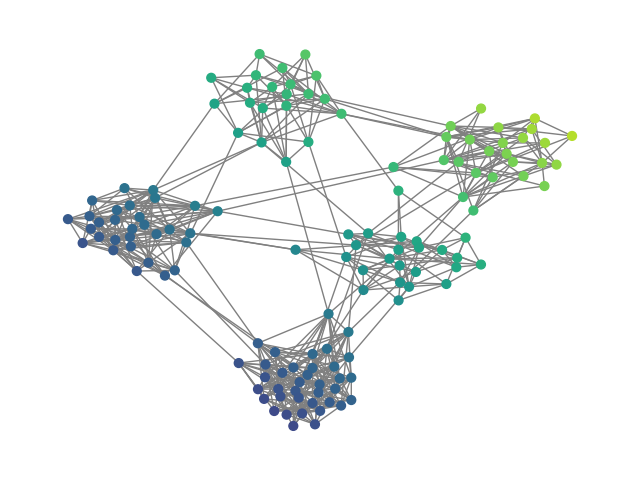} &
        \includegraphics[width=0.15\textwidth]{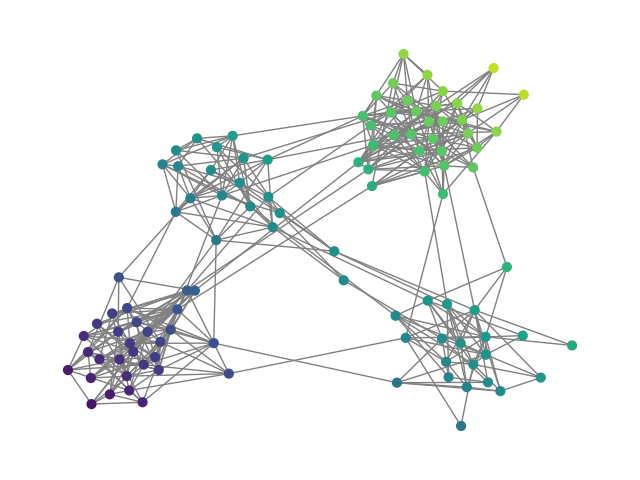} &
        \includegraphics[width=0.15\textwidth]{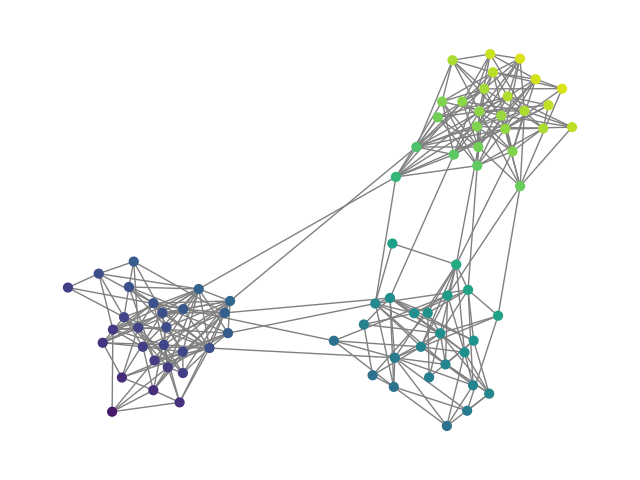} &
        \includegraphics[width=0.15\textwidth]{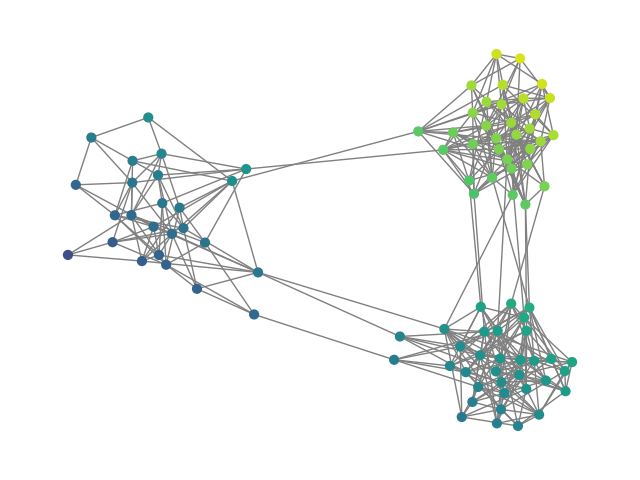}&
        \includegraphics[width=0.15\textwidth]{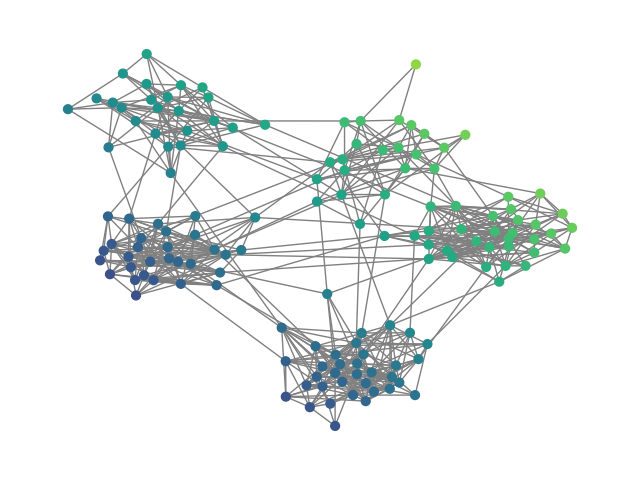} 
        \\
        \hline
        \end{tabular}
        \end{sc}
\end{figure}



\end{document}